\documentclass{article}

\PassOptionsToPackage{numbers, compress}{natbib}

\usepackage[preprint]{neurips_2024}

\usepackage[utf8]{inputenc} 
\usepackage[T1]{fontenc}    
\usepackage{url}            
\usepackage{booktabs}       
\usepackage{amsfonts}       
\usepackage{nicefrac}       
\usepackage{microtype}      
\usepackage{xcolor}         
\usepackage{times}
\usepackage{epsfig}
\usepackage{amsmath}
\usepackage{makecell} 
\usepackage{tabularx}
\usepackage{diagbox}
\usepackage{enumitem} 

\usepackage{xspace}
\usepackage{graphicx}
\usepackage{subcaption}
\usepackage{multirow} 
\usepackage{colortbl}
\usepackage{pifont}
\usepackage{wrapfig}
\usepackage{tcolorbox}
\usepackage{listings}
\usepackage{float}
\usepackage{marvosym}
\usepackage{CJKutf8}
\usepackage{xcolor}
\definecolor{citecolor}{HTML}{0071BC}
\usepackage[
    pagebackref,           
    breaklinks=true,       
    colorlinks=true,       
    linkcolor=red,         
    citecolor=citecolor,   
    urlcolor=black,        
    bookmarks=false        
]{hyperref}

\usepackage{array}

\title{
AndesVL Technical Report: An Efficient Mobile-side Multimodal Large Language Model
}

\author{
\textbf{AndesVL Team, OPPO AI Center}
\\
\github ~ \url{\ghlink}
\\
\huggingface ~ \url{\hflink}
}

\definecolor{baselinecolor}{gray}{.9}
\definecolor{reduce-color}{RGB}{67,178,68}

\begin{document}
\begin{CJK}{UTF8}{gbsn} 

\maketitle

\begin{abstract}
 
In recent years, while cloud-based MLLMs such as QwenVL, InternVL, GPT-4o, Gemini, and Claude Sonnet have demonstrated outstanding performance with enormous model sizes reaching hundreds of billions of parameters, they significantly surpass the limitations in memory, power consumption, and computing capacity of edge devices such as mobile phones. This paper introduces AndesVL, a suite of mobile-side MLLMs with 0.6B to 4B parameters based on Qwen3's LLM and various visual encoders. We comprehensively outline the model architectures, training pipeline, and training data of AndesVL, which achieves first-tier performance across a wide range of open-source benchmarks, including fields such as text-rich image understanding, reasoning and math, multi-image comprehension, general VQA, hallucination mitigation, multilingual understanding, and GUI-related tasks when compared with state-of-the-art models of a similar scale. Furthermore, we introduce a 1+N LoRA architecture alongside a Quantization-Aware LoRA Fine-Tuning (QALFT) framework to facilitate efficient task adaptation and model compression during mobile-side deployment of AndesVL. Our QALFT experiments reveal that AndesVL maintains performance with only ignorable degradation (3\%) after deployment on mobile devices compared to the original floating-point model.
Moreover, utilizing our cache eviction algorithm---OKV---along with customized speculative decoding and compression strategies, we achieve a 6.7x peak decoding speedup ratio, up to 30.9\% memory reduction, and 1.8 bits-per-weight when deploying AndesVL-4B on MediaTek Dimensity 9500 chips. We release all models on \url{https://huggingface.co/OPPOer}.

\end{abstract}

\section{Introduction}

\begin{figure}[tb]
    \centering
    \includegraphics[width=0.75\linewidth]{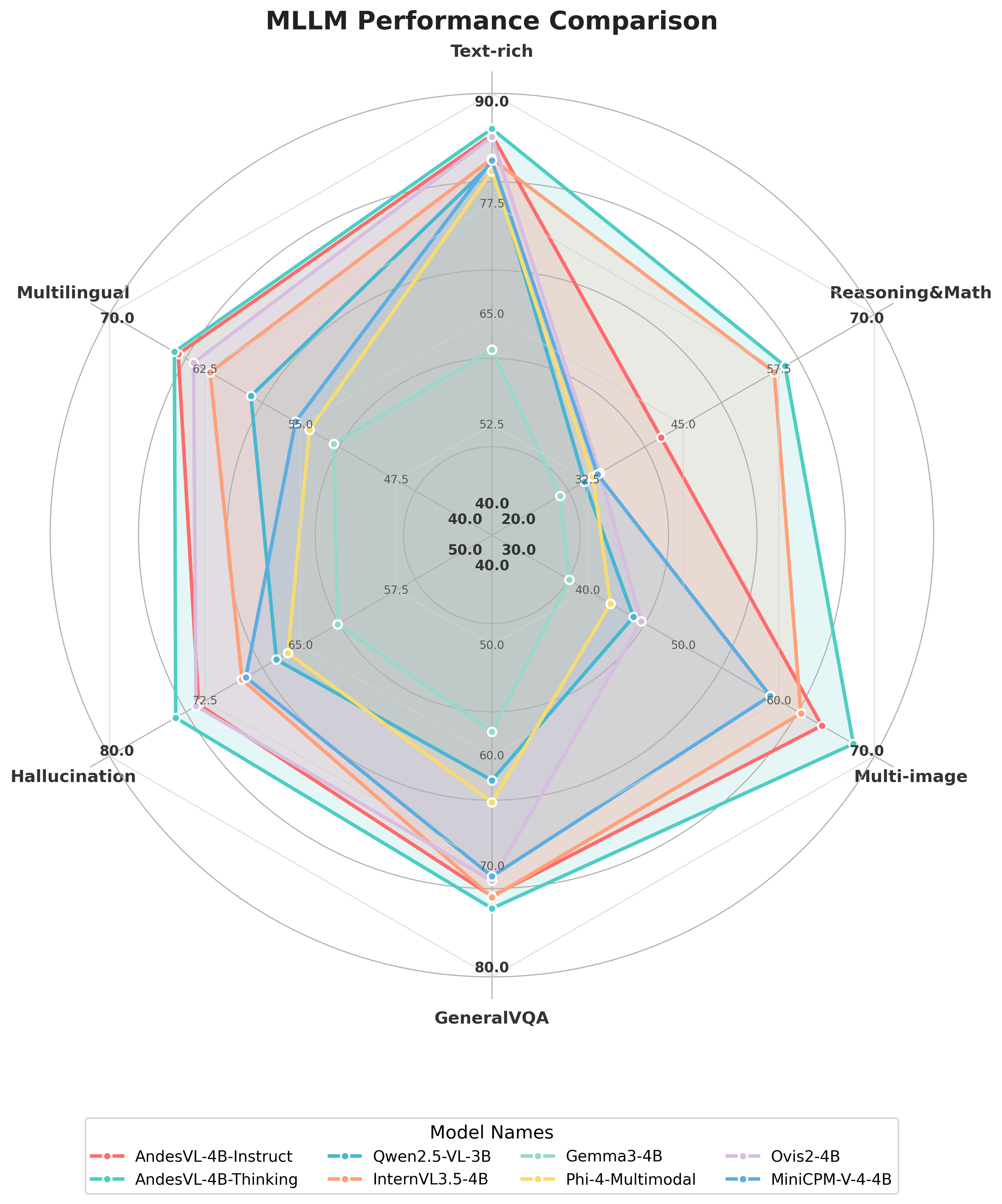}
    \caption{Overall performance comparisons over 6 domains (text-rich, reasoning and math, general VQA, multi-image, multilingual and hallucination) of different SOTA MLLMs with 4B parameters.}
    \label{fig:4b_performance}
\end{figure}

In recent years, the advent of large language models (LLMs) represented by ChatGPT~\cite{chatgpt}, the Qwen series~\cite{bai2023qwen, yang2024qwen2, yang2024qwen2.5, yang2025qwen3}, and the DeepSeek series~\cite{bi2024deepseekllm, liu2024deepseek, guo2025deepseek} has ushered in a new era of artificial intelligence. These LLMs have demonstrated remarkable capabilities in natural language processing tasks, such as text generation, question answering, and language translation. Building upon the success of LLMs, multimodal large language models (MLLMs) have emerged, expanding the functionality of large models from pure text to multiple modalities. MLLMs incorporate modalities such as image, video, and even audio, enabling more diverse and comprehensive interactions.

The typical training paradigm of MLLMs involves leveraging a pre-trained LLM. By aligning the LLMs with visual encoders and engaging in continual pre-training and fine-tuning, an MLLM that can process multimodal inputs and generate text outputs is developed. For effective training, a substantial amount of multimodal data is necessary, in addition to extensive text datasets. This data encompasses image-text pairs, optical character recognition (OCR) data, and visual question-answering (VQA) data. These datasets provide the model with a wide range of multimodal capabilities, such as image captioning, OCR, chart question answering, visual semantic recognition, and visual reasoning.

On the cloud side, there are numerous outstanding MLLMs. Models such as the QwenVL series~\cite{bai2023qwenvl, wang2024qwen2vl, bai2025qwen2_5}, the InternVL series~\cite{chen2023internvl, chen2024internvl_1_5, chen2024expanding, zhu2025internvl3, wang2025internvl3}, GPT-4o~\cite{chatgpt4o}, Gemini~\cite{team2023gemini, reid2024gemini1_5, gemini2_0, gemini2_0pro, geminipro2.5} and Claude Sonnet~\cite{claude3series2024} have demonstrated SOTA competence in multimodal tasks. Despite their groundbreaking performance, these models are generally oriented towards reaching maximum performance, involving parameter sizes running into hundreds of billions. Such large-scale parameter demands significantly exceed the processing capabilities of mobile devices such as smartphones and tablets, particularly in terms of memory capacity, running speed, and computing power of chips. Consequently, MLLMs that typically run on mobile platforms are limited to approximately 4 billion parameters, as illustrated by Qwen2.5-VL-3B~\cite{bai2025qwen2_5} and InternVL3.5-4B~\cite{wang2025internvl3}. To maintain optimal functionality on mobile hardware, additional measures, such as quantization-aware training (QAT) and deployment optimization on the mobile side, are essential.

Currently, only a limited number of mobile-device manufacturers and internet companies have started exploring mobile-side MLLMs. For example, vivo has introduced BlueLM-V-3B~\cite{lu2024bluelm} and BlueLM-2.5-3B~\cite{xiong2025bluelm}, Meituan has launched the MobileVLM series~\cite{chu2023mobilevlm, chu2024mobilevlm}, Xiaomi has concentrated on mobile agents with the development of MobileVLM~\cite{wu2024mobilevlm}, and Apple has released the Ferret-UI series~\cite{you2024ferret, li2024ferret} aimed at UI comprehension. Despite these initiatives, a thorough study explicating the training process, deployment strategies, and performance assessments on both general and mobile-specific benchmarks of mobile-side MLLMs is still absent.

In this paper, we introduce the AndesVL suite. By integrating Qwen3~\cite{yang2025qwen3} LLMs and various visual encoders, we have successfully developed mobile-side MLLMs with parameter sizes ranging from 0.6B to 4B. Our models focus on several key general-purpose capabilities, including knowledge acquisition, mathematical reasoning, handling text-rich content, dealing with hallucination issues, processing multi-image and multilingual inputs, and general VQA. We thoroughly introduce the model architectures, training pipeline, and data preparation strategies. Additionally, we have given special consideration to functions crucial for mobile-side MLLMs, such as user interface (UI) understanding. To evaluate the performance of our models, we have developed mobile-specific benchmarks. Specifically, as inspired by Qwen3-4B-Instruct and Qwen3-4B-Thinking~\cite{yang2025qwen3}, we propose to train the instruct and thinking models of AndesVL separately to achieve the best instruct-following and reasoning abilities, respectively.  Our floating-point models have achieved first-tier results among models of similar sizes across various benchmarks, as shown in Fig. \ref{fig:4b_performance}, including 32 open-source benchmarks related to the domains mentioned above.

For practical application on mobile devices, we have also designed a 1 + N Low-Rank Adaptation (LoRA)~\cite{hu2022lora} architecture to make the model adaptable to different tasks. Based on the AndesVL backbone model, downstream tasks can be clustered, and similar tasks can be fine-tuned using a single LoRA module to achieve optimal performance in specific domains. 
In addition to floating-point models within the 4B parameter range, to enable large models to run on the mobile side, the QAT and Quantization-Aware Lora Fine Tuning (QALFT) frameworks are necessary for model compression. With this pipeline, our mobile-side models have also demonstrated excellent results in various realistic applications. Additionally, we meticulously design a comprehensive mobile-side acceleration suite, with cache eviction, speculative decoding and sparsification, which achieve a block efficiency (BE) of up to 7.9, with about a 6.7x end-to-end decoding speedup over the baseline (with auto-regressive decoding and without compression optimization). Furthermore, we achieve a memory reduction of up to 30.9\% and a weight compression of 1.8 bits-per-weight when deploying AndesVL-4B on MediaTek Dimensity 9500 chips. This work represents a significant step forward in the development and deployment of mobile-side MLLMs. 

The structure of this article is as follows: Sec.~\ref{sec:related_works} introduces the mobile-side MLLM and related work on mobile-side deployment. Sec.~\ref{sec:andesvl} focuses on the model architecture, training data, and training scheme of AndesVL. Sec.~\ref{sec:mobile-side_depoloy_andesvl} introduces the mobile-side 1+N LoRA training architecture of AndesVL and the technical scheme for mobile-side deployment. Sec.~\ref{sec:experiments} details the performance of AndesVL on public benchmarks and self-built mobile-side benchmarks, as well as its comparison with SOTA models. Sec.~\ref{sec:experiments_on-device} presents the benchmark results and mobile-side performance of AndesVL after being deployed on mobile devices. Sec.~\ref{sec:future_directions} looks ahead to future directions. Sec.~\ref{sec:conclusion} summarizes the entire article.

The main contributions of this article can be summarized as follows:
\begin{itemize}
    \item First, addressing the speed and performance trade-off for mobile implementations of MLLM, we introduce the AndesVL suite, which is a collection of MLLMs designed for efficient deployment on edge devices, with parameter scales ranging from 0.6B to 4B, demonstrating competitive performance with SOTA models with comparable parameters.
    \item Second, we offer separate models for Instruct and Thinking versions, making each ideal for tackling the challenges associated with high-efficiency tasks in understanding and generation, as well as applications in complex mathematical reasoning and planning.
    \item Third, we design a 1+N Lora training pipeline for mobile deployment, which enables efficient task clustering and adaptation. We further propose the QALFT framework to ensure flexible application of the 1+N Lora architecture on mobile devices. 
    \item Finally, based on our mobile-side acceleration and compression strategies, \textit{e.g.}, customized cache eviction, sparsification, and speculative decoding, AndesVL-4B can achieve a 6.7x peak decoding speedup ratio, a memory reduction of up to 30.9\%, and 1.8 bits-per-weight on MediaTek Dimensity 9500 chips.
\end{itemize}

\section{Related Works}
\label{sec:related_works}

\subsection{Mobile-side MLLMs}
\label{sec:mobile-side_mllms}


Recent years have witnessed a proliferation of remarkable advances in MLLMs. Numerous remarkable MLLMs~\cite{bai2023qwenvl, wang2024qwen2vl, bai2025qwen2, chen2023internvl, chen2024internvl_1_5, chen2024expanding, zhu2025internvl3, wang2025internvl3, chatgpt4o, team2023gemini, reid2024gemini1_5, gemini2_0, gemini2_0pro, geminipro2.5, claude3series2024} have been introduced, primarily driven by the pursuit of exploring the upper bounds of model performance through scaling laws. This endeavor has resulted in models with astronomically large parameter counts, reaching hundreds of billions or even trillions. Nonetheless, this emphasis on large-scale models has left the development of mobile-side MLLMs relatively underexplored.

Among the efforts towards more mobile-friendly MLLMs, the Qwen series has made notable progress. Qwen2-VL~\cite{wang2024qwen2vl} and Qwen2.5-VL~\cite{bai2025qwen2} introduced model sizes of 2B and 3B, respectively, which are particularly suited for deployment on mobile devices. These model sizes effectively balance performance and the computational limitations of mobile hardware. Similarly, the InternVL series~\cite{chen2024expanding, zhu2025internvl3, wang2025internvl3} presented a range of model sizes---1B, 2B, and 4B---designed to fulfill various operational needs on mobile platforms.

In 2023, Meituan emerged as a pioneer in the mobile MLLM domain with the introduction of MobileVLM~\cite{chu2023mobilevlm}. Built upon MobileLLaMA in a LLaVA-like~\cite{liu2023llava} architecture, MobileVLM came in 1.7B and 3B model sizes. It achieved SOTA results in some benchmarks for models of similar sizes at that time. Meituan offered significant insights into the processing speeds on mobile and IoT platforms, reporting rates of 12.21 and 21.54 tokens per second, respectively. In 2024, the release of MobileVLM V2~\cite{chu2024mobilevlm} further advanced the field by exploring the data scaling law, improving training strategies, and optimizing the modality alignment design. These developments contributed to a comprehensive enhancement in the performance of the MobileVLM framework.

In 2024, the Apple MM series~\cite{mckinzie2024mm1, zhang2024mm1_5} demonstrated that even relatively compact models, specifically those with 1B and 3B parameters, could achieve impressive performance through meticulous data curation and optimized training strategies. The Ferret UI series~\cite{you2024ferret, li2024ferret} marked a significant step forward, as it was the first series extensively dedicated to improving the capabilities of screen UI understanding. It extended the capabilities of MLLMs to tasks such as referring and grounding on mobile UI screens and answering questions related to screen operations. However, Apple did not reveal the performance metrics for these models when deployed on mobile platforms.

Xiaomi's MobileVLM~\cite{wu2024mobilevlm} also made important contributions by leveraging carefully constructed UI understanding and APP operation trajectory data. This enabled the model to expand its capabilities from understanding within a single UI (intra-UI) to understanding and operating across multiple UIs (inter-UI). Nevertheless, Xiaomi's 9.8B MobileVLM model was not successfully deployed on mobile devices.

Finally, vivo's BlueLM-V-3B~\cite{lu2024bluelm} and BlueLM-2.5-3B~\cite{xiong2025bluelm} achieved mobile-side deployment of an MLLM through systematic optimizations in algorithms and hardware deployment. Specifically, BlueLM-V-3B achieved a running memory of 2.2G and a token throughput speed of 24.4 tokens/s on MediaTek Dimensity 9300 NPUs. This not only showcases its effectiveness but also provides practical performance metrics for mobile-side MLLMs.

Despite these efforts, there remains a gap in comprehensively documenting training processes, deployment solutions, and benchmark results for general and mobile-specific tasks of mobile-side MLLMs. Our work aims to fill this void by presenting the AndesVL suite, which offers a comprehensive approach to mobile-side MLLMs, including detailed training, deployment, and benchmarking aspects. 

\subsection{Mobile-Side Deployment of MLLM}
\label{sec:mobile-side_deploy}

The deployment of MLLMs on mobile devices presents unique challenges, including limited computational resources, diverse hardware architectures, and stringent energy constraints. To address these issues, various solutions~\cite{ONNX-Runtime, llamacpp, MLC, iyer2023automated, jiang2020mnn, li2024transformer, genimi-nano, apple-core-ML} have been proposed that take advantage of CPUs, GPUs, and NPUs. 

\paragraph{CPU-based Deployment}

In 2020, Alibaba developed the Mobile Neural Network (MNN)~\cite{jiang2020mnn}, an inference engine tailored for mobile applications. It introduces a ``pre-inference'' mechanism for runtime optimization, thorough kernel optimizations for optimal computation performance, and a back-end abstraction module that enables hybrid scheduling while maintaining a lightweight engine footprint on mobile CPUs.

In 2023, Georgi Gerganov~\cite{llamacpp} introduced llama.cpp, a lightweight, dependency-free C/C++ implementation designed for efficient LLM inference across diverse hardware platforms, including mobile CPUs. It includes support for several quantization levels (ranging from 1.5-bit to 8-bit), enabling reduced memory consumption and accelerated inference.

\paragraph{GPU-based Deployment}

In 2024, a machine learning compiler and high-performance deployment engine for LLMs, MLC LLM~\cite{MLC}, was developed, aiming to enable native deployment across various platforms, including mobile GPUs. It compiles models into optimized binaries compatible with platforms such as iOS, Android, and web browsers.

In addition, Li et al.~\cite{li2024transformer} proposed Transformer-Lite, which focuses on the high-efficiency deployment of LLM on mobile phone GPUs. It introduced four optimization techniques: a symbolic expression-based approach for dynamic shape model inference, operator optimizations with execution priority settings, an FP4 quantization method termed M0E4 to reduce dequantization overhead, and a sub-tensor-based technique to eliminate the need for copying key-value (KV) cache after inference. These optimizations enable significant speedups in both prefill and decoding phases compared to existing CPU-based and GPU-based inference engines.

\paragraph{NPU-based Deployment}

Gemini Nano~\cite{genimi-nano}, developed by Google, is designed for on-device use cases, running within Android's AICore system service to leverage device hardware for low-latency inference. It is accessible through the AI Edge SDK, which allows developers to customize the inference and prompts. Gemini Nano models, such as Nano-1 (1.8B parameters) and Nano-2 (3.25B parameters), are distilled from larger Gemini models and optimized for edge devices such as smartphones.

Finally, Apple's On-Device Deployment utilizes the Core ML framework to optimize and deploy large language models on Apple silicon~\cite{apple-core-ML}. Techniques such as grouped-query attention (GQA) mechanisms, mixed 2-bit and 4-bit quantization, and efficient memory management strategies enable the deployment of models like Llama-3.1-8B-Instruct on devices such as the iPhone 15 Pro, achieving decoding speeds of approximately 30 tokens per second.

Despite notable progress in mobile-side deployment of MLLMs, several challenges persist. These include balancing model performance with resource constraints, ensuring cross-device compatibility, standardizing deployment processes, and establishing comprehensive evaluation frameworks for multimodal tasks. To address these issues, we introduce the AndesVL series, which offers a comprehensive suite of optimized deployment solutions tailored for mobile platforms. This includes detailed training methodologies, quantization techniques, compilation strategies, and hardware-specific optimizations. Our work aims to bridge existing gaps, providing a robust foundation for future research and practical applications in mobile-side MLLM deployment.

\section{AndesVL}
\label{sec:andesvl}
\subsection{Model Architecture}
\begin{figure}
    \centering
    \includegraphics[width=0.9\linewidth]{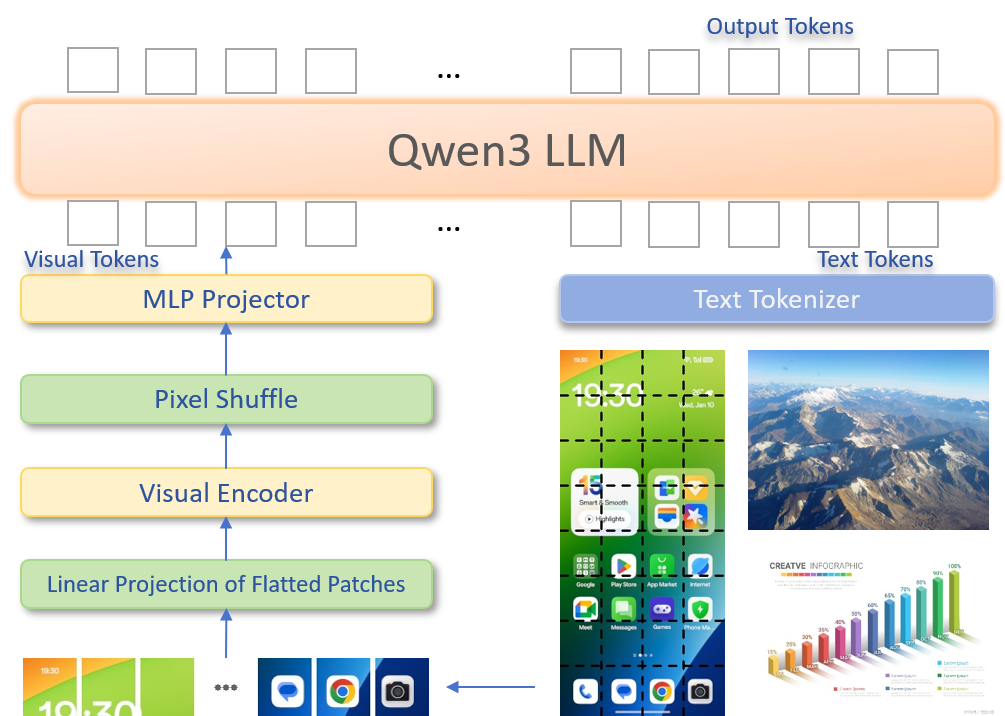}
    \caption{The overall architecture of AndesVL mainly includes a visual encoder, an MLP projector, and an LLM.}
    \label{fig:overall_architecture}
\end{figure}

\begin{table*}[htbp]
\centering
\begin{tabular}{lccc}
\hline
\textbf{Model Name} & \textbf{\#Param (B)} & \textbf{Vision Encoder} & \textbf{Language Model} \\ 
\hline
AndesVL-0.6B & 0.695 & SigLIP2-Base~\cite{tschannen2025siglip} & Qwen3-0.6B~\cite{yang2025qwen3} \\
AndesVL-1B   & 0.927 & AIMv2-Large~\cite{fini2025multimodal} & Qwen3-0.6B~\cite{yang2025qwen3} \\
AndesVL-2B   & 2.055 & AIMv2-Large~\cite{fini2025multimodal} & Qwen3-1.7B~\cite{yang2025qwen3} \\
AndesVL-4B   & 4.360 & AIMv2-Large~\cite{fini2025multimodal} & Qwen3-4B~\cite{yang2025qwen3} \\
\hline
\end{tabular}
\caption{AndesVL model architectures of different sizes.}
\label{tab:arch}
\end{table*}

AndesVL comprises models with parameters ranging from 0.6B to 4B parameters, with detailed architectures provided in Table~\ref{tab:arch}. Following the paradigm of typical MLLMs~\cite{liu2023llava, chen2024internevo, wang2024qwen2vl}, it consists of three fundamental components: a visual encoder, an MLP projector, and an LLM, as illustrated in Fig.~\ref{fig:overall_architecture}.

As a general-purpose MLLM, AndesVL is designed to handle image inputs with arbitrary aspect ratios and resolutions. To achieve this, we avoid the image cropping methods employed in other works \cite{liu2024llavanext, chen2024internevo, xu2024llava_uhd} and instead implement a Native Resolution ViT (NaViT) \cite{dehghani2023patch}-based strategy, allowing the visual encoder to process input of any resolution directly. This method is particularly beneficial for efficiently processing low-resolution images and ensures consistency between model inputs and the original data. The MLP projector includes two MLP layers used to align the ViT output with the LLM's embedding layer. To decrease the sequence length of the ViT output going into the LLM, a straightforward yet adaptable pixel shuffle operation is used to reduce the sequence length to a quarter of its original size. This operation combines and concatenates the data from adjacent 4×4 patches before passing them to the MLP. For the language model, AndesVL employs Qwen3~\cite{yang2025qwen3}, utilizing the 0.6B, 1.7B, and 4B models from the Qwen3 series. To save memory for embedding parameters, we preserve the tied word embeddings configuration across all LLM variations.

\subsection{Training Pipeline}

In this paper, inspired by the recently released Qwen3-4B-Instruct and Thinking models~\cite{yang2025qwen3}, we propose to develop AndesVL in two distinct models: an instruction model (supporting only non-thinking mode) and a thinking model (supporting both non-thinking and thinking modes). Consequently, the training datasets and methodologies are distinct from one another, as will be thoroughly explained in the subsequent subsections.

\begin{table}[tbh]
\centering
\resizebox{1\textwidth}{!}{
\begin{tabular}{l|ccc}
\toprule
\textbf{Stages} & \textbf{Visual-Language alignment} & \textbf{Joint V-L pre-training} & \textbf{Multi-task pre-training} \\
\midrule
\multirow{2}{*}{Main data type} & Caption & Interleaved image-text & All multi-task data \\
                        & + OCR + VQA & + Pure text + VQA + Long CoT* & + Long CoT* \\
\hline
Trainable parameters & ViT + MLP & Full model & Full model \\
ViT sequence length & 4,096 / 16,384 & 4,096 & 16,384 \\
LLM sequence length & 2,048 / 8,192  & 8,192 & 32,768 \\
Trained tokens & 100B / 100B & 160B & 160B \\
\bottomrule
\end{tabular}}
\caption{Pre-training stages of AndesVL. * indicates data exclusively used for the Thinking models.}
\label{tab:training_stages}
\end{table}

\subsubsection{Pre-train}
As illustrated in Table~\ref{tab:training_stages}, the pre-training phase of AndesVL consists of three stages: vision-language alignment, joint vision-language pre-training, and multi-task pre-training. To improve training efficiency, we pack not only the ViT input but also the LLM input tokens. 
Our pre-training commences with the Qwen3-Instruct/Thinking versions of the language model. Throughout all pre-training stages, we incorporate a proportion of instruction-following data. This allows us to maintain the model's instruction-following capabilities and monitor its performance progression directly via instruction-based evaluation. 

\paragraph{Vision-Language Alignment}
Our primary visual encoder leverages AIMv2-Large~\cite{fini2025multimodal}, a compact 300M parameter model that offers superior power efficiency compared to larger alternatives such as Qwen2VL-ViT-675M~\cite{wang2024qwen2vl}, making it particularly well-suited for mobile deployment. To enhance the encoder's versatility across varying input resolutions, we integrate 2D Rotary Position Embeddings (2D-RoPE)~\cite{kexuefm-8397}, whose strong extrapolation capabilities enable our vision encoder to effectively process high-resolution inputs even when trained on lower resolutions. To maintain model performance, we preserve the original position embeddings with a length of 1024 and adapt them to different resolutions using bicubic interpolation. 
\par
We employ a two-stage training procedure for the visual encoder within our MLLM framework, keeping the LLM frozen while utilizing diverse training data from captions, OCR, and VQA tasks. The first stage processes 100B tokens with a ViT sequence length of 4,096, applying higher learning rates specifically to the randomly initialized MLP layers, while the second stage continues with an additional 100B tokens with a ViT sequence length of 16,384. For our 1B and 2B model variants, we streamline the training process by directly leveraging the vision encoder from our 4B model and performing alignment by training the MLP layer only. For our most compact 0.6B model variant, we adopt the SigLIP2-Base-Patch16-512~\cite{tschannen2025siglip} model, which follows a similar adaptation strategy that combines bicubic interpolation for position embeddings with 2D-RoPE and two-stage training.

\paragraph{Joint Vision-Language Pre-training}
The second stage involves joint vision-language pre-training. After the visual encoder's output aligns well with the LLM's representations, we unfreeze the LLM parameters and conduct full-parameter pretraining using a relatively low learning rate.
In this stage, we utilize a large volume of unsupervised interleaved image-text data, enabling the model to acquire extensive visual knowledge. During pre-training on this data, we compute loss only on text tokens, excluding image tokens from the calculation.
\par
In unidirectional autoregressive transformers, inappropriate image positioning may prevent the model from learning multimodal knowledge effectively. For instance, images placed at the end of a sequence cannot contribute to learning even when encoded. To mitigate this issue, we employed a strategy where, with 50\% probability, we maintained the original image positions. With the remaining 50\% probability, we relocated all images in the data to precede all text content, replacing the images with their corresponding indices. Fig.~\ref{fig:image_reposition} illustrates this transformation.

\begin{figure}[h]
\centering
\begin{tcolorbox}[colback=gray!5, colframe=gray!50, boxrule=0.5pt, arc=2pt, left=5pt, right=5pt, top=5pt, bottom=5pt]
\textbf{Original interleaved document:}
\begin{lstlisting}[basicstyle=\small\ttfamily, breaklines=true, frame=none]
The sunset over the Pacific Ocean was breathtaking. 
<img>pacific_sunset.jpg</img> The vibrant colors painted 
the sky in shades of orange and pink. Later that evening, 
we hiked to the mountain viewpoint. <img>mountain_vista.jpg</img>
\end{lstlisting}

\vspace{0.3em}
\textbf{Transformed format:}
\begin{lstlisting}[basicstyle=\small\ttfamily, breaklines=true, frame=none]
<|image_0|> <img>pacific_sunset.jpg</img>
<|image_1|> <img>mountain_vista.jpg</img>
The sunset over the Pacific Ocean was breathtaking. 
<|image_0|> The vibrant colors painted the sky in shades 
of orange and pink. Later that evening, we hiked to the 
mountain viewpoint. <|image_1|>
\end{lstlisting}
\end{tcolorbox}
\caption{Image repositioning strategy for joint vision-language pre-training. Images are moved to the beginning of the sequence with 50\% probability to ensure effective multimodal learning.}
\label{fig:image_reposition}
\end{figure}

Since interleaved image-text data can be viewed as a multimodal extension of unlabeled text data, we also incorporate unlabeled text data from text pre-training. To maintain instruction-following capabilities, we include text instruction data in this stage as well. Furthermore, a certain proportion of multi-task pre-training data is added to enhance the model's overall multimodal abilities. For the Thinking version of the model, we additionally incorporate long CoT data, which will be detailed in Sec.~\ref{sec:pre-train_data}.

\paragraph{Multi-task Pre-training}
The final stage is multi-task pre-training. In this stage, our approach transitions from self-supervised learning with unsupervised data to supervised learning using annotated data, focusing solely on calculating the text token loss for the answer portions. Data types mainly consist of general VQA, captions, and OCR, alongside task-specific data like grounding/UI. For the Thinking model variant, we continue to incorporate long CoT data as in the previous stage, while increasing the proportion of multimodal types to enhance its step-by-step reasoning capabilities with visual inputs. Although we use 2D RoPE to allow model inference at high resolutions, we increased the ViT patch input from 4,096 to 16,384 to facilitate learning from data that require high resolution. To enhance the LLM's capabilities in long contexts, particularly its reasoning ability in Thinking mode, we expanded the LLM's sequence length from 8,192 to 32,768.

Consequently, by completing the three pre-training stages mentioned above, we developed the base versions for our Instruct and Thinking models, referred to as \textit{AndesVL-Instruct-Base} and \textit{AndesVL-Thinking-Base}, respectively, which are subsequently utilized for post-training.

\subsubsection{Post-train}

The AndesVL post-training process consists of two main stages: supervised fine-tuning (SFT) and reinforcement learning (RL). SFT is utilized for both instruction and thinking models. Notably, mixed preference optimization (MPO)~\cite{wang2024mpo} is adopted for refining the instruction models, while Group Relative Policy Optimization (GRPO)~\cite{guo2025deepseek} is employed for the thinking models. Following the application of SFT and MPO to AndesVL-Base, we derive the \textit{AndesVL-Instruct} model. Conversely, the \textit{AndesVL-Thinking} model is attained through the application of SFT and GRPO.


\paragraph{SFT}
Supervised fine-tuning (SFT) of the pre-trained AndesVL model is conducted utilizing meticulously formatted instruction data. Recognizing the critical influence of data diversity and quality on the performance of downstream tasks, an extensive array of multimodal instruction data is compiled, covering a wide range of task areas. To improve the model's conversational abilities, the Chat-ML instruction data format is employed.

The instruction dataset is meticulously crafted to introduce the model to multiple input modalities, enabling the development of strong representational learning capabilities. Additionally, the dataset encompasses a diverse range of task objectives, such as image captioning, visual question answering, text summarization, and code generation. This deliberate diversification in data sources and task outlines is designed to enhance the model's generalization capacity and remain robust across various application scenarios. Compliance with the Chat-ML format supports seamless integration with contemporary dialogue-oriented systems, thus promoting coherent and informative conversation exchanges. This strategic SFT method is essential for unlocking the full potential of the AndesVL model, thereby facilitating its effective use in real-world scenarios.

\paragraph{MPO}
Direct preference optimization (DPO) has emerged as the dominant approach for aligning LLMs with human preferences, as highlighted in \cite{rafailov2024dpo}, which can avoid complex on-policy RL pipelines and is suitable for training non-thinking models. Leveraging its effectiveness in language processing, recent research has extended the application of DPO to multimodal settings \cite{li2023silkie, zhou2024aligning}. Nonetheless, two challenges arise when implementing DPO in MLLM: the scarcity of comprehensive, high-quality multimodal reasoning preference datasets and DPO's inability to assess the absolute quality of individual responses. To address these issues, a novel approach known as Mixed Preference Optimization (MPO) was introduced by~\cite{wang2024mpo}, which has shown enhancements across various multimodal reasoning evaluation sets.

We borrowed the MMPR dataset and MPO from \cite{wang2024mpo}. During the training process, a joint loss consisting of preference loss $\mathcal{L}_p$, quality loss $\mathcal{L}_q$, and generation loss $\mathcal{L}_g$ was used, which can be formulated as
$$
\mathcal{L} = w_p \mathcal{L}_p + w_q \mathcal{L}_q + w_g \mathcal{L}_g.
$$
The preference loss $\mathcal{L}_p$ is formulated as
\begin{equation}
    \mathcal{L}_{p}=-\log \sigma\left(\beta \log \frac{\pi_{\theta}\left(y_{c} \mid x\right)}{\pi_{0}\left(y_{c} \mid x\right)}-\beta \log \frac{\pi_{\theta}\left(y_{r} \mid x\right)}{\pi_{0}\left(y_{r} \mid x\right)}\right),
\end{equation}
where $\beta$ is the KL penalty coefficient, $x$, $y_{c}$, and $y_{r}$ are user query, chosen response, and rejected response,
respectively. The policy model  $\pi_{\theta}$  is initialized from model $\pi_{0}$.

The quality loss $\mathcal{L}_q$ is formulated as
\begin{equation}
    \mathcal{L}_{q}=-\log \sigma(\beta \log \frac{\pi_{\theta}\left(y_{c} \mid x\right)}{\pi_{0}\left(y_{c} \mid x\right)}-\delta)-\log \sigma(-(\beta \log \frac{\pi_{\theta}\left(y_{r} \mid x\right)}{\pi_{0}\left(y_{r} \mid x\right)}-\delta)),
\end{equation}
where $\delta$ represents the reward shift, which is introduced by \cite{jung2024bco}, calculated as the moving average of previous rewards to stabilize training.

The generation loss $\mathcal{L}_g$ is the standard cross-entropy loss:
$$
\mathcal{L}_g = -\sum_{t=1}^{T} \log p_\theta(y_t \mid x, y_{<t}),
$$
where $p_\theta$ represents the conditional probability distribution of the language model over tokens.

\paragraph{GRPO}
In terms of training thinking model, subsequent to the SFT phase, our research transitions to on-policy RL training.
Initially, a dataset comprising approximately 200k high-quality data pairs is collected from different sources, which will be detailed in Sec.~\ref{sec:post-train_data}. A difficulty score is subsequently assigned to each data sample, serving as a metric derived from the number of correct responses elicited across eight successive rollouts of the SFT version of AndesVL. Empirical observations indicate that data samples exhibiting either extreme difficulty or excessive simplicity do not meaningfully contribute to learning gains following reinforcement training. Consequently, we strategically select a subset of data with difficulty scores ranging from 1 to 4 for our training regimen. This refinement yields a final training dataset of 43.6K examples, including 15.3K pure text samples and 28.3K multimodal data instances. 

Recent work on ReVisual-R1 \cite{chen2025advancing} has demonstrated that subsequent text-only RL training, following a multimodal RL phase, further enhances multimodal reasoning capabilities. Our experiments with AndesVL similarly reveal that this two-stage RL training paradigm significantly improves the model's multimodal reasoning. Furthermore, we observed that training the model with RL in an ``easy-to-hard'' manner more effectively enhances model performance; thus, training samples are ordered according to their difficulty scores. Consequently, AndesVL also undergoes a two-stage training process, incorporating this easy-to-hard curriculum, utilizing the GRPO~\cite{shao2024deepseekmath} algorithm. The empirical results unequivocally showcase a notable enhancement in AndesVL's domain-specific reasoning capabilities.




\subsection{Training Data}
\subsubsection{Pre-train Data}
\label{sec:pre-train_data}
In this section, we present in detail the data we utilized during the pre-training stage, including several open-source datasets and our in-house data, as illustrated in Table~\ref{tab:pretain_data}.

\begin{table}[tb]
    \belowrulesep=0pt
    \aboverulesep=0pt
        \centering
        \begin{tabular}{|c|c|}
            \toprule
            Task Type & Dataset Name \\
            \midrule
            Caption & \thead{Emu2~\cite{sun2024generative}, ShareGPT-4V~\cite{chen2023sharegpt4v}, Laion-ZH~\cite{schuhmann2022laion5b}, \\ Wukong~\cite{gu2022wukong}, Taisu~\cite{liu2022taisu}} \\
            \hline
            OCR & \thead{DocMatrix~\cite{2024docmatrix}, DocStruct~\cite{hu2024mplug,hu2024mplug2}, Leopard-Instruct~\cite{jia2024leopard}, \\Pixmo-doc~\cite{deitke2024molmo}, Anyword-3M~\cite{tuo2023anytext}, PIN-14M~\cite{wang2024pin}, \\In-house collected and synthesized (by Synthdog~\cite{wang2024pin}) OCR data} \\
            \hline
            Visual Grounding & \thead{Visual Genome~\cite{krishna2017vg}, RefCOCOs~\cite{yu2016refcoco, lin2014microsoft}, LVIS~\cite{plummer2015flickr30k},\\Flickr30k-Entities~\cite{plummer2015flickr30k}, Groma~\cite{ma2024groma}}  \\
            \hline
            GUI and Agent & \thead{AITW~\cite{rawles2024androidinthewild}, AITZ~\cite{zhang2024android}, AMEX~\cite{chai2024amex}, Android Control~\cite{lieffects}, Widget Caption~\cite{li2020widget}, \\Rico~\cite{deka2017rico}, SeeClick~\cite{cheng2024seeclick}, UIbert~\cite{bai2021uibert},  Screen2Words~\cite{wang2021screen2words}, MultiUI~\cite{liu2024harnessingwebpageuistextrich}, \\Aria-UI\cite{yang2025ariauivisualgroundinggui}, OS-Atlas~\cite{wu2024atlas}, Mind2Web~\cite{deng2024mind2web}, GUI-Odyssey~\cite{lu2024gui}, OmniAct~\cite{kapoor2025omniact},\\
            In-house AndesUI training set}  \\
            \hline
            VQA & \thead{Infinity-MM~\cite{gu2024aquilavl}, MAmmoTH~\cite{guo2024mammoth}, LLaVA-OneVision~\cite{li2024llavaov}, \\The~ Cauldron~\cite{Cauldron}, 
            VisualWebInstruct~\cite{visualwebinstruct},
            PangeaInstruct~\cite{yue2024pangeafullyopenmultilingual}}\\
            \hline
            Long CoT & \thead{OpenMathReasoning~\cite{moshkov2025aimo2}, OpenCodeReasoning~\cite{ahmad2025opencodereasoning}, OpenThoughts~\cite{guha2025openthoughtsdatarecipesreasoning}, Nemotron~\cite{NemotronPostTrainingDatasetV1}, \\In-house multiModal long CoT data}\\
            \hline
            Interleaved Image-Text & \thead{MMC4~\cite{zhu2023multimodal}, MINT~\cite{awadalla2024mint}, Multimodal Textbook~\cite{zhang20252}, \\Wanjuan~\cite{he2023wanjuan}, OmniCorpus~\cite{li2024omnicorpus},
            \\In-house crawled data from Chinese websites and Apps} \\
            \hline
            Pure Text & \thead{Fineweb-Edu-Chinese~\cite{yu2025opencsgchinesecorpusseries}, Fineweb-Edu~\cite{lozhkov2024fineweb-edu}, 
            FineMath~\cite{allal2025smollm2smolgoesbig},
            OpenCoder~\cite{Huang2024OpenCoderTO}, \\Infinity-Instruct~\cite{InfinityInstruct2024}}  \\
            \bottomrule
        \end{tabular}
        \caption{The detailed lists of pre-training datasets.}
        \label{tab:pretain_data}
\end{table}

\paragraph{Image Caption}
Our image caption data comprises both Chinese and English languages. The Chinese image caption data primarily originates from Laion~\cite{schuhmann2022laion5b}, Wukong~\cite{gu2022wukong}, and Tasisu~\cite{liu2022taisu} datasets. To refine the quality of these descriptions, we utilized Qwen2.5-VL-72B~\cite{bai2025qwen2_5} to generate re-captioned versions. During training, we randomly employed the original captions with a 50\% probability and the re-captioned versions also with a 50\% probability, culminating in a dataset of around 116 million samples. The English image caption data are primarily derived from the Infinity-MM~\cite{InfinityInstruct2024} stage 1 subset, using Emu2~\cite{sun2024generative} for caption generation and consisting of approximately 10 million samples.

\paragraph{OCR}
OCR data plays a critical role in bridging visual and textual modalities within vision-language models (VLMs). Our OCR dataset is primarily derived from three sources: open-source data, synthetic data, and in-house collected data. For real-world textual images, we curated and refined widely used open-source datasets through our dedicated data engine. To further enhance data diversity, we also extracted text-rich images from the LAION-2B~\cite{schuhmann2022laion5b} dataset using PaddleOCR~\cite{li2022paddleocr}.

Synthetic data serves as another essential component in strengthening the model's OCR capabilities. Specifically, to improve recognition accuracy for Chinese characters, we generated a large-scale Chinese OCR dataset using SynthDog~\cite{kim2022synthdog}. Additionally, we produced substantial amounts of non-semantic English OCR data to help mitigate the model’s tendency toward hallucination. To further improve robustness, we applied extensive data augmentation techniques, including geometric transformations, noise injection, and style variations, ensuring the model generalizes effectively across diverse and challenging real-world scenarios.

\paragraph{Visual Grounding}
We followed the bounding box structure utilized in Qwen2-VL~\cite{wang2024qwen2vl} and prepared data for both single and multiple grounding scenarios. The grounding datasets were chosen from publicly available sources like Visual Genome~\cite{krishna2017vg}, RefCOCOs~\cite{yu2016refcoco, lin2014microsoft}, Flickr30k-Entities~\cite{plummer2015flickr30k}, and Groma~\cite{ma2024groma}. These datasets were screened and categorized into four classifications: object referring, region captioning, referenced entity recognition, and grounded image captioning. Inspired by Ferret~\cite{li2024ferret} and Ferret-v2~\cite{ferretv2}, we ensured an equitable distribution of our grounding data across the Region-in-Text-out and Text-in-Region-Out formats.

\paragraph{GUI and Agent}
We divided the GUI data into four categories, which are detailed caption, recognition, action, and element grounding. In the pre-training stage, the data were formatted in single-turn style. For the element grounding data, we kept the structure the same as our visual grounding data. During the data synthesis and reconstruction stages, we kept the balance between different task categories and platform domains.

Besides the publicly available GUI data, we built an in-house GUI dataset using ColorOS UI and application widgets, namely AndesUI. We gathered 90 apps in total, including 65 popular download apps from the OPPO Software Store, spanning a variety of categories frequently used by users, along with 25 ColorOS pre-installed apps. Annotators were directed to capture screenshots of different heterogeneous pages within each app, ensuring that each screenshot contained unique layouts and content. Ultimately, we collected a total of 10k screenshots from third-party apps and 2.5k from ColorOS pre-installed apps.
Then, we aimed to annotate all the widgets within each screenshot. This included defining bounding boxes, identifying widget types, recording any available text on the widgets, and determining their clickability, among other details. On average, each interface produced 18 widgets. The training dataset resulted in a cumulative total of 227k widgets.
Finally, we needed to construct both basic and advanced data. Basic data consists of grounding and referring datasets, whereas advanced data comprises overall descriptive data and natural question-answer pairs. 
As a result, the training set produced 227k referring data entries, 186k grounding data entries, 13k comprehensive description data, and 108k natural question-answer pairs. All the details of the AndesUI dataset are presented in Appendix~\ref{app:andesui_dataset}.

\paragraph{VQA}
Our VQA dataset primarily originated from the open-source community, encompassing general VQA datasets, Infinity-MM~\cite{gu2024aquilavl}, Llava One Vision~\cite{li2024llavaov}, and The Cauldron~\cite{Cauldron}. Additionally, it included reasoning datasets such as MAmmoTH~\cite{guo2024mammoth} and VisualWebInstruct~\cite{visualwebinstruct}, along with multilingual and multicultural datasets like PangeaInstruct~\cite{yue2024pangeafullyopenmultilingual}.

\paragraph{Interleaved Image-Text}
Interleaved image-text data serves as a natural extension of pure text pretraining data into scenarios encompassing image inputs. Unlike instruction data differentiating between single-image and multi-image contexts, interleaved image-text data is inherently multi-modal and simplifies to pure text corpora when there are no image inputs. This is similar to pretraining on purely textual data, which enables models to develop in-context learning abilities. We collected interleaved image-text data from the open-source community, which includes MMC4~\cite{zhu2023multimodal}, MINT~\cite{awadalla2024mint}, Multimodal Textbook~\cite{zhang20252}, Wanjuan~\cite{he2023wanjuan}, and OmniCorpus~\cite{li2024omnicorpus}. Furthermore, to improve the model's understanding of Chinese language and culture, we created an in-house interleaved image-text dataset based on Chinese web content.

\paragraph{Pure-text}
Pure text data plays a crucial role in maintaining the text capabilities of MLLMs. Our openly accessible pure text corpus comprises the Chinese FineWeb-Edu~\cite{yu2025opencsgchinesecorpusseries}, the English FineWeb-Edu~\cite{lozhkov2024fineweb-edu}, the mathematical corpus FineMath~\cite{allal2025smollm2smolgoesbig}, the code corpus derived from OpenCoder~\cite{Huang2024OpenCoderTO} annealing data. Besides, we also constructed a large quantity of in-house text pre-training corpora. Furthermore, we incorporated text instruction data from Infinity-Instruct~\cite{InfinityInstruct2024}.

\begin{figure}[tbh]
    \centering
    \includegraphics[width=1\linewidth]{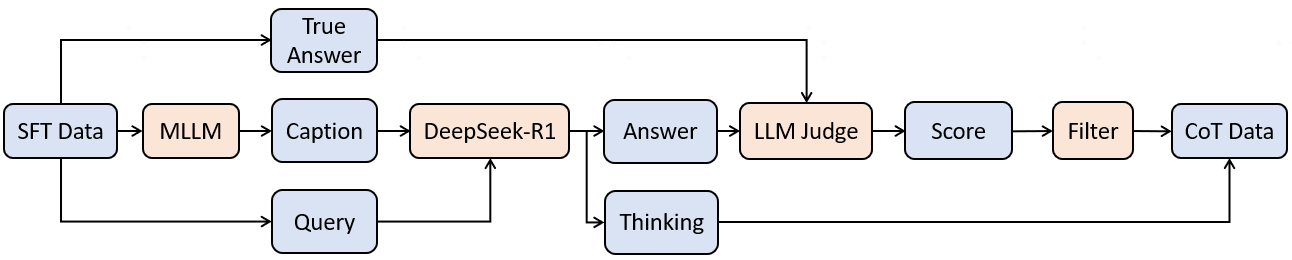}
    \caption{CoT data construction pipeline}
    \label{fig:CoT_data_construction_pipeline}
\end{figure}

\paragraph{Long COT data}
The long CoT data construction pipeline is illustrated in Fig.~\ref{fig:CoT_data_construction_pipeline}. Our long CoT dataset was constructed from two distinct pipelines: one leveraged human annotations in combination with a DeepSeek-based Chain-of-Thought (CoT) data generation pipeline, while the other relied on distilling knowledge from existing CoT models.
In the first pipeline, we began by collecting a diverse set of STEM (science, technology, engineering, and math) samples and common real-life images to serve as our visual inputs. Subsequently, human annotators developed pertinent questions based on these images and derived corresponding correct answers. Concurrently, MLLMs, such as GPT-4o~\cite{chatgpt4o}, were utilized to produce initial image descriptions. These descriptions were then manually refined, and query-related information was extracted to maintain alignment and relevance. 

In the second pipeline, which focuses on distillation from existing CoT models, we employed a hybrid strategy that merges MLLMs with DeepSeek-R1~\cite{guo2025deepseek}. Specifically, MLLMs are used to generate descriptive captions for the input images. These captions, along with the associated queries, are input into DeepSeek, which generates detailed reasoning chains as output responses, thereby producing high-quality CoT examples.

\subsubsection{Post-train Data}
\label{sec:post-train_data}
AndesVL undergoes two distinct post-training phases: SFT leveraging instruction data in a specific format and MPO utilizing sample \& reject data pairs for AndesVL-Instruct or GRPO data for AndesVL-Thinking.

\paragraph{SFT Data}
The SFT phase enhances the model's conversational capabilities and instruction-following abilities by using a large-scale, diverse dataset of instruction data obtained from publicly available resources and meticulously curated by closed-source strong models.

\begin{table}[tb]
\belowrulesep=0pt
\aboverulesep=0pt
    \centering
    \begin{tabular}{|c|c|}
        \toprule
        Task Type & Dataset Name \\
        \midrule
        VQA & \thead{EATEN \cite{guo2019eaten}, PMC-VQA \cite{zhang2023pmcvqa}, OmniAlign-V \cite{zhao2025omnialign}, Dvqa \cite{kafle2018dvqa},  \\
        mm-localized-narratives \cite{PontTuset_eccv2020},
        OCR-VQA\cite{mishra2019ocrvqa},
        Plotqa\cite{Methani_2020_WACV},
        ShareGPT-4o \cite{OpenGVLab_ShareGPT_4o},   \\
        geoqa-plus\cite{chen2021geoqa},
        Figureqa \cite{kahou2017figureqa}, Tallyqa \cite{acharya2019tallyqa}, 
        LACR-I2I \cite{nimapourjafar_LACR_I2I},
        Robut-wikisql \cite{nimapourjafar_robut_wikisql}, \\
        Mimic-cgd \cite{laurenccon2024matters},
        clevr \cite{johnson2017clevr}, textvqa \cite{singh2019towards}, scienceqa \cite{lu2022learn},
        mpdocvqa \cite{tito2022hierarchical},\\
        nlvr2 \cite{suhr2019nlvr2}, 
        ShareGPT4V \cite{chen2024sharegpt4v}, cocoqa \cite{du2019cocoqa}, ScreenQa \cite{hsiao2022screenqa}, raven \cite{zhang2019raven}, \\
        unigeo \cite{chen2022unigeo}, 
        docmatix \cite{laurenccon2024building},
        robut-wtq \cite{zhao2023robut},
        mapqa \cite{chang2022mapqa}, 
        iconqa \cite{lu2021iconqa}, chart2text \cite{obeid2020chart},   \\
        docreason25k \cite{hu2024mplug},
        kvqa \cite{kim2019korean},
        scitsr \cite{chi2019complicated},
        mm-LADD \cite{nimapourjafar_mm_LADD}, tabmwp \cite{lu2022dynamic},\\
        KonIQ-10k \cite{hosu2020koniq},  
        SVIT \cite{zhao2023svit},
        sujet-finance-qa-vision \cite{Sujet-Finance-QA-Vision-100k}, \\
        viet-sharegpt-4o-text-vqa \cite{doan2024vintern1befficientmultimodallarge}, 
        ChartQA \cite{liu2024chatqa}, cambrian-10m \cite{tong2024cambrian}, mm-aokvqa \cite{schwenk2022okvqa},  \\
        ai2diagram \cite{kembhavi2016diagram}, Clevr-CoGenT-TrainA-70K-Complex \cite{johnson2017clevr}, 
        laion-gpt4v \cite{laion_laion_gpt4v}, \\
        geometry3k \cite{lu2021inter},
        DocVQA \cite{mathew2021docvqa},
        vistext \cite{tang2023vistext}, simchart9k \cite{xia2023structchart}, \\
        geomverse \cite{kazemi2023geomverse}, spot-the-diff \cite{tan2013spot},
        robut-sqa \cite{zhao2023robut}, 
        multihiertt \cite{zhao2022multihiertt}, \\
        lrv-instruction \cite{liu2023aligning}, objects365 \cite{shao2019objects365},
        mtvqa-train \cite{tang2024mtvqa},
        iam \cite{cheng2022iam},
        finqa \cite{chen2021finqa},  \\
        viquae \cite{lerner2022viquae}, fsc147-controlnet \cite{Theonewhomadethings_fsc147_controlnet},
        mm-tat-qa \cite{zhu2021tat},
        infographic-vqa \cite{gonzalez2024metrics},  \\
        vsr \cite{zhang2021vsr},
        orand-car-a \cite{Labeled_ORAND_CAR_A},  mm-tqa \cite{nimapourjafar_mm_tqa},
        mm-intergps \cite{nimapourjafar_mm_intergps}, \\
        JourneyBench-Hallucination \cite{JourneyBench_Hallucination}, mm-vqarad \cite{nimapourjafar_mm_vqarad}, 
        mm-diagram-image-to-text \cite{nimapourjafar_diagram_image_to_text}, \\
        ctw\cite{liu2019curved}, naf\cite{davis2019deep},
        LIME-DATA-ai2d-train \cite{LIME_DATA_ai2d_train},
        mmc-inst \cite{MMC_Instructed_Dataset}, COCO-Text \cite{howard_hou_COCO_Text},\\
        HME100K \cite{VLM_Perception_HME100k_400}, 
        st-vqa \cite{vikhyatk_st_vqa}, fintabnet \cite{apoidea_fintabnet} 
        , CoSyn-400K \cite{deitke2024molmo}, PuzzleVQA \cite{chia2024puzzlevqa} } \\
        \hline
        OCR & \thead{anytext \cite{tuo2023anytext}, CORD \cite{park2019cord}, invoice-to-json \cite{Invoice-to-Json},
        arxiv-ocr \cite{nz_arxiv_ocr}, textocr \cite{MiXaiLL76_TextOCR_OCR},  \\
        invoices-and-receipts-ocr-v2 \cite{mychen76_invoices_receipts_ocr_v2},
        mall-receipt-extraction \cite{CC1984_mall_receipt_extraction_dataset}, \\
        invoices-and-receipts-ocr-v1 \cite{mychen76_invoices_receipts_ocr_v1}, 
        ds-receipts-v2-train \cite{ds_receipts_v2_train}, dataset-receipt \cite{ilhamxx_dataset_receipt}, \\
        invoices-donut-data-v1 \cite{katanaml_invoices_donut_data_v1}, 
        Vision-OCR-Financial-Reports \cite{Vision_OCR_Financial_Reports_10k},\\
        handwritten-text-ocr \cite{toghrultahirov_handwritten_text_ocr}, 
        nutritional-data-poie \cite{kashindra_mahato_nutritional_data_poie} } \\
        \hline
        Pure Text Dialogue & \thead{sharegpt-gpt4 \cite{shibing624_sharegpt_gpt4}, 
        ruozhiba \cite{LooksJuicy_ruozhiba}, 
        Ner-sentiment-analysis-sharegpt \cite{Yeenyi_ner_sentiment_analysis_sharegpt},  \\
        chinese-ner-sft \cite{qgyd2021_chinese_ner_sft},
        few-shot-ner-sft \cite{qgyd2021_few_shot_ner_sft}, 
        SystemChat \cite{cognitivecomputations_SystemChat}
        }
        \\
        \hline
        Image Captioning & \thead{Detailed Caption \cite{li2023monkey}, VizWiz \cite{gurari2018vizwiz} }\\
        \hline
        Code Generation & \thead{WebSight \cite{laurencon2024unlocking}, mm-datikz \cite{nimapourjafar_mm_datikz}} \\
        \hline
        Function Calling & \thead{Function-Calling-Dataset-V1 \cite{Hermes-Function-Calling-Dataset-V1}, glaive-function-calling \cite{glaiveai_glaive_function_calling}
        }
 \\
        \hline
        Markdown & Docomini\cite{hu2024mplug}, Mdocr Chinese markdown \cite{wang2024pin}  \\
        \hline 
        Math & \thead{cmm-math\cite{liu2024cmmmath}, MMR1-Math-RL-Data-v0 \cite{MMR1_Math_RL_Data_v0}, Codegebragpt-multimodal \cite{sr5434_CodegebraGPT_data},  \\
        Geo170K \cite{Luckyjhg_Geo170K},
        MathV360K \cite{shi2024math}, Multimath-300k \cite{multimath_300k}, 
        Unimer-math-ocr \cite{wang2024unimernetuniversalnetworkrealworld},  \\
        Openr1-math-220k \cite{mOpenR1_Math_220k},
        MathInstruct \cite{yue2023mammoth}, MetaMathQA \cite{yu2023metamath}}  \\
        \hline
        In-house Data & Meticulously auto-generated, labeled, and curated instruction data \\
        \bottomrule
    \end{tabular}
    \caption{The detailed lists of SFT datasets.} 
    \label{tab:sft_data}
\end{table}

\begin{figure}
    \centering
    \includegraphics[width=1\linewidth]{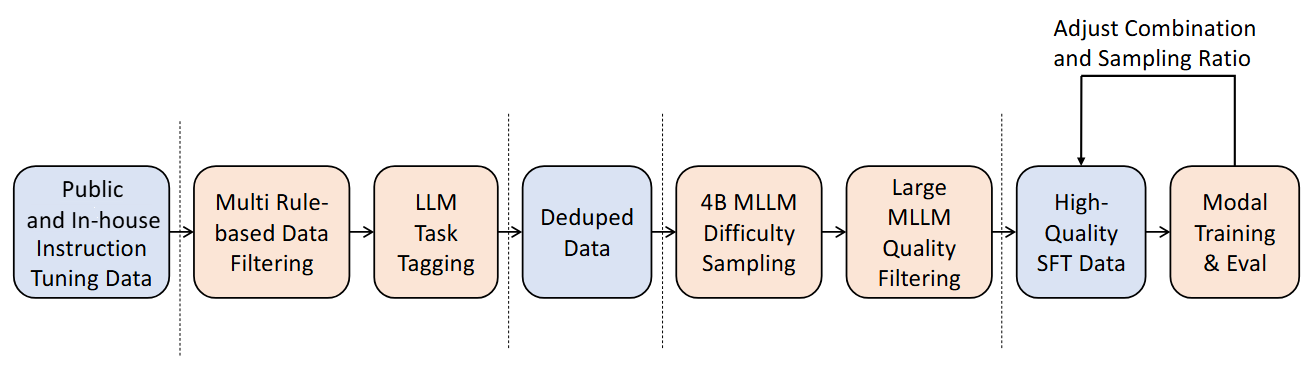}
    \caption{SFT Data Filtering Pipeline}
    \label{fig:SFT_Data_Construction_Pipeline}
\end{figure}

As illustrated in Fig.~\ref{fig:SFT_Data_Construction_Pipeline}, we adopted a multi-stage data filtering process to further enhance the quality of these datasets. Initially, we utilized traditional rule-based single-modality filtering only on text and images, eliminating basic noise (\textit{e.g.}, invalid or blurry images, incorrect instructions) and inappropriate data within each dataset. After this, we employed Qwen2.5-VL-72B~\cite{bai2025qwen2} that clusters all datasets into different task categories. Then, these clustered image-text pairs were filtered on the measurement of quality and difficulty level by the LLM-as-a-judge \cite{zheng2023vicuna} approach. We used GPT-4o~\cite{chatgpt4o} to measure quality according to factual accuracy, image-text correspondence, and hallucination levels. We employed previously trained checkpoints for difficulty filtering to generate multiple responses for the image-query pair. The image-text pair was considered unsuitable for the SFT training if most of the generated responses were judged to be the same as or above the level of the image-text pairs.

The final SFT dataset encompasses approximately 16 million entries, distributed between unimodal text data (10\%) and multimodal data, incorporating interleaved image-text sequences. This data composition, characterized by a significant proportion of multimodal data complemented by a supplementary portion of pure text, enables the model to maintain robust performance even in purely text-based scenarios.

To facilitate the acquisition of more comprehensive and superior capabilities, the instruction data covers a wide range of task types, including VQA, OCR, image captioning, pure text dialogue, code generation, function calling, markdown format generation, pure text math, and multimodal math. We list the details of the SFT datasets in Table~\ref{tab:sft_data}.

\paragraph{MPO Data}
The MPO dataset was derived from two distinct sources: one is constructed through our in-house data generation pipeline, and the other is the publicly available MMPR-v1.2 dataset~\cite{wang2024mpo} introduced by InternVL~\cite{zhu2025internvl3}.
\begin{figure}
    \centering
    \includegraphics[width=1\linewidth]{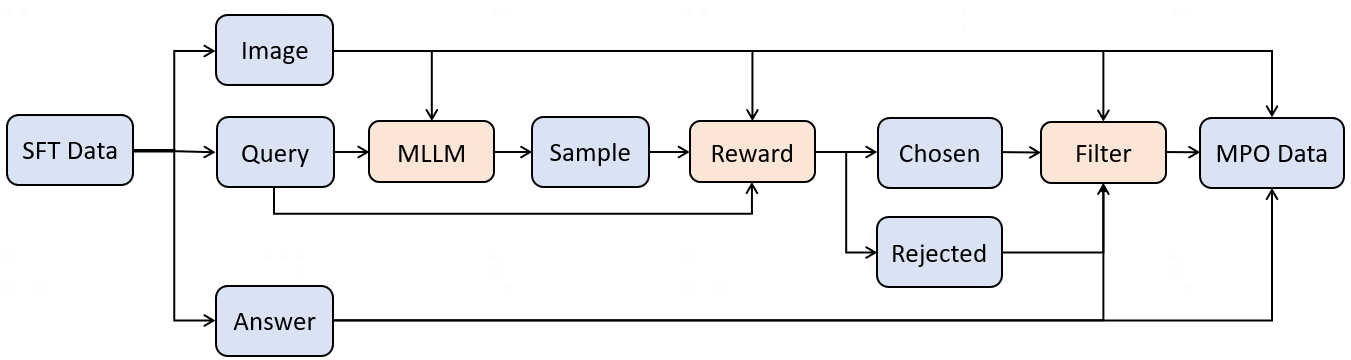}
    \caption{MPO data construction pipeline}
    \label{fig:MPO_data_construction_pipeline}
\end{figure}

The MPO data construction pipeline, as illustrated in Fig.~\ref{fig:MPO_data_construction_pipeline}, began with the collection of SFT data from specific domains to act as input sources. Following this, sampling-based inference was conducted using
AndesVL after SFT, yielding an n-best list of responses for each input. Subsequently, an MLLM, \textit{e.g.}, GPT-4o~\cite{chatgpt4o}, was leveraged to evaluate these candidate responses, selecting the most precise and coherent response as the \textit{chosen response} and the least relevant or erroneous one as the \textit{rejected response}, based on the input image, query, and reference answer.

Then a filtering strategy was applied to guarantee the quality of data. In particular, two similarity scores were calculated: (a) comparing the chosen and rejected responses and (b) comparing the rejected response with the ground-truth answer. Instances were discarded if (a) they fell below a designated threshold—indicating insufficient contrast—or if (b) they exceeded a specific level, implying that the rejected response is too similar to the correct answer, which could mislead the preference learning process. Finally, we obtained about 80k valid MPO data through this process.

\paragraph{GRPO Data}
The GRPO data primarily comprised tasks from STEM-related domains, encompassing both unimodal text and multimodal inputs. The data sources included publicly available datasets as well as annotated in-house data. Specifically, the GRPO dataset integrated the following: We-Math2.0~\cite{qiao2025wemath}, MathV360K~\cite{shi2024mathv}, KVQA~\cite{shah2019kvqa}, ChartQA~\cite{masry2022chartqa}, ThinkLite~\cite{wang2025sota}, STEM-500k~\cite{shen2024measuring}, deepscaler~\cite{deepscaler2025} and Statics-5k. Among these, Stats-5K is an in-house annotated dataset specifically designed for computational tasks involving statistical charts in English-language contexts. To enhance data quality and training efficacy, we applied post-processing procedures, including difficulty grading and content categorization.

\textbf{Difficulty Grading} Difficulty grading refers to performing \textit{n} rollouts on these data using the thinking model after SFT training and then categorizing the difficulty levels based on the number of correct answers obtained in the \textit{n} rollouts. A higher number of correct responses corresponds to a lower difficulty level. 

\textbf{Content Categorization} This involved first identifying the model’s deficient capabilities, followed by employing an LLM to select and group data instances with semantically similar content.

Through these refinement strategies, we constructed a reinforcement learning dataset predominantly composed of mathematical and STEM tasks, amounting to approximately 43.6k samples.






\section{Mobile-side Application of AndesVL}
\label{sec:mobile-side_depoloy_andesvl}

Based on the AndesVL model after both pre-training and post-training, we build a \textbf{1+N LoRA}~\cite{yang2023longqlora} on-device AI framework. This architecture comprises a foundational model and multiple scenario-specific LoRA adapters for each scenario. Based on this framework, we further perform quantization and mobile-side acceleration and release multiple on-device AI applications on OPPO AI phones.

\subsection{Multi-scenario LoRA Training}
\begin{figure}
    \centering
    \includegraphics[width=1\linewidth]{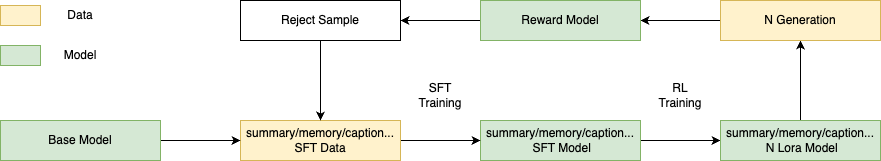}
    \caption{Multi-scenario LoRA training based on AndesVL.}
    \label{fig:Multi-scenario training}
\end{figure}

During multi-scenario deployment of AndesVL models, it is imperative to balance the generalization capacity of the model with its domain-specific adaptability. To address practical requirements during application, based on AndesVL, we further designed a dedicated multi-scenario LoRA training stage, structured as in Fig.~\ref{fig:Multi-scenario training}.

\begin{table}[tb]
\centering
\begin{tabular}{ll}
\toprule
\textbf{Task Type} & \textbf{Reward Criteria} \\
\midrule
OCR tasks & Text detection accuracy and structural integrity \\
Caption generation & Semantic relevance and linguistic conciseness \\
Text summarization & Content coverage and instruction alignment \\
\bottomrule
\end{tabular}
\caption{Reward signals for different specific real-world tasks}
\label{tab:reward_signal_for_multi_scenarios}
\end{table}

In practical deployment scenarios, task-specific fine-tuning is often required while computational resources remain limited. It is infeasible to train separate large models for each individual scenario. Therefore, based on our AndesVL described above, we trained multiple LoRA models for different scenarios while keeping the base model parameters frozen. This approach only requires fine-tuning a minimal number of parameters to adapt to various application scenarios.
It significantly reduces training resource consumption, while effectively prevents catastrophic forgetting and enhances the model's generalization across multiple scenarios. The LoRA training for each scenario consist of two phases: SFT and RL.

The SFT data construction for LoRA fine-tuning was more scenario-specific and customized, where for each scenario we collected and annotated high-quality, strongly relevant dedicated data; designed data labels and task definitions that closely align with actual requirements; and implemented customized training loss functions tailored to specific scenarios.
For example, in image caption generation tasks, to increase entity density, we designed an entity-weighted cross-entropy loss that assigns higher loss weights to entity words (\textit{e.g.}, colors, quantities, object categories) in captions, thereby encouraging the model to focus more on generating these critical pieces of information.
The entity-weighted cross-entropy loss enhances key information generation through differential weighting:
\begin{equation}
\mathcal{L}_{\text{entity}} = -\frac{1}{N} \sum_{i=1}^{N} \sum_{t=1}^{T_i} \alpha_{i,t} \log P(w_{i,t} | x_i, w_{i,<t}, \theta),
\end{equation}
where $\alpha_{i,t}$ denotes the weighting factor ($\alpha_{i,t} > 1$ for entity tokens, $=1$ otherwise), $N$ represents the batch size, $T_i$ is the sequence length of the $i$-th sample, and $w_{i,t}$ indicates the $t$-th token in the $i$-th sample.
The total training objective that combines entity-focused and fluency-preserving terms is
\begin{equation}
\mathcal{L}_{\text{total}} = \lambda_1 \mathcal{L}_{\text{entity}} + \lambda_2 \mathcal{L}_{\text{BLEU/ROUGE}},
\end{equation}
where $\lambda_1$ is the weight for entity preservation, $\lambda_2$ is the fluency control coefficient maintaining grammatical quality and $\mathcal{L}_{\text{BLEU/ROUGE}}$ denotes standard metric-based loss for text quality.

The RL phase in the multi-scenario LoRA stage is also highly customized. For example, in the captioning task, in addition to ensuring the accuracy and conciseness of the generated captions, it is also necessary to achieve a higher density of entity words (\textit{i.e.}, including useful information such as color, quantity, etc.). This enables the model to output more informative content within the same token length, thereby facilitating improvements in downstream tasks such as album search. We define the \textbf{Entity Density Reward} as
\begin{equation}
R_{\text{entity}} = \frac{\text{Number of entity words in caption}}{\text{Total number of words in caption}},
\end{equation}
the \textbf{Key Information Reward}, 
\begin{equation}
R_{\text{info}} = \beta_1 \cdot \mathbb{I}(\text{caption contains color}) + \beta_2 \cdot \mathbb{I}(\text{caption contains number}),
\end{equation}
where $\mathbb{I}(\cdot)$ is the indicator function (1 if the condition is satisfied and 0 otherwise), and $\beta_1, \beta_2$ are weighting coefficients. So finally, the \textbf{Total Reward} is
\begin{equation}
R_{\text{total}} = \lambda_1 R_{\text{entity}} + \lambda_2 R_{\text{info}} + \lambda_3 R_{\text{BLEU/ROUGE}},
\end{equation}
where $\lambda_1, \lambda_2, \lambda_3$ are weighting coefficients and $\lambda_3$ ensures the fluency and relevance of the caption. Various detailed examples of reward signals are presented in Table~\ref{tab:reward_signal_for_multi_scenarios}. This mechanism ensures consistent and high-quality outputs in diverse scenarios.

The multi-scenario LoRA training phase focuses on \textbf{deep customization} and \textbf{strong adaptation}, utilizing scenario-specific data along with customized loss and reward functions to significantly enhance model precision and practical utility in targeted application scenarios.

\subsection{Quantization and Deployment}
We have established an end-to-end quantization optimization pipeline, comprising a QAT framework for base models and a scenario-specific Quantization-Aware LoRA Fine-Tuning (QALFT) framework. This pipeline leverages cloud-based computational resources and engineering investments, to maximally preserve AndesVL performance on edge devices, while simultaneously enhance on-device inference efficiency through fine-grained mixed-precision quantization.

\subsubsection{Quantization-Aware Training for AndesVL}
\label{sec:QAT_AndesVL}
Although post-training quantization (PTQ) techniques have advanced rapidly, directly deploying models to mobile devices via PTQ still incurs significant performance degradation. Moreover, the inherent unpredictability of PTQ-induced accuracy loss imposes an additional burden on algorithm validation and testing. 

To address these challenges, we have developed a robust and flexible Quantization-Aware Training (QAT) framework. It supports multiple quantization configurations: weights can be quantized to 2, 3, 4, or 8 bits, and activations to 8 or 16 bits. The framework also enables fine-grained mixed-precision combinations and includes automated precision assignment strategies to maintain model accuracy while maximizing inference efficiency. Furthermore, through close collaboration with silicon vendors, we have established a deterministic mapping mechanism that directly translates static-QAT models into hardware-compatible, edge-deployable quantized representations. This approach aims to fundamentally eliminate the performance uncertainty on edge devices that arises from PTQ.

\subsubsection{QALFT}
QAT effectively satisfies the accuracy requirements for deploying a single base model on edge devices. However, in multi-LoRA scenarios, the activation quantization encodings of the base model must jointly account for the activation ranges introduced by all LoRA adapters. Consequently, any update to a LoRA adapter necessitates re-quantizing both the base model and all associated LoRAs to maintain optimal performance across diverse use cases—an impractical requirement for edge deployment.

To overcome this limitation, we co-developed the Quantization-Aware LoRA Fine-Tuning (QALFT) framework in collaboration with MediaTek. QALFT begins by applying PTQ to a QAT-pretrained base model and permanently freezing its quantization encodings. Subsequent LoRA weights are then trained on top of this fixed, quantized backbone—following a paradigm analogous to QLoRA~\cite{dettmers2023qloraefficientfinetuningquantized}. This design enables independent updates of scenario-specific LoRA modules without re-quantizing the base model, thereby eliminating quantization-induced performance degradation during deployment and significantly streamlining the iteration cycle for task-specific algorithms. Empirical evaluations demonstrate that, in this framework, the on-device performance of AndesVL degrades by only 3\% relative to its full-precision model, the marginal loss validates the efficacy of QALFT in real-world applications.

\begin{figure}
    \centering
    \includegraphics[width=0.8\linewidth]{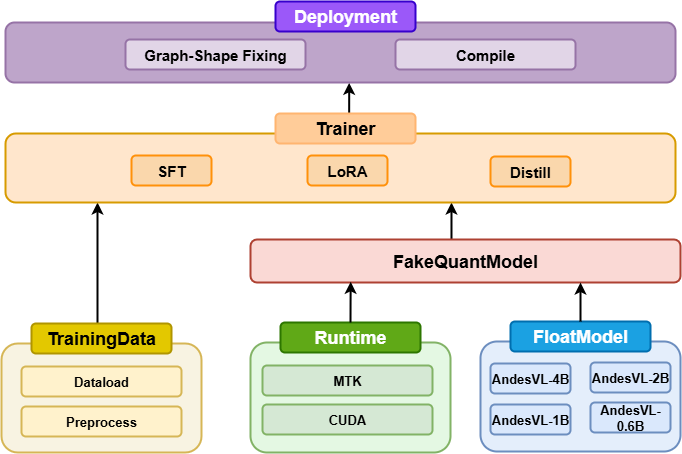}
    \caption{QALFT framework.}
    \label{fig:QALFT_framework}
\end{figure}

As illustrated in Fig.~\ref{fig:QALFT_framework}, QALFT employs a layered architectural design. Its core principle is the complete decoupling of three essential components: the floating-point base model, training data, and the QALFT trainer. This decoupling ensures that the training logic remains agnostic to and isolated from vendor-specific hardware infrastructure, thereby facilitating seamless and efficient deployment on MediaTek platforms.

\subsection{Mobile-side Acceleration with Cache Eviction}
The key-value cache (KV cache) plays a crucial role in enhancing the inference performance of LLMs. Nevertheless, as the input sequence length expands, the size of the KV cache also grows proportionally—this not only imposes significant pressure on memory resources but also undermines time efficiency. This issue is particularly pronounced for edge devices such as mobile devices: on these platforms, both performing inference on long text inputs and storing massive volumes of KV cache data are highly inefficient and impractical.

Surprisingly, the long text prompt itself is extremely sparse, which means that only a small number of tokens contribute most of the value. Therefore, we can perform an eviction operation on the KV cache.

Classic cache eviction solutions include streamingLLM~\cite{xiao2023efficient}, H2O~\cite{zhang2023h2o}, and snapKV~\cite{li2024snapkv}. The essence of these solutions lies in retaining the latest token and the previous key token based on observations, attention weights, etc. We designed a new solution called OKV that outperforms these solutions while maintaining the same compression rate and supports context lengths up to 128K.

\subsection{Mobile-Side Acceleration with Speculative Decoding}
Due to the sequential nature of auto-regressive LLMs, the decoding phase is expensive and slow. Speculative decoding has been proven to be an effective solution to this problem: EAGLE-2~\cite{li2024eagle} performs auto-regression at the feature level, reusing top-layer features from the target model in drafting to achieve better results than vanilla speculative decoding; HASS~\cite{zhang2024learning} proposes a training-time-testing method, which further improves accept length by reusing features of the draft model in the training phase to maintain consistency in the inference phase.

Based on the characteristics of mobile-side devices, we made some customizations and adaptations to existing Eagle-like methods reusing top layer features, to fully utilize the storage and computation resources on devices. We experimentally evaluate the AndesVL model with speculative decoding on multiple specific tasks. The results show that our customized method achieves a \textbf{block efficiency (BE)} of up to \textbf{7.9}. Additionally, when combined with hardware compression and LLM sparsification, it obtains a \textbf{6.7x} peak speedup ratio over the baseline.

\section{Experiments}
\label{sec:experiments}
In this section, we provide experimental results to demonstrate the comprehensive capabilities of AndesVL. The general multimodal capabilities of AndesVL are compared with those of SOTA MLLMs using widely adopted multimodal benchmarks. Following this, the domain-specific performances of AndesVL are detailed, covering text-rich image understanding (including OCR, chart, and document comprehension), reasoning and math, multi-image comprehension, general VQA, hallucination mitigation, multilingual capability, and GUI-related tasks. Most of the benchmarks are tested using VLMEvalKit~\cite{duan2024vlmevalkit}. 

\subsection{Overall Performance}
\subsubsection{Benchmarks}
We evaluated the performance of AndesVL in comparison to several advanced multimodal models of comparable model size, including Qwen2-VL~\cite{wang2024qwen2vl}, Qwen2.5-VL~\cite{bai2025qwen2_5}, InternVL3~\cite{zhu2025internvl3}, InternVL3.5~\cite{wang2025internvl3}, Gemma3~\cite{gemmateam2025gemma3}, BlueLM-2.5-3B~\cite{xiong2025bluelm}, Phi-3.5-Vision~\cite{abdin2024phi3}, Phi-4-Multimodal~\cite{abouelenin2025phi}, MiniCPM-V~\cite{yao2024minicpm}, R-4B~\cite{jiang2025r}, Qianfan-VL~\cite{dong2025qianfan}, Ovis2~\cite{lu2024ovis}, SAIL-VL-1.5-2B~\cite{dong2025scalable}, SAIL-VL2-2B~\cite{yin2025sail}, and SmolVLM2~\cite{marafioti2025smolvlm}. For fair comparisons, these models are grouped by their parameter sizes in the following evaluations.

The diverse multimodal capabilities of the proposed AndesVL are assessed using 32 commonly adopted benchmarks, covering various multimodal tasks across six domains: reasoning and math, text-rich, multi-image, general VQA, hallucination and multilingual capability. Detailed benchmark information is presented in subsequent subsections.

\subsubsection{Evaluation Results}

\begin{table}
    \centering
    \resizebox{1\textwidth}{!}{
    \begin{tabular}{l|cccccc|c}
        \hline
         Model & Text-rich & \makecell{Reasoning\\\&Math} & Multi-image & \makecell{General\\VQA}  & Hallucination & Multilingual & Overall\\
         \hline
         Phi-3.5-Vision~\cite{abdin2024phi3}  & 65.8& 24.0& 36.8& 55.4& 63.9& 47.0& 48.8\\
         Phi-4-Multimodal~\cite{abouelenin2025phi}  & 81.2& 33.2& 42.4& 64.2& 66.0& 54.3& 56.9\\
         Gemma3-4B~\cite{gemmateam2025gemma3}  & 61.0& 28.9& 38.1& 57.8& 62.1& 52.4& 50.0\\
         Qwen2.5-VL-3B~\cite{bai2025qwen2_5} & 82.1& 32.1& 44.8& 62.2& 66.9& 58.9& 57.8\\
         Ovis2-4B~\cite{lu2024ovis}  & 85.1& 34.1& 45.6& 71.3& 73.2&63.4&62.1\\
         MiniCPM-V-4-4B~\cite{yao2024minicpm} & 82.4& 33.8& 59.1& 70.9& 69.3& 55.4&61.8\\
         R-4B-RL~\cite{jiang2025r} & - & 57.1 & - & -& -&- &-\\
         InternVL3.5-4B~\cite{wang2025internvl3} & 82.6& 56.9& 62.3& 72.8& 69.6& 62.1& 67.7\\
        \rowcolor{gray!15} AndesVL-4B-Instruct & 85.3& 42.1& 64.5& 72.7& 73.0& 64.6& 67.0\\
        \rowcolor{gray!15} AndesVL-4B-Thinking & \textbf{86.0}& \textbf{58.3}& \textbf{67.8}& \textbf{73.8}& \textbf{74.8}& \textbf{64.9}& \textbf{70.9}\\
         \hline
         Qwen2-VL-2B~\cite{wang2024qwen2vl}  & 75.8& 23.1& 49.7& 60.5& 66.1& 52.8& 54.7\\
         MiniCPM-V-2-2B~\cite{yao2024minicpm}  & 60.5& 20.8& 50.5&  53.5& 63.6& 40.2& 48.2\\
         SAIL-VL-1.5-2B~\cite{dong2025scalable}  & 82.1& 29.5& 55.6& 68.4& 70.5& 57.3& 60.6\\
         SAIL-VL2-2B~\cite{yin2025sail}  & \textbf{84.1} & 33.1& 59.0& \textbf{71.8} & 71.2& \textbf{61.7} & 63.5\\
         Ovis2-2B~\cite{lu2024ovis}  & 81.8& 29.5& 59.7& 67.2& 70.3& 58.0& 61.1\\
         InternVL3-2B~\cite{zhu2025internvl3} & 78.3& 31.6& 54.3& 69.4 & 67.9& 57.4& 59.8\\
         InternVL3.5-2B~\cite{wang2025internvl3} & 79.6& \textbf{49.9}& 56.6& 68.3& 70.5& 58.0& 63.8\\
         \rowcolor{gray!15} AndesVL-2B-Instruct & 82.4& 33.8& 56.5& 66.1& 70.9& 60.3& 61.7\\
         \rowcolor{gray!15} AndesVL-2B-Thinking & 81.3& 45.7& \textbf{59.8}& 68.3& \textbf{71.8}& 59.4& \textbf{64.4}\\
         \hline
         Ovis2-1B~\cite{lu2024ovis}  & 77.3& 24.3& 53.0& 59.5& 65.4& 52.4& 55.3\\
         InternVL3-1B~\cite{zhu2025internvl3}  & 71.2& 23.8& 47.8& 61.2& 65.4& 47.9& 52.9\\
         InternVL3.5-1B~\cite{wang2025internvl3}  & 73.5& 32.8& 52.2& 59.9& 65.4& 49.1& 55.5\\
         \rowcolor{gray!15} AndesVL-1B-Instruct  & 76.7& 27.4& 52.1& 60.7& 67.0&  53.3& 56.2\\
         \rowcolor{gray!15} AndesVL-1B-Thinking  & \textbf{77.4}& \textbf{35.8}& \textbf{54.3}&\textbf{ 63.4}& \textbf{67.4}& \textbf{54.1}& \textbf{58.8}\\
         \hline
         SmolVLM2-0.5B~\cite{marafioti2025smolvlm}  &55.5& 18.4& 42.0& 43.6& 54.4& 26.1& 40.0\\
         \rowcolor{gray!15} AndesVL-0.6B-Instruct & \textbf{73.5}  & 26.0 & 51.5 & 55.3  &  65.7 & \textbf{51.0} & 53.8\\
        \rowcolor{gray!15}  AndesVL-0.6B-Thinking & 73.3& \textbf{29.4}& \textbf{53.1}& \textbf{57.1}&  \textbf{65.9}& 49.7& \textbf{54.7}\\
         \hline
    \end{tabular}}
    \caption{The overall comparison of AndesVL with existing MLLMs on 32 benchmarks, which are grouped into 6 domains. The best results are marked in \textbf{bold}.}
    \label{tab:benchmark_all}
\end{table}

Table~\ref{tab:benchmark_all} summarizes the overall performance of various existing MLLMs across 32 benchmarks spanning six different categories: Text-rich, Reasoning \& Math, Multi-image, General VQA, Hallucination, and Multilingual. We compute the average scores, drawn from the models' original papers or the OpenCompass leaderboard~\cite{opencompass2023}, to represent their capabilities across specific domains and overall.

Our proposed AndesVL series substantially outperform existing models of similar sizes on multiple test sets, across all evaluated scales. These statistics highlight the effectiveness of our advanced training strategies and the quality of the training corpus utilized.

Specifically, across 32 benchmarks, the AndesVL-4B-Thinking model achieves an overall score of 70.9, outperforming the second-best model, InternVL3.5-4B~\cite{wang2025internvl3}, by a margin of 3.2 points. Across every multimodal task category, the AndesVL-4B-Thinking model secures a significant margin of 0.9 to 5.5 points, underscoring its universal superiority in diverse multimodal scenarios. AndesVL-4B-instruct also demonstrates remarkably strong performance across multiple vertical domains, especially on multilingual and multi-image tasks. 

At the 2B scale, the AndesVL-2B-Thinking model achieves the highest overall score of 64.4. It exhibits a clear advantage in multi-image understanding and hallucination mitigation over existing models, even surpassing some 4B-scale models. 

For even more compact and lightweight models, our proposed 1B and 0.6B models command a decisive advantage across all metrics, with their Thinking and Instruct versions occupying the top spots and suppressing other leading models in the literature. Notably, our 0.6B variants, the AndesVL-0.6B series, achieve a performance even comparable to existing 1B models, such as InternVL3.5-1B. 

Above results underscore the model's proficiency in addressing a wide range of real-world tasks that require multimodal perception, understanding, knowledge, and reasoning. Moreover, the diversity in our models' sizes, combined with their strong performance, enables them suitable for deployment in a wide range of mobile scenarios, including those with highly limited computing resources.

\subsection{Reasoning and Math}
\subsubsection{Benchmarks}
To evaluate AndesVL’s multimodal reasoning and mathematical capabilities, we extensively evaluate the model on various benchmarks for mathematical reasoning as follows:

\textbf{MMMU}~\cite{yue2024mmmu}: MMMU evaluates MLLMs on college-level tasks across six disciplines, testing expert-level reasoning and advanced perception in specific fields. The accuracy results achieved from the model's direct answer on its validation set are recorded. 

\textbf{MMMU Pro}~\cite{yue2024mmmu}: MMMU Pro evaluates the multimodal understanding and reasoning capabilities of the model from a wide range of academic disciplines. It is the upgraded version of the MMMU benchmark. The overall accuracy score of the direct answer is reported.

\textbf{MathVista}~\cite{lu2023mathvista}: MathVista evaluates the mathematical reasoning ability, such as algebra, geometry, and statistics, of MLLMs with visual contexts. The accuracy scores on the testmini set are recorded.

\textbf{MathVision}~\cite{wang2024mathvision}: MathVision is made up of math problems with visual contexts. The problems are sourced from real math competitions. The results on full set of the benchmark are reported.

\textbf{MathVerse}~\cite{zhang2025mathverse}: MathVerse evaluates a model's capability of solving visual diagram-based math problems. The performance of its vision-only set is reported.

\textbf{DynaMath}~\cite{zou2024dynamath}: DynaMath consists of variant-generated questions for a seed question under various conditions. The worst-case accuracy is reported to reflect the model's reliability of MLLMs' reasoning abilities. 

\textbf{WeMath}~\cite{qiao2025wemath}: WeMath decomposes composite visual math problems into sub-problems to hierarchically assess inherent issues in MLLMs' reasoning, covering 67 knowledge concepts across 5 levels of granularity.

\textbf{LogicVista}~\cite{xiao2024logicvista}: LogicVista evaluates models across five logical reasoning tasks: spatial, deductive, inductive, numeric, and mechanical reasoning, leveraging a diverse dataset of visual multiple-choice questions. 

\subsubsection{Evaluation Results}

\begin{table}[tb]
    \resizebox{1\textwidth}{!}{
    \centering
    \begin{tabular}{l|cccccccc|c}
        \hline
         Model &  \makecell{MMMU\\(val)} & \makecell{MMMU\\Pro} & \makecell{MathVista\\(mini)} & MathVision & \makecell{MathVerse\\(vision-only)} & \makecell{DynaMath\\(worst case)} & WeMath & LogicVista & Overall\\
         \hline
         Qwen2.5-VL-3B~\cite{bai2025qwen2_5}   & 51.2 & 30.9*& 60.9*& 18.8*& 25.7*& 11.0*& 23.2*& 35.1*& 32.1\\
         BlueLM-2.5-3B~\cite{xiong2025bluelm} & 47.5 & - & 70.8 & 28.5 & - & - & - &- & -\\
         BlueLM-2.5-3B-thinking~\cite{xiong2025bluelm} & 51.3 & - & 78.4 & 47.7 & - & - & - &- & -\\
         Qianfan-VL-3B~\cite{dong2025qianfan} & 46.4 & - & - & - & - & - & - &- & -\\
         Gemma3-4B~\cite{gemmateam2025gemma3} &  47.3 & 24.5*& 46.3 & 23.1*& 23.2*& 11.0*& 23.1*& 32.7*& 28.9\\
         Phi-3.5-Vision-4B~\cite{abdin2024phi3} &44.6 & 23.6*& 43.3 & 14.9*& 19.7*& 9.8*& 11.2*& 25.1*& 24.0\\
         Phi-4-Multimodal~\cite{abouelenin2025phi} &55.1& 38.5& 62.4& 19.7*& 22.0*& 13.0*& 19.2*& 35.6*& 33.2\\
         Ovis2-4B~\cite{lu2024ovis} &49.0 & 28.0*& 69.6 & 21.1*& 39.0*& 12.6*& 18.0*& 35.1*& 34.1\\
         MiniCPM-V-4-4B~\cite{yao2024minicpm} & 51.2 & 33.4*& 66.9 & 20.7 & 22.0*& 14.2* & 30.0*& 32.0*& 33.8\\
         R-4B-RL~\cite{jiang2025r} & 68.1 & 46.5 & 78.0 & 47.8 & 64.9 & \textbf{39.5} & 52.8 & \textbf{59.1} & 57.1\\
         InternVL3.5-4B~\cite{wang2025internvl3} & 66.6 & \textbf{53.5}*& 77.1 & \textbf{54.4} & 61.7 & 35.7 & 50.1 & 56.4 & 56.9\\
         \rowcolor{gray!15} AndesVL-4B-Instruct & 58.0 & 37.6 & 73.3 & 27.1 & 34.3 & 21.2 & 33.7 & 41.6 & 40.9\\
         \rowcolor{gray!15} AndesVL-4B-Thinking & \textbf{66.9} & 51.4 & \textbf{79.4} & 51.0 & \textbf{66.9} & 35.5 & \textbf{57.4} & 57.7 & \textbf{58.3} \\
         \hline
         Qwen2-VL-2B~\cite{wang2024qwen2vl} &  42.2 & 19.9*& 48.0 & 17.3*& 16.8*& 4.0*& 11.3*& 25.5*& 23.1 \\
         MiniCPM-V-2B~\cite{yao2024minicpm} &  38.2 & 20.8*& 39.8 & 15.0*& 16.8*& 2.8*& 6.3*& 26.6*& 20.8\\
         SAIL-VL-1.5-2B~\cite{dong2025scalable} &  46.7*& 23.6*& 67.3 &  18.0*& 21.7*& 8.6*& 16.5*& 33.8*& 29.5\\
         SAIL-VL2-2B~\cite{yin2025sail} &  47.7 & 29.1*& 71.1 &  23.4& 24.5*& 10.2& 22.7& 36.2& 33.1\\
         SAIL-VL2-2B-Thinking~\cite{yin2025sail} &  - & -& 68.5 &  27.5& -& 20.2& 38.8& 47.0& -\\
        Ovis2-2B~\cite{lu2024ovis} &  45.6 & 23.8*& 64.1 & 17.6*& 30.7*& 10.0*& 10.4*& 33.6*& 29.5\\
         InternVL3-2B~\cite{zhu2025internvl3} &  43.2 & 26.9*& 57.0 & 19.5*& 21.8*& 14.6  & 22.4 & 47.7 & 31.6\\
         InternVL3.5-2B~\cite{wang2025internvl3} &  \textbf{59.0} & \textbf{42.6}*& 71.8 & \textbf{42.8} & 53.4 & \textbf{31.5}  & \textbf{48.5} & \textbf{49.4} & \textbf{49.9}\\
         \rowcolor{gray!15} AndesVL-2B-Instruct & 46.1 & 30.7 & 64.9 & 22.4 & 26.8 & 15.2 & 30.3 & 34.0 & 33.8 \\
         \rowcolor{gray!15} AndesVL-2B-Thinking & 52.1 & 37.3 & \textbf{73.3} & 35.2 & \textbf{54.8} & 27.5 & 41.1 & 44.3 & 45.7 \\
         \hline
         Ovis2-1B~\cite{lu2024ovis} &  36.1 & 20.9*& 59.4 & 16.0 & 23.9*& 2.8*& 9.6*& 26.0*& 24.3\\
         InternVL3-1B~\cite{zhu2025internvl3} & 43.4 & 20.1*& 45.8 & 18.8 & 18.7 & 5.8 & 13.4 & 29.8 & 24.5\\
         InternVL3.5-1B~\cite{wang2025internvl3} & \textbf{44.2} & 25.7*& 59.3 & \textbf{27.3} & 37.8 & \textbf{17.2} & 21.5 & 29.3 & 32.8\\
         \rowcolor{gray!15} AndesVL-1B-Instruct & 43.1 & 24.4 & 53.8 & 18.1 & 18.5 & 10.2 & 21.0 & 30.2 & 27.4 \\
         \rowcolor{gray!15} AndesVL-1B-Thinking & 44.0 & \textbf{27.9} & \textbf{66.4} & 23.5 & \textbf{45.1} & 11.6 & \textbf{33.9} & \textbf{34.0} & \textbf{35.8} \\
         \hline
         SmolVLM2-0.5B~\cite{marafioti2025smolvlm} &  34.1 & 14.7*& 37.5 & 13.2*& 14.0*& 3.2*& 7.2*& 23.5*& 18.4*\\
         \rowcolor{gray!15} AndesVL-0.6B-Instruct & 40.7 & \textbf{24.9} & 51.8 & \textbf{19.2} & 18.7 & 6.4 & 16.2 & \textbf{29.8} & 26.0 \\
         \rowcolor{gray!15} AndesVL-0.6B-Thinking & \textbf{43.3} & 24.3 & \textbf{54.9} & \textbf{19.2} & \textbf{34.0} & \textbf{7.0} & \textbf{22.8} & 29.3 & \textbf{29.4} \\
         \hline
    \end{tabular}}
    \caption{Comparison of reasoning and mathematical performance. The best results are marked in \textbf{bold}. Data marked with * are from our evaluation, while others are from their original papers or the OpenCompass leaderboard.}
    \label{tab:benchmark_math}
\end{table}

As shown in Table~\ref{tab:benchmark_math}, AndesVL-4B-Thinking achieves the highest overall score of 58.3 across various math and reasoning benchmarks among exiting models. Notably, AndesVL-4B exhibits considerable superiority over advanced models on the MathVista,  MathVerse and WeMath benchmarks. With an overall score of 45.7, the AndesVL-2B-Thinking model ranks second, performing very close to the top score of 49.9 in literature. Furthermore, the AndesVL's 1B and 0.6B Thinking models deliver dominant performance within their respective size groups, achieving top ranks not only overall but also on most individual benchmarks. 

These improvements over exiting models highlight the efficacy of our training strategy. Our approach enhances the visual-text joint reasoning ability by leveraging a large corpus of refined, long Chain-of-Thought (CoT) multimodal data in pre-training and through an intricately designed reinforcement learning process in post-training.

Collectively, these findings underscore AndesVLs' comprehensive capabilities in addressing multimodal mathematical problems, as well as reasoning challenges in scientific, engineering, and real-world contexts.

\subsection{Text-rich Image Understanding}
\label{sec:exp_text-rich}
\subsubsection{Benchmarks}
In order to evaluate the OCR, chart, and document understanding capabilities of AndesVL, we perform assessments over a variety of text-rich datasets, including the following seven benchmarks. 

\textbf{AI2D}~\cite{kembhavi2016ai2d}: AI2D consists of visual questions based on elementary school science diagrams. The results of its test set with and without mask settings are reported.

\textbf{OCRBench}~\cite{liu2023ocrbench}: OCRBench evaluates the overall OCR capabilities of MLLMs across five tasks: text recognition, scene text VQA, document VQA, key information extraction, and handwritten math expression recognition.

\textbf{ChartQA}~\cite{masry2022chartqa}: ChartQA requires a model to comprehend charts and graphs visually. The average relaxed accuracy across both human and augmented test sets in ChartQA is taken as the evaluation metric.

\textbf{TextVQA}~\cite{singh2019textvqa}: TextVQA evaluates a model's capability on visual reasoning with visual context from texts within images. The accuracy in the validation set is reported.

\textbf{DocVQA}~\cite{mathew2021docvqa}: DocVQA requires a model to read, comprehend, and retrieve texts within document images to answer related questions. Performance is reported on the test set using the ANLS text similarity metric.

\textbf{InfoVQA}~\cite{mathew2022infographicvqa}: InfoVQA consists of various complex infographics that combine text, graphics, and visual elements in creative layouts. The ANLS similarity score computed on the test set is reported. 

\textbf{SEEDBench-2-Plus}~\cite{li2024seedbench2plus}: SEEDBench-2-Plus evaluates a model's multimodal capability on text-rich visual tasks across charts, maps, and webs. The average accuracy on this dataset is reported.


\subsubsection{Evaluation Results}
\begin{table}[tb]
    \centering
    \resizebox{1\textwidth}{!}{
    \begin{tabular}{l|cccccccc|c}
        \toprule
         Model &  \makecell{AI2D\\(w M)} & \makecell{AI2D\\(w/o M)}  & \makecell{ChartQA\\(test)} & \makecell{TextVQA\\(val)} & \makecell{DocVQA\\(test)} & \makecell{InfoVQA\\(test)} & \makecell{OCR\\Bench} & \makecell{SEED\\2-Plus} & Overall\\
         \hline
         Qwen2.5-VL-3B~\cite{bai2025qwen2_5}   & 81.4& 91.3*& 84.2*& 79.2*& 93.0*& 77.0*& 82.6*& 68.2*& 82.1  \\
         BlueLM-2.5-3B~\cite{xiong2025bluelm} & 83.0& - & - & - & - & - & 82.6& - & -\\
         BlueLM-2.5-3B-thinking~\cite{xiong2025bluelm} &  82.6& - & - & - & - & - & 84.0& - & -\\
         Qianfan-VL-3B~\cite{dong2025qianfan} &  81.4& - & 81.8& 80.1& - & -  & 83.1& 67.6& - \\
         Gemma3-4B~\cite{gemmateam2025gemma3} &  70.7& 86.3*& 33.7& 57.7& 70.0*& 43.0*& 66.0& 60.7& 61.0\\
         Phi-3.5-Vision-4B~\cite{abdin2024phi3} & 77.8& 87.6*& 70.0*& 65.1*& 69.0*& 35.0*& 59.9& 62.2& 65.8\\
         Phi-4-Multimodal ~\cite{abouelenin2025phi} & 82.3& 91.7*& 81.4& 75.6& 93.2& 72.7& 84.4& 68.5& 81.2\\
         Ovis2-4B~\cite{lu2024ovis} & 85.7& 94.2*& 84.2*& \textbf{83.2}*& 94.0*& 79.0*& \textbf{91.1}&69.3& 85.1\\
         MiniCPM-V-4-4B~\cite{yao2024minicpm} & 82.9& 93.0*& 84.4& 80.8& 93.0*& 69.0*& 89.4& 67.0& 82.4\\
         R-4B-RL~\cite{jiang2025r} & \textbf{86.2}& -& -& -& 91.0& -& 83.6& -& -\\
         InternVL3.5-4B~\cite{wang2025internvl3} & 82.6& 92.3& 86.0& 77.9& 92.4& 78.0*& 82.2& 69.4& 82.6\\
         \rowcolor{gray!15} AndesVL-4B-Instruct &84.5& \textbf{94.6}& 87.8& 81.6& \textbf{96.0}& 81.0& 86.1& 70.9& 85.3\\
         \rowcolor{gray!15} AndesVL-4B-Thinking & 84.9& 94.1& \textbf{90.4}& 82.1&95.4& \textbf{81.9}& 87.0& \textbf{72.0}& \textbf{86.0}\\
         \hline
         Qwen2-VL-2B~\cite{wang2024qwen2vl} & 74.7& 84.1*& 72.5*& 79.5*& 90.0*& 65.0*& 79.7& 61.2& 75.8\\
         MiniCPM-V-2B~\cite{yao2024minicpm} &  62.9& 68.8*& 55.6& 73.2& 71.0*& 40.0*& 60.5& 51.9& 60.5\\
         SAIL-VL-1.5-2B~\cite{dong2025scalable} &  \textbf{83.7}& 92.4*& 78.4*&  82.0 & 92.0*& 72.0*& 88.5& 68.0*& 82.1\\
          SAIL-VL2-2B~\cite{yin2025sail} & 83.0& \textbf{92.8}*& 85.0* & \textbf{83.0}* & 93.1& 77.0* & \textbf{89.5} & 69.1* & \textbf{84.1}\\
        Ovis2-2B~\cite{lu2024ovis} &  82.7& 91.8*& 81.3*& 80.0*& 92.0*& 72.0*& 87.3& 67.4& 81.8\\
         InternVL3-2B~\cite{zhu2025internvl3} & 78.7& 87.4& 80.2& 77.0& 88.0*& 67.0*& 83.5& 64.6& 78.3\\
         InternVL3.5-2B~\cite{wang2025internvl3} & 78.8& 89.1& 80.7& 76.5& 89.4& 70.8& 83.6& 68.0& 79.6\\
         \rowcolor{gray!15} AndesVL-2B-Instruct & 80.1& 89.9&\textbf{87.4}& 79.9& \textbf{94.2}& \textbf{74.2}& 84.6& \textbf{68.8}& 82.4\\
         \rowcolor{gray!15} AndesVL-2B-Thinking & 77.8& 89.3& 86.6& 80.0& 93.9& 72.9& 82.9& 67.1& 81.3\\
         \hline
         Ovis2-1B~\cite{lu2024ovis} &  \textbf{76.4}& 85.3*& 74.9*& \textbf{78.4}*& 89.0*& 64.0*& \textbf{89.0}& 61.4& 77.3
\\
         InternVL3-1B~\cite{zhu2025internvl3} & 69.4& 78.3& 75.3& 74.1& 81.9& 53.7& 79.0& 58.2& 71.2
\\
         InternVL3.5-1B~\cite{wang2025internvl3} & 69.3& 81.8& 77.7& 71.5& 85.6& 60.5& 79.5& 62.3& 73.5
\\
         \rowcolor{gray!15} AndesVL-1B-Instruct & 71.5& 83.8& 80.4& 77.0& \textbf{91.5}& 65.3& 78.9& 64.8& 76.7
\\
         \rowcolor{gray!15} AndesVL-1B-Thinking & 74.4& \textbf{86.1}& \textbf{82.3}& 76.2& 91.4& \textbf{65.8}& 77.7& \textbf{65.5}& \textbf{77.4}\\
         \hline
         SmolVLM2-0.5B~\cite{marafioti2025smolvlm} &  57.3& 59.5*& 59.6& 60.3& 70.0*& 29.0*& 60.9& 47.7& 55.5
\\
         \rowcolor{gray!15} AndesVL-0.6B-Instruct & 68.4& 82.1& \textbf{78.2}& 69.7& \textbf{89.5}& \textbf{63.4}& 72.2& \textbf{64.3}& \textbf{73.5}
\\
         \rowcolor{gray!15} AndesVL-0.6B-Thinking & \textbf{68.8}& \textbf{82.9}& \textbf{78.2}& \textbf{68.9}& 88.8& 61.0& \textbf{73.8}& 64.0& 73.3\\
         \bottomrule
    \end{tabular}}
    \caption{Comparison of OCR, chart, and document understanding performance. The best results are marked in \textbf{bold}. Data marked with * are from our evaluation, while others are from their original papers or the OpenCompass leaderboard.}
    \label{tab:benchmark_text_rich}
\end{table}

Table~\ref{tab:benchmark_text_rich} shows a detailed comparison of AndesVL with several existing promising MLLMs on OCR-related benchmarks. AndesVL demonstrates superior or competitive performance to them. 

Among existing models, our AndesVL-4B-Thinking model claims the top rank with an overall score of 86.0, and it achieves the top results on four of eight benchmarks.  Meanwhile, the AndesVL-4B-Instruct model also delivers strong and comparable performance on text-rich tasks with a score of 85.3. Most notably, on ChartQA, the AndesVL-4B-Thinking model scores 90.4, exceeding the previous best, InternVL3.5-4B (86.0), by 4.4 points. A similar marked advantage is observed on DocVQA. AndesVL's success on the ChartQA and DocVQ benchmarks, featuring long-text images and complex questions, directly illustrates its ability to not only recognize long texts accurately but also apply advanced reasoning to solve challenging, contextual problems effectively.

Moreover, the advantages over existing models on text-rich tasks persist down to our smaller-scale versions. To be specific, our proposed AndesVL-2B-Instruct, AndesVL-1B-Thinking, and AndesVL-0.6B-Instruct models all achieve leading results in their respective model-size groups, with overall scores of 82.4, 77.4, and 73.5, respectively.

These outcomes demonstrate the effectiveness of our models' multimodal recognition and comprehension capabilities across a variety of text-rich tasks.

\subsection{Multi-Image Understanding}
\subsubsection{Benchmarks}

To evaluate AndesVL's capabilities in perception and understanding of multi-image relation, we conducted assessments on various multi-image benchmarks.  

\textbf{BLINK}~\cite{fu2024blink}: BLINK contains visual questions on multiple images from 14 computer vision tasks. Over half of the questions involve multiple images. The accuracy result on the validation set is reported.

\textbf{MMT-Bench}~\cite{mmtbench}: MMT-Bench consists of multimodal tasks across recognition, reasoning, and planning, with many sub-tasks requiring multi-image understanding. The accuracy metric in the validation set is reported.

\textbf{MuirBench}~\cite{wang2024muirbench}: MuirBench evaluates MLLMs' capabilities in multi-image understanding on 12 tasks and 10 types of multi-image relations. The accuracy score is reported.

\textbf{Q-Bench}~\cite{wu2024qbench}: Q-Bench assesses the abilities of MLLMs in low-level visual perception and understanding. The accuracy metric in the validation set is reported.

\subsubsection{Evaluation Results}
\begin{table}[tb]
    \centering
    \begin{tabular}{l|cccc|c}
        \toprule
         Model &  BLINK & Q-Bench1 val &  MMT val & MuirBench & Overall\\
         \hline
         Qwen2.5-VL-3B~\cite{bai2025qwen2_5}   & 49.3* & 30.3* & 61.0* & 38.7* & 44.8 \\
         Qianfan-VL-3B~\cite{dong2025qianfan} &  50.0 & 73.5 & 62.2 & - & -\\
         Gemma3-4B~\cite{gemmateam2025gemma3} &  43.9 & 14.2* & 53.9 & 40.5* & 38.1\\
         Phi-3.5-Vision-4B~\cite{abdin2024phi3} & 58.3 & 3.7* & 61.6 & 23.6* & 36.8\\
         Phi-4-Multimodal~\cite{abouelenin2025phi} & \textbf{61.3}& 10.2* & 60.4 & 37.6* & 42.4\\
         Ovis2-4B~\cite{lu2024ovis} & 53.0 & 20.9* & 65.2 & 43.4* & 45.6\\
         MiniCPM-V-4-4B~\cite{yao2024minicpm} & 54.0 & 76.5* & 59.7 & 46.1 & 59.1\\
         R-4B-RL~\cite{jiang2025r} & 56.3 & - & - & - & -\\
         InternVL3.5-4B~\cite{wang2025internvl3} & 58.1 & 73.8* & 64.3 & 53.1 & 62.3\\
         \rowcolor{gray!15} AndesVL-4B-Instruct & 58.2 & \textbf{77.8} & 66.4 & 55.5 & 64.5\\
         \rowcolor{gray!15} AndesVL-4B-Thinking & 58.4& 77.5 & \textbf{66.5} & \textbf{68.8} & \textbf{67.8}\\
         \hline
         Qwen2-VL-2B~\cite{wang2024qwen2vl} &  45.2 & 72.8* & 55.0 & 25.9* & 49.7\\
         MiniCPM-V-2B~\cite{yao2024minicpm} &  41.2 & 67.0* & 53.5 & 40.1* & 50.5 \\
         SAIL-VL-1.5-2B~\cite{dong2025scalable} &  45.4* & 75.9* & \textbf{61.7}* &  39.5* & 55.6\\
         SAIL-VL2-2B~\cite{yin2025sail} &  54.5* & \textbf{77.1}* & 60.9* &  43.6* & 59.0\\
        Ovis2-2B~\cite{lu2024ovis} & \textbf{65.7} & \textbf{76.2}* & 55.0 & 41.9* & 59.7\\
         InternVL3-2B~\cite{zhu2025internvl3} & 50.3 & 68.4* & 59.5 & 38.8* & 54.3 \\
         InternVL3.5-2B~\cite{wang2025internvl3} & 51.3 & 72.4* & 58.5 & 44.0 & 56.6\\
         \rowcolor{gray!15} AndesVL-2B-Instruct &  48.1 & 73.8 & 58.8 & 45.5 & 56.5\\
         \rowcolor{gray!15} AndesVL-2B-Thinking &  48.6 & 74.6 & 58.5 & \textbf{57.4} & \textbf{59.8}\\
         \hline
         Ovis2-1B~\cite{lu2024ovis} & 44.0 & 71.3 & 54.7* & 42.0* & 53.0\\
         InternVL3-1B~\cite{zhu2025internvl3} & 42.9 & 63.4* & 53.5* & 31.2 & 47.8\\
         InternVL3.5-1B~\cite{wang2025internvl3} & 44.0 & 68.5* & 54.5 & 41.7 & 52.2\\
         \rowcolor{gray!15} AndesVL-1B-Instruct & \textbf{44.7} & 70.4 & 55.2 & 38.0 & 52.1\\
         \rowcolor{gray!15} AndesVL-1B-Thinking & \textbf{44.7} & \textbf{72.4} & \textbf{57.0} & \textbf{43.2} & \textbf{54.3}\\
         \hline
         SmolVLM2-0.5B~\cite{marafioti2025smolvlm} &  40.7 & 56.5  & 44.7 & 26.2* & 42.0\\
         \rowcolor{gray!15} AndesVL-0.6B-Instruct & \textbf{46.6} & 69.2 & 52.0 & 38.0 & 51.5 \\
         \rowcolor{gray!15} AndesVL-0.6B-Thinking & 46.0 & \textbf{71.7} & \textbf{52.7} & \textbf{42.0} & \textbf{53.1}\\
         \bottomrule
    \end{tabular}
    \caption{Comparison of multi-image understanding performance. The best results are marked in \textbf{bold}. Data marked with * are from our evaluation, while others are from their original papers or OpenCompass leaderboard.}
    \label{tab:benchmark_multi-image}
\end{table}

The detailed results presented in Table~\ref{tab:benchmark_multi-image} indicate that AndesVL-4B-Thinking achieves superior outcomes across various multi-image benchmarks, culminating in a top overall score of 67.8, outperforming the previous best (InternVL3.5-4B, 62.3) by a margin of 5.5 points. It also scores the top on three out of four individual multi-image benchmarks. Moreover, as the model scale decreases, the models persist to demonstrate highly competitive accuracy, with the 0.6B variant attaining a score of 53.1. 

This superiority suggests that the advanced pre-training strategies and enhanced training datasets employed in AndesVL significantly enhance its ability to capture and reason about inter-image relationships by concurrently understanding and analyzing the relationships among multiple images.

\subsection{General VQA}
\subsubsection{Benchmarks}

We evaluate AndesVL’s general visual question-answer ability through a range of benchmarks, including real-world understanding and comprehensive benchmarks. These evaluations test the model's capabilities under complex realistic tasks and various comprehensive tasks. The following six benchmarks are included in these evaluations.

\textbf{MME} ~\cite{fu2023mme}: MME evaluates models’ perception and cognitive abilities across 14 sub-tasks. The overall summarization score across all tasks is reported. Notice that the MME score is divided by 28 to calcute the overall average score.

\textbf{MMBench v1.1}~\cite{liu2023mmbench}: MMBench v1.1 evaluates the multimodal understanding capability of MLLMs. It consists of multimodal questions over 20 dimensions and supports English and Chinese versions. The average performance scores on both test sets are reported. 

\textbf{MMVet~\cite{yu2023mmvet}}: MMVet evaluates six core competencies for MLLMs: recognition, knowledge, spatial awareness, language generation, OCR, and mathematics, across 16 integrated tasks. 

\textbf{MMStar}~\cite{chen2024mmstar}: MMStar evaluates the multimodal capabilities of MLLMs, focusing on advanced perception, reasoning, math, and science\&technology for visual and language understanding.

\textbf{RealWorldQA}~\cite{realworldqa}: RealWorldQA evaluates the spatial understanding capabilities of MLLMs under various real-world scenarios.

\textbf{R-Bench}~\cite{li2024r}: R-Bench focuses on evaluating the robustness of MLLMs to distortion in the real world, which covers 33 distortion dimensions. The accuracy on the distortion set is reported.

\subsubsection{Evaluation Results}
\begin{table}
    \resizebox{1\textwidth}{!}{
    \centering
    \begin{tabular}{l|cccccc|c}
        \toprule
         Model &  MME\_sum & \makecell{MMBench\\v1.1} & MMVet & MMStar & RealWorldQA & \makecell{R-Bench\\(dis)} & Overall \\
         \hline
         Qwen2.5-VL-3B~\cite{bai2025qwen2_5}   & 2,181*& 51.2 & 60.0 & 56.3 & 66.3*& 61.8*& 62.2 \\
         BlueLM-2.5-3B~\cite{xiong2025bluelm} & - & 82.1 & 66.7 & 64.5 & - & - & - \\
         BlueLM-2.5-3B-thinking~\cite{xiong2025bluelm} & - & 78.3 & 65.1 & 66.3 & - & - & -  \\
         Qianfan-VL-3B~\cite{dong2025qianfan} & - & - & 48.2 & 57.9 & 65.8 & - & -  \\
         Gemma3-4B~\cite{gemmateam2025gemma3} &  1,744 & 66.4 & 57.8 & 47.9 & 55.6 & 56.6*& 57.8 \\
         Phi-3.5-Vision-4B~\cite{abdin2024phi3} &1,838 & 67.4 & 43.2 & 47.5 & 53.6 & 55.4*& 55.4 \\
         Phi-4-Multimodal~\cite{abouelenin2025phi} &1,962& 77.2& 51.9& 58.9& 64.1& 62.8*& 64.2\\
         Ovis2-4B~\cite{lu2024ovis} & 2,162 & 81.4 & 65.5 & 61.9 & 71.1*& 70.5*& 71.3 \\
         MiniCPM-V-4-4B~\cite{yao2024minicpm} & 2,298& 79.7& 68.0& 62.8& 68.5& 64.7*& 71.0\\
         R-4B-RL~\cite{jiang2025r} & - & \textbf{84.8} & 73.1 & \textbf{81.9} & 69.1 & - & - \\
         InternVL3.5-4B~\cite{wang2025internvl3} & 2,272 & 79.3*& \textbf{76.6} & 65.0 & 66.3 & 68.7 & 72.8 \\
         \rowcolor{gray!15} AndesVL-4B-Instruct & 2,345& 81.2& 61.2& 66.1& 72.2 & \textbf{71.7}& 72.7\\
         \rowcolor{gray!15} AndesVL-4B-Thinking & \textbf{2,412} & 81.7 & 61.9 & 69.9 & \textbf{73.2} & 69.9 & \textbf{73.8} \\
         \hline
         Qwen2-VL-2B~\cite{wang2024qwen2vl} &  1,899 & 72.2*& 51.5 & 47.5 &  60.7 &  62.8*& 60.5 \\
         MiniCPM-V-2B~\cite{yao2024minicpm} &  1,808 & 65.8*& 41.0 & 39.1 & 55.8& 64.7*& 53.5 \\
         SAIL-VL-1.5-2B~\cite{dong2025scalable} & 2,063 & 78.5* & 61.4 &  62.8 & 67.1 & 66.7*& 68.4 \\
        SAIL-VL2-2B~\cite{yin2025sail} & 2,144 & \textbf{80.3}* & 68.7 & \textbf{64.1} & \textbf{72.0}* & 69.1* & \textbf{71.8} \\
        Ovis2-2B~\cite{lu2024ovis} &  2,005 & 77.0*& 67.9 & 56.7 & 66.0 & 64.2*&  67.2 \\
         InternVL3-2B~\cite{zhu2025internvl3} & 2,221 & 78.0*& 62.2 & 60.7 & 64.3 & \textbf{71.4}*& 69.4 \\
         InternVL3.5-2B~\cite{wang2025internvl3} & 2,123 & 75.3*& \textbf{71.7} & 62.7 & 62.0 & 62.4 & 68.3 \\
         \rowcolor{gray!15} AndesVL-2B-Instruct & 2,081 & 77.3 & 52.0 & 60.0 & 67.8& 65.3 & 66.1 \\
         \rowcolor{gray!15} AndesVL-2B-Thinking & \textbf{2,326} & 75.4 & 59.5 & 62.7 & 64.8 & 64.2 & 68.3 \\
         \hline
         Ovis2-1B~\cite{lu2024ovis} &  1,720 & 68.4*& 50.0*& 52.1 & 63.9 & 61.0*& 59.5 \\
         InternVL3-1B~\cite{zhu2025internvl3} & 1,935 & 68.2*& \textbf{59.5}*& 51.5 & 58.2 & 60.4 & 61.2\\
         InternVL3.5-1B~\cite{wang2025internvl3} & 1,910 & 67.6*& 56.5*& 51.9 & 57.6 & 57.4 & 59.9 \\
         \rowcolor{gray!15} AndesVL-1B-Instruct & 1,938 & 70.9& 43.5 & 52.5 & 65.0 & 63.4 & 60.7 \\
         \rowcolor{gray!15} AndesVL-1B-Thinking & \textbf{2,006} & \textbf{73.1} & 48.9 & \textbf{57.9} & \textbf{65.1} & \textbf{64.0} & \textbf{63.4} \\
         \hline
         SmolVLM2-0.5B~\cite{marafioti2025smolvlm} & 1,448 & 41.6*& 29.9& 38.2 & 52.7*& 47.4*& 43.6 \\
         \rowcolor{gray!15} AndesVL-0.6B-Instruct  & 1,866 & 65.3 & \textbf{39.7} & 44.3 & 58.6 & 57.2 & 55.3 \\
         \rowcolor{gray!15} AndesVL-0.6B-Thinking  & \textbf{1,925} & \textbf{66.3} & 36.1 & \textbf{49.7} & \textbf{61.6} & \textbf{59.8} & \textbf{57.1} \\
         \bottomrule
    \end{tabular}}
    \caption{Comparison of general VQA performance. Notice that the MME score is divided by 28 to calculate the overall score. The best results are marked in \textbf{bold}. Data marked with * are from our evaluation, while others are from their original papers or the OpenCompass leaderboard.}
    \label{tab:benchmark_general_vqa}
\end{table}

As illustrated in Table~\ref{tab:benchmark_general_vqa}, the AndesVL series (4B, 1B, and 0.6B) achieve the top performance in their respective groups, while the 2B model also delivers a highly competitive result. 
This suggests that our model extracts robust representations and displays a strong ability to comprehend real-world scenarios, enabling it to effectively tackle complex and dynamic tasks.

\subsection{Hallucination Evaluation}
\subsubsection{Benchmarks}

We evaluate AndesVL’s hallucination alleviation ability through a range of widely used hallucination benchmarks. These evaluations test the model's capabilities under  visual hallucination settings. The following three benchmarks are included in these evaluations.

\textbf{HallusionBench}~\cite{guan2023hallusionbench}: HallusionBench mainly evaluates a model's capabilities under language hallucination and visual illusion settings. The average of its three metrics---aAcc, fAcc, and qAcc---is taken as the reported performance score. 

\textbf{CRPE}~\cite{wang2024allCRPE}: CRPE quantitatively evaluates the object recognition and relation comprehension ability of MLLMs. The accuracy on the CRPE Relation subset is reported.

\textbf{POPE}~\cite{li2023pope}: POPE evaluates object hallucination tendencies in MLLMs. The overall average score is reported.

\subsubsection{Evaluation Results}
\begin{table}
    \centering
    \begin{tabular}{l|ccc|c}
        \toprule
         Model &  Hallucination & CRPE (relation) & POPE (avg) & Overall \\
         \hline
         Qwen2.5-VL-3B~\cite{bai2025qwen2_5}   & 46.6 & 64.9*& 89.3*& 66.9 \\
         BlueLM-2.5-3B~\cite{xiong2025bluelm} & 53.7 & - & - & -\\
         BlueLM-2.5-3B-thinking~\cite{xiong2025bluelm} & 57.3 & - & - & -\\
         Qianfan-VL-3B~\cite{dong2025qianfan} &  - & - & 85.1 & - \\
         Gemma3-4B~\cite{gemmateam2025gemma3} &  40.8 & 61.0*& 84.6 & 62.1 \\
         Phi-3.5-Vision-4B~\cite{abdin2024phi3} & 40.5 & 68.5*& 82.8 & 63.9\\
         Phi-4-Multimodal~\cite{abouelenin2025phi} & 40.5 & 72.0*& 85.6 & 66.0\\
         Ovis2-4B~\cite{lu2024ovis} & 53.8 & \textbf{77.0}*& 88.7 & 73.2 \\
         MiniCPM-V-4-4B~\cite{yao2024minicpm} & 50.8 & 74.6*& 82.4 & 69.3\\
         R-4B-RL~\cite{jiang2025r} & 58.9 & - & - & - \\
         InternVL3.5-4B~\cite{wang2025internvl3} & 44.8 & 75.0 & 88.9 & 69.6 \\
         \rowcolor{gray!15} AndesVL-4B-Instruct & 54.7& 75.8& 88.5& 73.0\\
         \rowcolor{gray!15} AndesVL-4B-Thinking & \textbf{59.2} & 75.5 & \textbf{89.8} & \textbf{74.8}\\
         \hline
         Qwen2-VL-2B~\cite{wang2024qwen2vl} &  42.4 & 68.5*& 87.3 & 66.1 \\
         MiniCPM-V-2B~\cite{yao2024minicpm} &  36.1 & 68.5*& 86.3 & 63.6 \\
         SAIL-VL-1.5-2B~\cite{dong2025scalable} &  49.8 & 73.9*& 87.7* &  70.5 \\
        SAIL-VL2-2B~\cite{yin2025sail} &  51.7 & 75.2& 86.8 & 71.2 \\
        Ovis2-2B~\cite{lu2024ovis} &  50.2 & 73.0*& 87.8 & 70.3 \\
         InternVL3-2B~\cite{zhu2025internvl3} & 42.5 & 71.5 & 89.6 & 67.9\\
         InternVL3.5-2B~\cite{wang2025internvl3} & 48.6 & \textbf{75.6}& 87.2 & 70.5\\
         \rowcolor{gray!15} AndesVL-2B-Instruct & \textbf{51.8} & 73.0 & 87.9 & 70.9 \\
         \rowcolor{gray!15} AndesVL-2B-Thinking & 51.4 & 74.1& \textbf{89.8} & \textbf{71.8} \\
         \hline
         Ovis2-1B~\cite{lu2024ovis} &  45.2 & 63.2 & 87.7 & 65.4\\
         InternVL3-1B~\cite{zhu2025internvl3} & 41.4 & 64.0 & \textbf{90.7} & 65.4 \\
         InternVL3.5-1B~\cite{wang2025internvl3} & 41.0 & 68.4 & 86.8 & 65.4\\
         \rowcolor{gray!15} AndesVL-1B-Instruct & 43.2 & 68.7 & 89.2 & 67.0\\
         \rowcolor{gray!15} AndesVL-1B-Thinking & \textbf{45.6} & \textbf{68.8} & 87.7 & \textbf{67.4} \\
         \hline
         SmolVLM2-0.5B~\cite{marafioti2025smolvlm} & 27.7 & 52.9*& 82.7 & 54.4 \\
         \rowcolor{gray!15} AndesVL-0.6B-Instruct & \textbf{45.3} & 67.4 & 84.3 & 65.7\\
         \rowcolor{gray!15} AndesVL-0.6B-Thinking & 42.5 & \textbf{68.3} & \textbf{86.8} & \textbf{65.9} \\
         \bottomrule
    \end{tabular}
    \caption{Comparison of hallucination alleviation performance. The best results are marked in \textbf{bold}. Data marked with * are from our evaluation, while others are from their original papers or the OpenCompass leaderboard.}
    \label{tab:benchmark_hallucination}
\end{table}

As illustrated in Table~\ref{tab:benchmark_hallucination}, the AndesVL series achieve exceptionally high scores overall: the 4B, 2B, 1B and 0.6B models attaining 74.8, 71.8, 67.4, and 65.9 points, respectively. It maintains a substantial lead over other models of comparable size by a margin of 1.5 to 11.5 points - a lead that becomes even more pronounced with smaller model scales. This finding demonstrates that our architecture delivers superior hallucination alleviation capabilities while maintaining high accuracy, a key strength that persists even in the smallest 0.6B variants.

\subsection{Multimodal Multilingual Understanding}
\subsubsection{Benchmarks}

We evaluate AndesVL’s multilingual understanding capabilities through the following three benchmarks:

\textbf{MMMB}~\cite{sun2024parrot}: MMMB assesses multilingual capabilities of MLLMs, comprising 6 languages, 15 categories, and 12,000 questions. The average score is reported.

\textbf{Multilingual MMBench}~\cite{sun2024parrot}: Multilingual MMBench extends the original MMBench~\cite{liu2023mmbench} dataset to six languages, including English, Chinese, Portuguese, Arabic, Turkish, and Russian. The average score is reported.

\textbf{MTVQA}~\cite{tang2024mtvqa}: MTVQA evaluates the multilingual capability of MLLMs with human-annotated, text-rich images across 9 diverse languages. The average accuracy on the test set is reported.

\subsubsection{Evaluation Results}
\begin{table}
    \centering
    \begin{tabular}{l|ccc|c}
        \toprule
         Model &  MMMB & Multilingual MMBench & MTVQA & Overall \\
         \hline
         Qwen2.5-VL-3B~\cite{bai2025qwen2_5}   & 76.9*& 74.9*& 24.8 & 58.9 \\
         Qianfan-VL-3B~\cite{dong2025qianfan} & - & - & 26.5 & - \\
         Gemma3-4B~\cite{gemmateam2025gemma3} &  69.6*& 65.3*& 22.1 & 52.4 \\
         Phi-3.5-Vision-4B~\cite{abdin2024phi3} & 61.3*& 67.4*& 12.3*& 47.0\\
         Phi-4-Multimodal~\cite{abouelenin2025phi} & 74.5*& 74.2*& 14.3& 54.3\\
         Ovis2-4B~\cite{lu2024ovis} &  79.7*& 81.2*& 29.4 & 63.4\\
         MiniCPM-V-4-4B~\cite{yao2024minicpm} & 72.8*& 70.7*& 22.6*& 55.4\\
         InternVL3.5-4B~\cite{wang2025internvl3}  & 80.2& 76.4& 29.6& 62.1\\
         \rowcolor{gray!15} AndesVL-4B-Instruct & \textbf{81.9} & 80.8 & \textbf{31.2}*& 64.6 \\
         \rowcolor{gray!15} AndesVL-4B-Thinking & 81.7 & \textbf{83.2} & 29.9 & \textbf{64.9} \\
         \hline
         Qwen2-VL-2B~\cite{wang2024qwen2vl} &  71.3*& 66.3*& 20.8 & 52.8 \\
         MiniCPM-V-2B~\cite{yao2024minicpm} &  60.0*& 51.3*& 9.3 & 40.2 \\
         SAIL-VL-1.5-2B~\cite{dong2025scalable} & 76.0*& 72.9*& 22.9*&  57.3 \\
         SAIL-VL2-2B~\cite{yin2025sail} & \textbf{79.9}*& \textbf{78.1}*& 27.2*&  \textbf{61.7} \\
        Ovis2-2B~\cite{lu2024ovis} &76.6*& 72.0*& 25.6 & 58.0 \\
         InternVL3-2B~\cite{zhu2025internvl3} &73.6& 71.9& 26.7 & 57.4 \\
         InternVL3.5-2B~\cite{wang2025internvl3} &74.6& 70.9& 28.5 & 58.0 \\
         \rowcolor{gray!15} AndesVL-2B-Instruct & 76.5 & 75.3 & \textbf{29.1} & 60.3 \\
         \rowcolor{gray!15} AndesVL-2B-Thinking & 76.5 & 75.0 & 26.7 & 59.4 \\
         \hline
         Ovis2-1B~\cite{lu2024ovis} & 70.8*& 62.6*& 23.7 & 52.4\\
         InternVL3-1B~\cite{zhu2025internvl3} & 63.2& 58.2& 22.2 & 47.9 \\
         InternVL3.5-1B~\cite{wang2025internvl3} & 66.0& 58.5& 22.9 & 49.1\\
         \rowcolor{gray!15} AndesVL-1B-Instruct & \textbf{72.0} & 63.0 & \textbf{24.9} & 53.3 \\
         \rowcolor{gray!15} AndesVL-1B-Thinking & 71.3 & \textbf{67.5} & 23.6 & \textbf{54.1} \\
         \hline
         SmolVLM2-0.5B~\cite{marafioti2025smolvlm} &  46.8*& 23.7*& 7.7 & 26.1 \\
         \rowcolor{gray!15} AndesVL-0.6B-Instruct & \textbf{70.3} & \textbf{60.8} & \textbf{21.8} & \textbf{51.0} \\
         \rowcolor{gray!15} AndesVL-0.6B-Thinking & 69.4 & 58.4 & 21.3 & 49.7 \\
         \bottomrule
    \end{tabular}
    \caption{Comparison of multilingual performance. The best results are marked in \textbf{bold}. Data marked with * are from our evaluation, while others are from their original papers or the OpenCompass leaderboard.}
    \label{tab:benchmark_multilingual}
\end{table}

As demonstrated in Table~\ref{tab:benchmark_multilingual}, both the Thinking and Instruct variants of AndesVL-4B demonstrate exceptional multilingual capabilities, achieving a leading score of 64.9, which surpasses the previous best model, Ovis2-4B~\cite{lu2024ovis}, by 1.5 points. This advantage persists in the smaller-scale variants of AndesVL, with each one achieving leading multilingual scores within their respective sub-groups.

The model's professional-grade multilingual capability provides a foundation for the cross-lingual transfer of its multimodal functions, a feature paramount for extending its global utility in mobile applications.

\subsection{GUI Tasks}
\subsubsection{Benchmarks}
In order to validate the capability of the AndesVL in terms of UI understanding, we conducted experiments on ScreenSpot~\cite{cheng2024seeclick}, ScreenSpot-V2~\cite{wu2024atlas}, ScreenSpot-Pro~\cite{li2025screenspot}, and our proposed AndesUI-Bench.

\textbf{ScreenSpot}~\cite{cheng2024seeclick}: ScreenSpot is a realistic GUI grounding benchmark that encompasses mobile, desktop, and web environments. It contains over 600 screenshots and 1200 instructions from iOS, Android, macOS, Windows, and webpages. It specifically includes both text-based elements and a variety of widgets and icons.

\textbf{ScreenSpot-V2}~\cite{wu2024atlas}: ScreenSpot-V2 is an enhanced version of the ScreenSpot benchmark that addresses annotation errors and ambiguities in the original dataset. Specifically, it corrects spelling errors, clarifies ambiguous instructions, removes overly similar questions, and fixes mislabeled ground-truth bounding boxes. These improvements ensure a more accurate and reliable evaluation for GUI grounding tasks.

\textbf{ScreenSpot-Pro}~\cite{li2025screenspot}: ScreenSpot-Pro is a new benchmark designed to evaluate the grounding capabilities of MLLMs in high-resolution professional settings. It includes 1,581 unique instructions in high-resolution screenshots sourced from 23 applications across five industries (development, creative, CAD, scientific, and office) and three operating systems (Linux, macOS, and Windows). The benchmark highlights the challenges of high-resolution displays, smaller target sizes, and complex environments in professional applications.

For the Screenspot, ScreenSpot-V2, and ScreenSpot-Pro datasets, the annotation format is bounding boxes. For each data instance, the model is required to output a specific coordinate; if the coordinate falls within the annotated bounding box, it is considered a correct prediction and contributes to the final accuracy. 

\textbf{AndesUI-Bench}: The AndesUI-Bench was developed to evaluate the smartphone UI understanding capabilities of MLLMs. As mentioned in Appendix~\ref{app:andesui_dataset}, the AndesUI-Bench represents the test set of the AndesUI dataset. This dataset includes 9k referring entries, 7.6k grounding entries, 455 comprehensive description entries, and 1.2k complex question-answer entries.

\subsubsection{Evaluation Results}
\begin{table}
    \centering
    \begin{tabular}{l|ccc|c}
        \toprule
         Model &  ScreenSpot & ScreenSpot\_v2 & ScreenSpot\_Pro & Overall \\
         \hline
         Qwen2.5-VL-3B~\cite{bai2025qwen2_5}   & 55.5*& 80.9* & 27.3* & 54.6  \\
         OS-Atlas-4B ~\cite{xiong2025bluelm} & 70.1 & 71.9 & - & - \\
         InternVL3.5-4B~\cite{wang2025internvl3} & 83.6& 85.1 & 18.1* & 62.3  \\
         \rowcolor{gray!15} AndesVL-4B-Instruct & 84.3 & 86.1 & 28.2 & 66.2  \\
         \rowcolor{gray!15} AndesVL-4B-Thinking & \textbf{85.2}& \textbf{87.4}& \textbf{32.5} & \textbf{68.4} \\
         \hline
         UI-TARS-2B ~\cite{wang2024qwen2vl} &  \textbf{82.3} & \textbf{84.7} & \textbf{27.7} &  \textbf{64.9}\\
         InternVL3-2B~\cite{zhu2025internvl3} & 45.1*& 47.0*& 1.0*& 31.0\\
         InternVL3.5-2B~\cite{wang2025internvl3} & 77.2*& 79.6*& 12.2*& 56.3\\
         \rowcolor{gray!15} AndesVL-2B-Instruct & 74.6 & 76.3 & 20.9 & 57.3\\
         \rowcolor{gray!15} AndesVL-2B-Thinking & 67.2 & 70.2 & 19.6 & 52.4\\
         \hline
         InternVL3-1B~\cite{zhu2025internvl3} & 31.3*& 30.9*& 0.6*& 20.9\\
         InternVL3.5-1B~\cite{wang2025internvl3} & 60.9*& 61.9*& 9.2*& 44.0\\
         \rowcolor{gray!15} AndesVL-1B-Instruct & 71.8 & 73.2 & \textbf{23.1} & 56.0\\
         \rowcolor{gray!15} AndesVL-1B-Thinking & \textbf{73.9} & \textbf{74.4} & 20.9 & \textbf{56.4}\\
         \bottomrule
    \end{tabular}
    \caption{Comparison of UI understanding performance with other general and specific models on ScreenSpot testset. The best results are marked in \textbf{bold}. Data marked with * are from our evaluation, while others are from their original papers.}
    \label{tab:benchmark_screenspot}
\end{table}

\begin{table}
    \centering
    \begin{tabular}{l|ccc|c}
        \toprule
         Model &  Grounding & Referring & QA & Overall \\
         \hline
         Qwen2.5-VL-3B~\cite{bai2025qwen2_5}   & 61.4 & 62.1 & 15.0 & 46.2  \\
         InternVL3.5-4B~\cite{wang2025internvl3} & 91.9 & 68.9 & 82.4 & 81.1  \\
         \rowcolor{gray!15} AndesVL-4B-Instruct & \textbf{95.1} & 72.5 & \textbf{82.6} & 83.4  \\
         \rowcolor{gray!15} AndesVL-4B-Thinking & 94.5& \textbf{73.4}& 82.5 & \textbf{83.5} \\
         \bottomrule
    \end{tabular}
    \caption{Comparison of UI understanding performance on AndesUI-Bench testset. The best results are marked in \textbf{bold}. All results are from our evaluation.}
    \label{tab:benchmark_andesui}
\end{table}

In this study, we present a comprehensive analysis of various models' performance in UI understanding tasks. As illustrated in Tab.~\ref{tab:benchmark_screenspot}, AndesVL-4B surpasses other models of comparable size in accuracy, achieving a leading score of 68.4. While slightly trailing behind UI-TARS-2B, a specialized model in the GUI domain, the AndesVL 2B and 1B variants maintain highly competitive performance, demonstrating robust UI comprehension capabilities.

Tab.~\ref{tab:benchmark_andesui} shows the performance comparison between AndesVL and other leading open-source models on the AndesUI-Bench testset. AndesVL-4B delivers outstanding performance across all evaluation metrics, achieving the top score of 83.5. These results collectively demonstrate our model's substantial expertise and distinct competitive advantage in UI understanding and application.

\subsection{Ablation Studies}

\begin{table}[tb]
    \centering
    \resizebox{1\textwidth}{!}{
    \begin{tabular}{c|c|c|c|c|c|c}
    \toprule
         \multirow{1}{*}{Model}& \multirow{1}{*}{MMVet}& \makecell{MathVerse\\(vision-only)}&\multirow{1}{*}{RealWorldQA}&\multirow{1}{*}{OCRBench}& \multirow{1}{*}{HallusionBench }& \multirow{1}{*}{Overall}\\
         \hline
         AndesVL-2B-Instruct-Base & 48.3 & 22.8 & 65.1 & 82.4 & 49.3 & 53.6\\
         AndesVL-2B-Instruct-SFT  & 51.0 & 25.9 & 66.9 & 83.9 & 49.4 & 55.4\\
         AndesVL-2B-Instruct-MPO  & 52.0 & 26.8 & 67.8 & 84.6 & 51.8 & 56.6\\
         \bottomrule
    \end{tabular}}
    \caption{Comparison on several general benchmarks among AndesVL-2B-Instruct-Base, AndesVL-2B-Instruct-SFT and AndesVL-2B-Instruct-MPO.}
    \label{tab:ablation_1}
\end{table}

\begin{table}[tb]
    \centering
    \resizebox{1\textwidth}{!}{
    \begin{tabular}{c|c|c|c|c|c|c|c}
    \toprule
         \multirow{1}{*}{Model}& \makecell{MathVista\\(mini)}& \makecell{MathVision}& \multirow{1}{*}{WeMath} &\makecell{MathVerse\\(vision-only)} &\multirow{1}{*}{MMMU} & \multirow{1}{*}{MMMU\_Pro} & \multirow{1}{*}{Overall}\\
         \hline
         AndesVL-2B-Thinking-Base & 68.1 & 32.1 & 38.2 & 51.7 & 48.0 & 35.7 & 45.6\\
         AndesVL-2B-Thinking-SFT  & 69.7 & 32.0 & 37.0 & 52.3 & 52.6 & 35.5 & 46.5\\
         AndesVL-2B-Thinking-RL   & 73.3 & 35.2 & 41.1 & 54.8 & 52.1 & 37.3 & 49.0\\
         \hline
         AndesVL-4B-Thinking-Base & 76.2 & 48.1 & 49.5 & 64.9 & 62.3 & 46.0 & 57.8\\
         AndesVL-4B-Thinking-SFT  & 77.4 & 48.4 & 54.2 & 66.4 & 64.8 & 48.7 & 60.0\\
         AndesVL-4B-Thinking-RL   & 79.4 & 51.0 & 57.4 & 66.9 & 66.9 & 51.4 & 62.2\\
         \bottomrule
    \end{tabular}}
    \caption{Comparison on several reasoning and math benchmarks among  AndesVL-2B-Thinking-Base, AndesVL-2B-Thinking-SFT, AndesVL-2B-Thinking-RL, AndesVL-4B-Thinking-Base, AndesVL-4B-Thinking-SFT and AndesVL-4B-Thinking-RL.}
    \label{tab:ablation_2}
\end{table}

In Table \ref{tab:ablation_1}, we present ablation results for AndesVL-2B-Instruct-Base, AndesVL-2B-Instruct-SFT and AndesVL-2B-Instruct-MPO. We find that MPO enhances the mathematical reasoning and multimodal understanding capabilities of the Instruct model, with the MPO model achieving improvements of over 1.0 pp and 0.9 pp on MMVet and MathVerse respectively. Furthermore, MPO improves the model's ability to resist hallucinations, yielding a 1.4 pp gain on HallusionBench. Notably, RealWorldQA and OCRBench show 0.9 pp and 0.7 pp improvement respectively. We thought that this is because MPO corrects errors made by the SFT model on instances it was originally capable of solving correctly. 

In Table \ref{tab:ablation_2}, we present the ablation studies for AndesVL-2B-Thinking-Base, AndesVL-2B-Thinking-SFT, AndesVL-2B-Thinking-RL, AndesVL-4B-Thinking-Base, AndesVL-4B-Thinking-SFT and AndesVL-4B-Thinking-RL. As shown, the model trained with RL exhibits significant improvements in mathematical reasoning. For instance, performance increases by about 2 pp on MathVista, MathVision and WeMath on both AndesVL-2B-Thinking and AndesVL-4B-Thinking. Moreover, the RL-enhanced model also demonstrates improved performance on complex multimodal understanding tasks, such as over 2 pp improvement on MMMU\_Pro. These results indicate that RL significantly enhances the model’s multimodal understanding and mathematical reasoning capabilities evenif the model has only 2B parameters.

Overall, both the Instruct and Thinking models exhibit improved performance after the SFT stage. For the Instruct model, MPO leads to significant gains in mathematical reasoning, multimodal understanding capabilities, OCR accuracy, and hallucination resistance. For the Thinking model, RL notably enhances its abilities in mathematical reasoning and complex multimodal understanding.

\section{On-Device Performance}
\label{sec:experiments_on-device}

\subsection{Results of Quantization-Aware Training}
To evaluate the capabilities of our on-device models, we use OCR capabilities as a testbed and conduct experiments on multiple OCR-related benchmarks, including DocVQA~\cite{mathew2021docvqa}, InfoVQA~\cite{mathew2022infographicvqa}, TextVQA~\cite{singh2019textvqa} and ChartQA~\cite{masry2022chartqa}.

As mentioned in Sec.~\ref{sec:QAT_AndesVL}, directly applying PTQ to floating-point models can significantly degrade model performance, and we introduced QAT to solve this. We compare the quantized and floating-point models based on Top-1 overlap across multiple OCR-related benchmarks. The experimental results are shown in Table~\ref{tab:QAT_top1_qcc}. In Table~\ref{tab:QAT_top1_qcc}, AndesVL-4B-Instruct-Base (PTQ) represents the model of AndesVL-4B-Instruct-Base post-trained on OCR data  with PTQ, AndesVL-4B-Instruct-Base (QAT+PTQ) is the model of AndesVL-4B-Instruct-Base post-trained on OCR data with QAT and PTQ. The results demonstrate that QAT+PTQ can achieve 95\% Top-1 overlap~\cite{li2025bild} between the quantized and floating-point models, and achieves significant improvement over PTQ alone.

\begin{table}[tb]
\centering
\begin{tabular}{l|cccc|c}
    \toprule
    Model           &\makecell{DocVQA\\(test)} &\makecell{InfoVQA\\(test)}& \makecell{TextVQA\\(val)} &  \makecell{ChartQA\\(test)}  & Overall   \\ 
    \toprule
    AndesVL-4B-Instruct-Base (PTQ)       & 93.2   &  89.0 &  91.4 & 89.3    & 90.7 \\
    AndesVL-4B-Instruct-Base (QAT+PTQ)   & 95.4   &  95.2 & 97.5  & 95.1     & 95.8 \\ 
    \bottomrule
    \end{tabular}
\caption{Top-1 overlap between AndesVL-4B-Instruct-Base (PTQ) and AndesVL-4B-Instruct-Base (QAT+PTQ) on 4 OCR benchmarks.}
\label{tab:QAT_top1_qcc}
\end{table}

\subsection{Results of QALFT}
To further improve performance across various on-device scenarios, we utilize the QLAFT framework to train LoRA weights specific to each situation.  The experiments mentioned in the Table \ref{tab:qat_qalft_performance} are all completed based on the pre-trained AndesVL-4B-Instruct-Base model. AndesVL-4B-Instruct-Base-LoRA (floating point) represents LoRA fine-tuned floating-point model trained on  OCR data, AndesVL-4B-Instruct-Base-LoRA (PTQ) represents AndesVL-4B-Instruct-Base-LoRA (floating point) with PTQ, AndesVL-4B-Instruct-Base-LoRA (QAT+PTQ) represents AndesVL-4B-Instruct-Base-LoRA (floating point) with QAT and PTQ, AndesVL-4B-Instruct-Base (QAT+PTQ+QALFT) represents QALFT training on AndesVL-4B-Instruct-Base with QAT and PTQ. The experimental results in Table \ref{tab:qat_qalft_performance} show that the performance of the model only with PTQ decreases significantly, QAT and QALFT can significantly improve the performance of the model on the device side, while QALFT can decrease slightly by 3\% compared to the
floating-point model.
\renewcommand{\arraystretch}{1.5}
\begin{table}[tb]
\resizebox{1\textwidth}{!}{
\centering
\begin{tabular}{l|ccc|c}
    \toprule
    LoRA Models   & \makecell{TextVQA\\(val)} & \makecell{ChartQA\\(test)} &  \makecell{AI2D\\(w M)} & Overall \\ 
    \hline
    AndesVL-4B-Instruct-Base-LoRA (floating point)  & 81.1  & 87.5  &  83.4 &  84.0       \\ 
    \hline
    AndesVL-4B-Instruct-Base-LoRA (PTQ)             &  67.2 &  66.1 &  65.5 &  66.3       \\
    AndesVL-4B-Instruct-Base-LoRA (QAT+PTQ)         &  77.2 &  84.0 &  80.8 &   80.7     \\ 
    \hline
    AndesVL-4B-Instruct-Base (QAT+PTQ+QALFT)        &  80.8 &   86.4 & 81.3  &     82.8    \\ 
    \bottomrule
\end{tabular}}
\caption{Comparison on 3 OCR benchmarks performance among AndesVL-4B-Instruct-Base-LoRA (floating point), AndesVL-4B-Instruct-Base-LoRA (PTQ), AndesVL-4B-Instruct-Base-LoRA (QAT+PTQ) and AndesVL-4B-Instruct-Base (QAT+PTQ+QALFT), the results prove that QAT and QALFT significantly improve performance.}
\label{tab:qat_qalft_performance}
\end{table}

\subsection{Results of Cache Eviction}
Our cache eviction strategy is tailored for tasks with long prompts. We use the call summary task, which is a popular and pioneering feature of OPPO AI phones and involves substantial input information redundancy, to verify its effectiveness.
In this task, our proprietary OKV cache eviction algorithm results in a more than 10\% improvement in Rouge-1 relative to SnapKV with 50\% eviction ratios. In certain instances, it even outperformed the baseline with full KV caches. Comprehensive results presented in Table \ref{tab:cache_eviction_performance}. All experiments are based on the same AndesVL-4B-Instruct-Base model and are carried out on one device. The baseline AndesVL-4B-Instruct-Base  is supervised fine-tuned on the call summary task; SnapKV and OKV are applied to the model for inference respectively.

\begin{table}[tb]
\centering
\begin{tabular}{l l ccc}
\toprule
\textbf{Eviction Ratio} & \textbf{Method} & \textbf{ROUGE-1} & \textbf{ROUGE-2} & \textbf{ROUGE-L} \\
\midrule
0\% (Baseline)          & AndesVL-4B-Instruct-Base & 0.59 & \textbf{0.33} & \textbf{0.42} \\
\midrule
25\%                    & SnapKV                   & 0.55 & 0.30          & 0.39 \\
                        & OKV                      & \textbf{0.60} & \textbf{0.33} & 0.41 \\
\midrule
50\%                    & SnapKV                   & 0.50 & 0.25          & 0.36 \\
                        & OKV                      & 0.56 & 0.30          & 0.39 \\
\bottomrule
\end{tabular}
\caption{ROUGE performance of the reproduced SnapKV and our OKV under 25\% and 50\% key-value cache eviction ratios on the call summary task.}
\label{tab:cache_eviction_performance}
\end{table}

\subsection{Results of Speculative Decoding}
Our customized speculative decoding achieves significant decoding acceleration across multiple multimodal and text-only tasks. We combined it with our key breakthrough in LLM sparsification and MediaTek's hardware-aware compression, and show the final results in Table \ref{tab:speculative_decoding_performance}. In this table, the PTQ (baseline) represents the quantized version of the floating point AndesVL-4B-Instruct-Base, + Hardware-aware compression represents PTQ (baseline) with hardware compression, + Sparsification denotes PTQ (baseline) with hardware-aware compression and sparsification, and + Speculative decoding denotes PTQ (baseline) with speculative decoding, sparsification, and hardware-aware compression. The results show that we can achieve 6.7x peak decoding speedup ratio and 1.8 bits-per-weight under extreme sparsification and hardware-aware compression. Moreover, we achieved a memory reduction of up to 30.9\% on the MediaTek Dimensity 9500 chips.

\begin{table}[tb]
\centering
\begin{tabular}{lcc}
\toprule
\textbf{Compression \& Acceleration Method} & \textbf{Peak Speedup} & \textbf{BPW} \\
\midrule
PTQ (baseline)                              & 1.0$\times$           & 3.0 \\
\quad + Hardware-aware compression                  & 1.1$\times$           & 3.0 \\
\quad\quad + Sparsification                                  & 1.6$\times$           & 1.8 \\
\quad\quad\quad + Speculative decoding                      & 6.7$\times$           & 1.8 \\
\bottomrule
\end{tabular}
\caption{Peak decoding speedup ratio and bits-per-weight (BPW) of AndesVL-4B-Instruct-Base under various compression and acceleration techniques on an edge device. The baseline is PTQ-only.}
\label{tab:speculative_decoding_performance}
\end{table}

\section{Future Directions}
\label{sec:future_directions}

In the future, several promising directions can be explored to further enhance the capabilities of mobile-side MLLMs. First, designing more optimal visual encoder solutions holds great potential. By leveraging advanced network architectures and novel feature extraction strategies, we aim to improve the efficiency and accuracy of visual information processing, enabling the model to better understand complex visual content on resource-constrained mobile-side devices.

Second, developing superior post-training schemes is crucial. Refining the post-training process can optimize the model performance in handling various multimodal tasks, reduce hallucinations, and enhance the consistency and reliability of generated outputs. This may involve exploring new types of training data, adjusting training objectives, and optimizing training algorithms to make the model more adaptable to real-world scenarios.

Third, implementing effective distillation schemes between large and small models can significantly improve the performance-to-resource ratio of mobile-side models. By transferring knowledge from large, high-performance cloud-based models to smaller mobile-side counterparts, we can boost the capabilities of the latter while maintaining low computational costs and memory requirements.

Finally, the development of a unified mobile-side model integrating text, image, and speech modalities (a three-mode integrated model) represents an exciting frontier. Such a model would enable seamless interaction with users across multiple modalities, providing more natural and intelligent user experiences. This will require in-depth research on multimodal fusion techniques, cross-modal representation learning, and efficient inference algorithms to ensure the model's effectiveness and efficiency on mobile-side devices. These research directions will not only drive the progress of mobile-side MLLMs but also expand their application scope in various fields.

\section{Conclusion}
\label{sec:conclusion}

This paper presents AndesVL, a suite of mobile-side MLLMs with parameter sizes ranging from 0.6B to 4B. By integrating Qwen3's LLM and various visual encoders, AndesVL achieves first-tier performance on multiple open-source benchmarks and the self-developed AndesUI benchmark, especially excelling in mobile UI understanding. The proposed 1+N LoRA architecture and Quantization-Aware LoRA Fine Tuning (QALFT) framework enable efficient task adaptation and model compression. QALFT ensures that AndesVL maintains high precision performance with only ignorable degradation (3\%) after deployment on mobile devices compared to the original floating-point model. By employing our proposed OKV, meticulously designed speculative decoding techniques and compression strategies, we can achieve 1.8 bits-per-weight,  6.7x peak decoding speed ratio and up to 30.9\% memory reduction when deploying AndesVL-4B on MediaTek Dimenisity 9500 chips. This work bridges the gap between cloud-based MLLMs and edge devices, providing a practical solution for mobile-side MLLM  and paving the way for future advancements in edge AI.

{
    \small
    \bibliographystyle{plain}
    \bibliography{main}
}
\newpage
\appendix
\section{Contributor}
\label{app:contributor}

Zhiwei Jin, Xiaohui Song, Nan Wang, Yafei Liu, Chao Li, Xin Li, Ruichen Wang, Zhihao Li, Qi Qi, Long Cheng, Dongze Hao, Quanlong Zheng, Yanhao Zhang, Haobo Ji, Jian Ma, Zhitong Zheng, Zhenyi Lin, Haolin Deng, Xin Zou, Xiaojie Yin, Ruilin Wang, Liankai Cai, Haijing Liu, Yuqing Qiu, Ke Chen, Zixian Li, Chi Xie, Huafei Li, Chenxing Li, Chuangchuang Wang, Kai Tang, Zhiguang Zhu, Kai Tang, Wenmei Gao, Rui Wang, Jun Wu, Chao Liu, Qin Xie, Chen Chen\footnote{chenchen4@oppo.com}, Haonan Lu\footnote{luhaonan@oppo.com}
\begin{table}
    \begin{tabular}{l|l}
        \hline
        \textbf{Category} & \textbf{APP Names} \\ 
        \hline
        Shopping & Alibaba, Dewu, JD, Pinduoduo, Taobao, Taote, Xianyu, Vipshop \\ 
        \hline
        Transportation & Baidu Maps, Amap, Tencent Maps, Hello, Didi, Traffic 12123, Railway 12306 \\ 
        \hline
        Lifestyle Services & Meituan, Dazhong Dianping, Ele.me, Meituan Waimai, Ctrip, \\ 
                       & Qunar, SF Express \\ 
        \hline
        Automotive & Dongchedi, Autohome \\ 
        \hline
        Telecommunications & State Grid Online, China Telecom, China Unicom, China Mobile \\ 
        \hline
        Video & Tencent Video, iQIYI, Bilibili, Youku, Kuaishou, Douyin, \\ 
              & Migu Video, Tencent Animation, Hongguo Short Drama \\ 
        \hline
        Social Media & Toutiao, Weibo, WeChat, Xiaohongshu, Douban, \\ 
                     & Zhihu, Baidu Tieba, Momo, Facebook, YouTube \\ 
        \hline
        Gaming & Xiaohonghe, League of Legends Mobile, Happy Match \\ 
        \hline
        Music & NetEase Cloud Music, Ximalaya \\ 
        \hline
        Fitness & Keep \\ 
        \hline
        Tools & Tianyancha, Quark, Cloud Flash Pay, Industrial and Commercial Bank of China, \\ 
              & 58 City, Meitu Xiuxiu \\ 
        \hline
        OPPO Built-in Apps & Settings, Phone Migration, Xiaobu Assistant, Clock, Weather, \\ 
                  & Calendar, Notes, Calculator, Compass, Camera, \\ 
                  & Recorder, Album, Music, OPPO Video, Reader, \\ 
                  & Contacts, Dialer, Messages, Mini Games, Game Center, \\ 
                  & Wallet, Cloud Services, My OPPO, OPPO Store, Main App Store \\ 
        \hline
    \end{tabular}
    \caption{App List from Andes-UI Dataset Collection}
    \label{tab:app_category}
\end{table}
\begin{table}
    \centering
    \begin{tabular}{l|c|c}
        \hline
        \textbf{Data Type} & \textbf{Training Set} & \textbf{Test Set} \\ 
        \hline
        Total Screenshots & 13002 & 455 \\ 
        \hline
        Referring Data Count & 226901 & 8642 \\ 
        \hline
        Grounding Data Count & 185968 & 7194 \\ 
        \hline
        Overall descriptive data & 13002 & 455 \\ 
        \hline
        Natural Q\&A Pairs & 107688 & 1181 \\ 
        \hline
    \end{tabular}
    \caption{AndesUI Dataset Statistics}
    \label{tab:andesui_stat}
\end{table}
\section{AndesUI Dataset}
\label{app:andesui_dataset}
In this section, we provide a comprehensive presentation of the AndesUI dataset construction pipeline, including the data collection process, human annotation, and data generation.

\textbf{Selection of APPs.}
We collected a total of 90 APPs, comprising 65 popular download APPs from the OPPO Software Store, covering a wide range of categories commonly used by users, along with 25 ColorOS pre-installed APPs. These APPs are listed in Table~\ref{tab:app_category}.

\textbf{Screenshot Data Collection.}
For each APP, we instructed annotators to capture screenshots of various diverse pages within the app, ensuring that each screenshot had distinct layouts and content. If two screenshots had similar layout structures but differed solely in text and images, they were classified as homogeneous interfaces. Our objective was to maximize diversity within the dataset while covering all typical interfaces of the app. Depending on the homogeneity degree, we collected between 1 and 10 screenshots for each heterogeneous page. For example, in the Xiaohongshu post interface, the display of different users' posts is similar enough to be regarded as a homogeneous page; however, since some posts include images while others do not, we aimed to collect additional screenshots from this homogeneous interface.

Throughout the screenshot collection process, we focused on capturing various atypical scenarios, including network interruptions and pop-ups (encompassing advertisement, log-in, confirmation, and phone pop-ups). For the training dataset, we collected a total of 10,747 screenshots from third-party apps and 2,255 screenshots from system pre-installed apps. In the testing set, there were a total of 455 screenshots. These screenshots were heterogeneous to reduce duplicate and similar pages. All detailed statistics of the dataset is shown in Table~\ref{tab:andesui_stat}

\begin{figure}[htbp]
    \centering
    \begin{subfigure}[b]{0.3\textwidth}
        \centering
        \includegraphics[width=\textwidth]{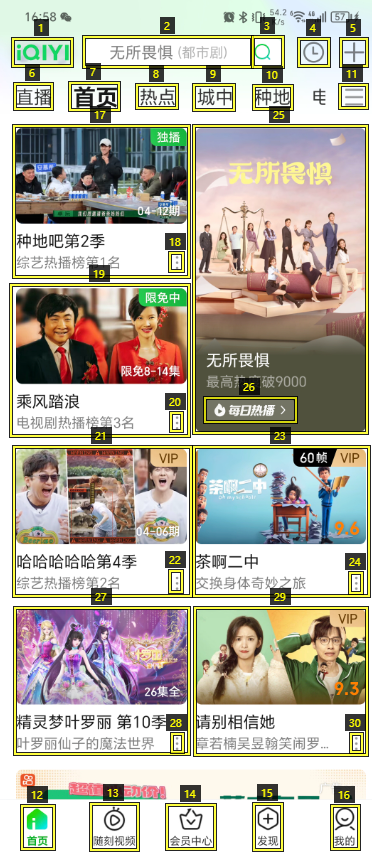}
        \label{fig:subfig1}
    \end{subfigure}
    \hfill
    \begin{subfigure}[b]{0.3\textwidth}
        \centering
        \includegraphics[width=\textwidth]{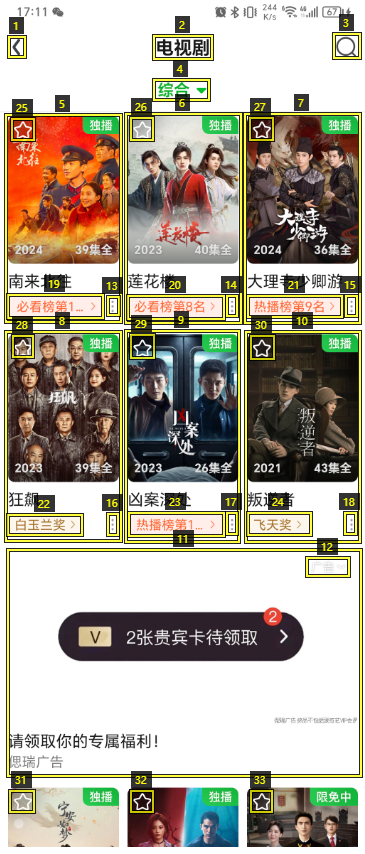}
        \label{fig:subfig2}
    \end{subfigure}
    \hfill
    \begin{subfigure}[b]{0.3\textwidth}
        \centering
        \includegraphics[width=\textwidth]{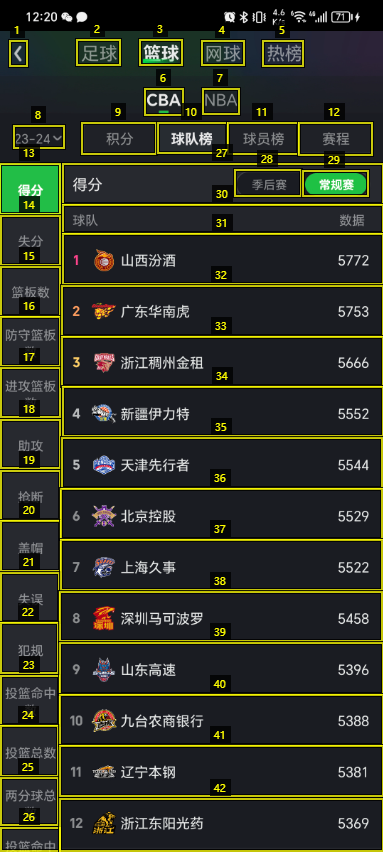}
        \label{fig:subfig3}
    \end{subfigure}
    \caption{Examples of widget labels in the AndesUI dataset.}
    \label{fig:ui_widget}
\end{figure}

\textbf{Annotation of Widgets.}
Our objective was to provide annotations for all widgets present within each screenshot. This included delineating bounding boxes, identifying widget types, recording any text on the widgets (when available), and indicating whether they are clickable, among other details. For this process, we employed the VIA-2.0.12 tool~\cite{vgg_via}. Annotating all widgets manually from scratch is a labor-intensive endeavor; hence, we initially used Qwen2-VL-72B~\cite{wang2024qwen2vl} to generate preliminary annotations on each screenshot, converting these annotations into a JSON format compatible with VIA. Subsequent modifications and refinements were then carried out by annotators. On average, each interface resulted in 18 widgets. The training dataset contained a total of 226,901 widgets, while the testing dataset included 9,068 widgets. Examples of labeled widgets of screenshots are provided in Fig.~\ref{fig:ui_widget}.

We needed to construct both basic and advanced data. Basic data includes grounding and referring data, while advanced data consists of comprehensive descriptive data and natural question-answer pairs. Basic data can be generated through programmatic means. In particular, for each widget, a single grounding data entry and a single referring data entry are generated. As an illustration, for a ``send'' widget with coordinates [3212, 1045, 3550, 2242], the associated grounding and referring data are
\begin{itemize}
    \item Question: ``Can you tell me the coordinates of the widget named 'send'?''
Answer: "<|box\_start|>(3212, 1047),(3550, 2242)<|box\_end|>"
\item Question: ``What is the widget located within the bounding box <|box\_start|>(3212, 1047),(3550, 2242)<|box\_end|>?''
Answer: ``send''
\end{itemize}
Consequently, the training dataset produced 226,901 data entries for referring and 185,968 for grounding. The test dataset included 7,194 grounding entries and 8,642 referring entries. This discrepancy occurs because a single page can contain several widgets sharing the same name, leading to fewer grounding entries. The questions were randomly selected from a seed library of questions. Initially, GPT-4 was employed to create 100 different question formulations. The bounding box coordinates underwent normalization and were then scaled by a factor of 10,000.

\textbf{Generation of Advanced Data.}
For comprehensive descriptive data, each screenshot was analyzed by aggregating the details of individual widgets. Subsequently, GPT-4 was utilized to generate a detailed description of the current page, including the theme, function, spatial arrangement of principal widgets, and a general usage guide for the interface.

For dataset creation involving natural question-answer pairs, we constructed several pairs for each screenshot. To achieve this, we initially utilized the information from each widget to instruct GPT-4 to formulate approximately ten question-answer pairs, emulating possible user inquiries during real-world application. The natural question-answer pairs were divided into four categories: descriptive questions, locating questions, interaction questions, and questions regarding natural scenes. They can also be classified by difficulty level: easy, medium, and hard. Questions classified as ``easy'' can be immediately answered, whereas ``hard'' questions might necessitate reasoning or multiple steps to resolve. Initially, GPT-4 was employed to generate several preliminary questions, which were then refined by annotators. Ultimately, we generated 107,688 natural question-answer pairs for the training set and constructed 1,181 pairs for the test set. Below is the system prompt used to guide GPT-4 in generating the natural question-answer pairs:

\textit{
You are an AI visual assistant capable of analyzing mobile screens. You will receive a screenshot from the \{app\_info\} app of the \{page\_description\} page, along with a string representation of a widget dictionary.
Each element in the dictionary is a dictionary that represents a UI widget, where the key is the widget number and the value contains information about the widget, including its bounding box coordinates, widget type, and widget description. The bounding box coordinates are represented as (x1, y1, x2, y2), with floating-point values ranging from 0 to 1.
Based on the provided text and coordinates, please design several simulated question-and-answer dialogues that represent interactions between the user and the system. These dialogues should focus on the user's potential actions on the screen (rather than perceptions).
The questions you create should be divided into three levels of difficulty: easy, medium, and hard. Easy questions can be answered directly from the widget dictionary. Medium and hard questions require some reasoning.
The questions can also be categorized into four content types: descriptive questions, locating questions, interaction questions, and natural scene questions. Here are four examples for reference; they may not be related to this image, and please do not restrict yourself to these few questions.
Examples of descriptive questions: Can you describe the function of widget\_23? / How many video list items are there in this screenshot?
Examples of locating questions: How do I access the creation page from the current page? / How can I view detailed information about the fourth video?
Examples of interaction questions: Can widget\_2 be swiped?
Examples of natural scene questions: How can I search for the latest movies? / How do I share the second video on social media?
For non-descriptive questions, you do not need to specify the type information of the widget in your responses.
When creating dialogues involving specific widgets, please strictly refer to the widget number (e.g., "widget\_3") rather than using the bounding box coordinates. This is necessary for me to use this data for function calls, so clear reference is required.
Your response format should be: [\{"User":"...","System":"...","Difficulty Level":"","Content Category":""\},...]
}
\section{Qualitative Examples}
\label{app:qualitative_examples}
In this section, a series of qualitative examples are illustrated for various capabilities of the AndesVL model by presenting responses generated from AndesVL-4B. Fig. \ref{tab:general_case_caption} shows strong image understanding; Fig. \ref{tab:general_case_multiimage_1}, \ref{tab:general_case_multiimage_2} and \ref{tab:general_case_multiimage_3} demonstrate multi-image perception ability; Fig.~\ref{tab:general_case_reasoning_1} shows basic reasoning capability of AndesVL; and Fig.~\ref{tab:case_ocr_1} and \ref{tab:case_ocr_2} illustrate document understanding and OCR capabilities.

Examples demonstrating AndesVL-4B-Thinking’s visual reasoning capability are provided in Fig. \ref{tab:case_math_1}, \ref{tab:case_math_2} and \ref{tab:case_math_3}. Fig. \ref{tab:case_math_1} shows a case of AndesVL-4B-Thinking solving a geometry problem. Fig. \ref{tab:case_math_3} illustrates the model's capability of handling reasoning tasks related to charts.  

UI understanding and reasoning cases of AndesVL-4B are shown in Fig.~\ref{tab:case_ui_1} and \ref{tab:case_ui_2}.

\begin{table}[h]
\centering
\begin{tabular}{c|l}
\hline
\multirow{8}*{{\includegraphics[width=3cm]{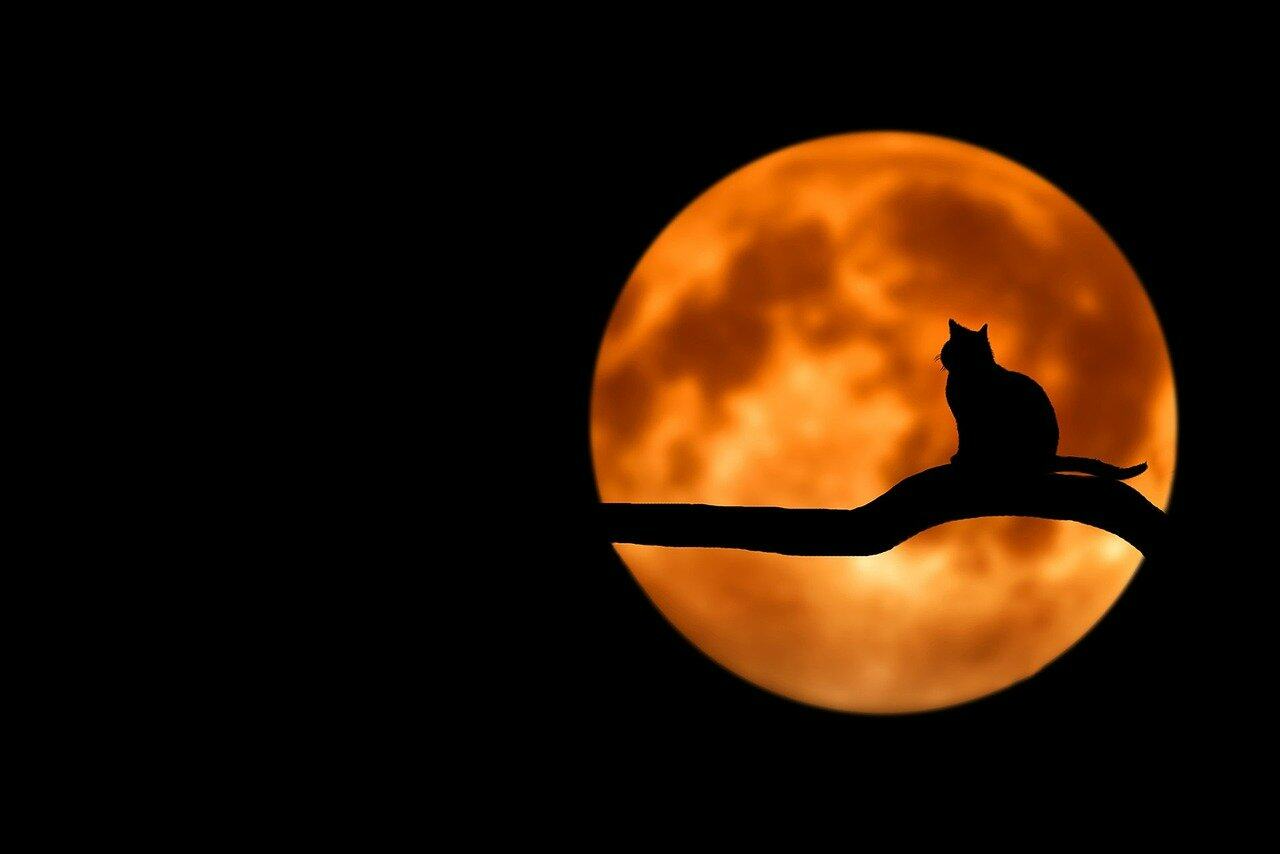}}}
& \textbf{Prompt}:  详细描述一下这张图片的内容。\\
& \textbf{Model Response}: \\
& 这张图片展示了一个夜晚的场景。在图片的右侧，有一只猫，它正坐在一\\
& 根树枝上，猫的轮廓清晰地映衬在背景的明亮光线中。这只猫的姿势看起\\
& 来很放松，它正面对着树枝的另一端。背景中，有一个巨大的满月，月光\\
& 呈现出橙红色调，与夜晚的黑色背景形成鲜明对比。月球的表面细节不是\\
& 很清晰，但可以看到一些暗淡的云状纹理。整个场景给人一种宁静而神秘\\
& 的感觉。\\
\hline
\end{tabular}
\caption{A dense captioning example of AndesVL.}
\label{tab:general_case_caption}
\end{table}

\begin{table}[h]
\centering
\begin{tabular}{c|l}
\hline
\multirow{9}*{{\includegraphics[width=3cm]{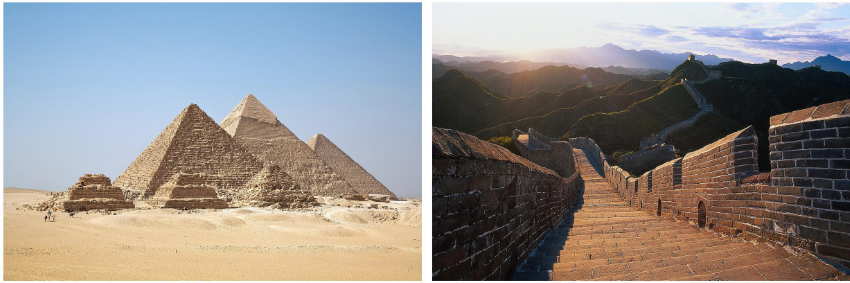}}}
& \textbf{Prompt}: What are these attractions? Please give a short description of the \\
& attractions and output in Markdown format.\\
& \textbf{Model Response}: \\
& - **Left**: The Pyramids of Giza, located in Egypt, are the largest and oldest \\
& of the three pyramids in the Giza pyramid complex. They were built as tombs \\
& for the Pharaohs of Ancient Egypt.\\
& - **Right**: The Great Wall of China is an ancient series of walls and \\
& fortifications that was built to protect China from invasions. It stretches for \\
& over 13,000 miles and is one of the most famous landmarks in the world. \\
\hline
\end{tabular}
\caption{A knowledge question answering example of AndesVL.}
\label{tab:general_case_multiimage_1}
\end{table}

\begin{table}[h]
\centering
\begin{tabular}{c|l}
\hline
\multirow{16}*{{\includegraphics[width=3cm]{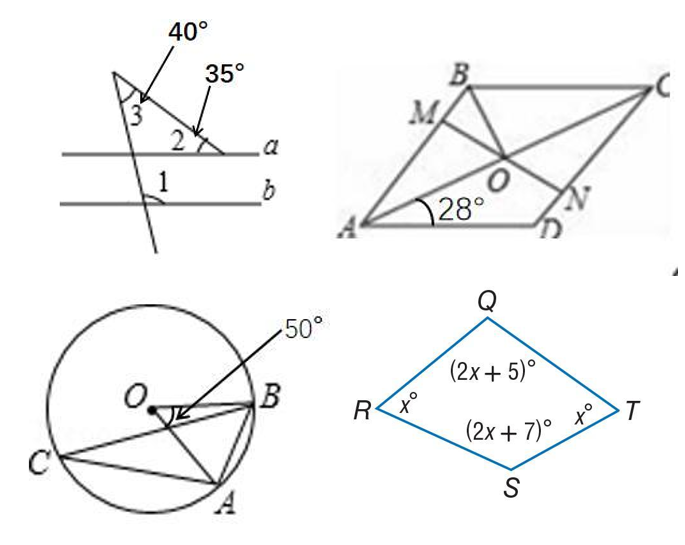}}}
& \textbf{Prompt}: Here is one math problem with its ground-truth answer, \\
& the solving of which requires interpreting the paired diagram. \\
& Please select the correct paired math diagram from the four given images. \\
& Question: As shown in the figure, the straight line a is parallel to b, \\
& then the degree of angle 1 is (). \\
& \# Choices: \\
& - A:75° \\
& - B:105° \\
& - C:140° \\
& - D:145° \\
& Answer the question first and give a brief explanation of your choice. \\
& \textbf{Model Response}: \\
& B: 105° \\
& The given angles are 40° and 35°. Since line a is parallel to line b, \\
& angle 2 is equal to 40°. Therefore, angle 1 is the sum of angles 2 and 35°, \\
& which is 40° + 35° = 75°. However, angle 1 is an exterior angle to the \\
& triangle formed by angles 35° and 40°, so the exterior angle theorem tells\\
& us that angle 1 is equal to 180° - 75° = 105°. \\
\hline
\end{tabular}
\caption{An Example of Multi-Image Understanding of AndesVL.}
\label{tab:general_case_multiimage_2}
\end{table}

\begin{table}[h]
\centering
\begin{tabular}{c|l}
\hline
\multirow{16}*{{\includegraphics[width=3cm]{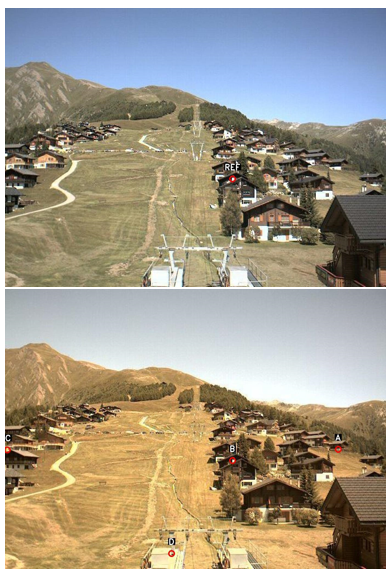}}}
& \textbf{Prompt}: A point is circled on the first image, labeled with REF. \\
& We change the camera position or lighting and shoot the second image. \\
& You are given multiple red-circled points on the second image, \\
& choices of "A, B, C, D" are drawn beside each circle. \\
& Which point on the second image corresponds to the point in the first image? \\
& \# Select from the following options first and briefly explain the reason: \\
& - A: Point A \\
& - B: Point B \\
& - C: Point C \\
& - D: Point D \\
& \textbf{Model Response}: \\
& B: Point B  \\
& The first image's REF point is in the center of the ski lift area, and \\
& Point B in the second image is in the same relative position, maintaining \\
& the same perspective and orientation. \\
\hline
\end{tabular}
\caption{An Example of Multi-Image Understanding of AndesVL.}
\label{tab:general_case_multiimage_3}
\end{table}

\begin{table}[h]
\centering
\begin{tabular}{c|l}
\hline
\multirow{8}*{{\includegraphics[width=3cm]{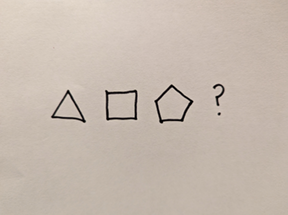}}}
& \textbf{Prompt}: Look at this sequence of three shapes. What shape should come as the\\
& fourth shape? Explain your reasoning with detailed descriptions of the first shapes. \\
& \textbf{Model Response}: \\
& The sequence consists of a triangle, a square, and a pentagon. The fourth shape \\ 
& in the sequence should be a hexagon. The reasoning is as follows: Each shape in \\
& the sequence has one more side than the previous shape. The triangle has three  \\
& sides, the square has four sides, and the pentagon has five sides. Therefore,\\
& the next shape should have six sides, which is a hexagon.\\
\hline
\end{tabular}
\caption{A visual reasoning example of AndesVL.}
\label{tab:general_case_reasoning_1}
\end{table}

\begin{table}[h]
\centering
\begin{tabular}{c|l}
\hline
\multirow{40}*{{\includegraphics[width=3cm]{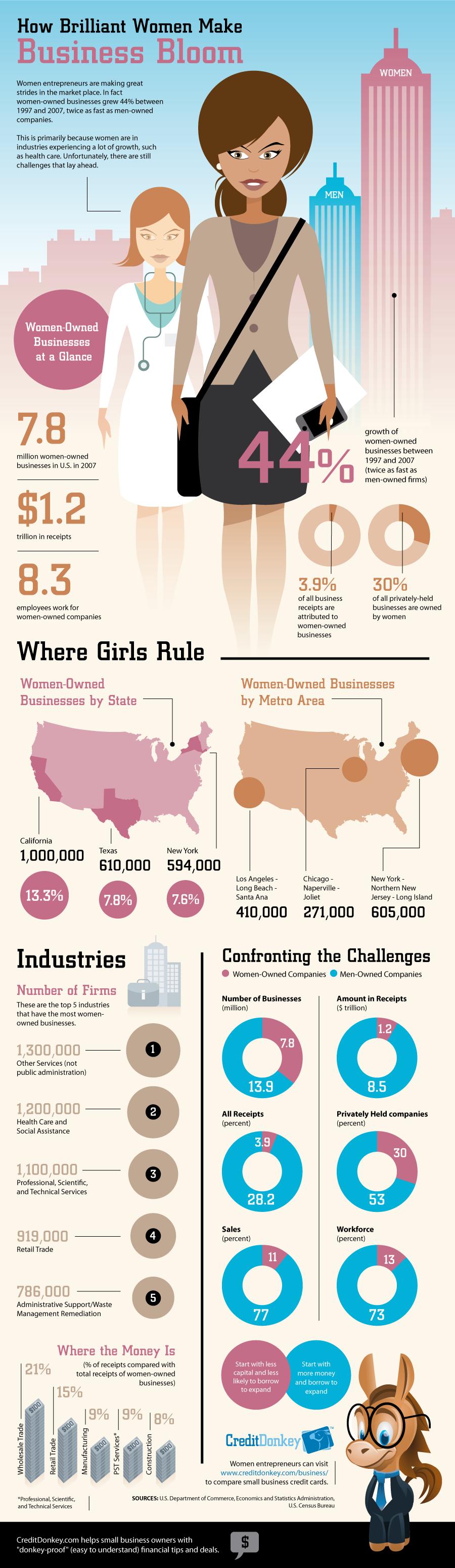}}}

& \textbf{Prompt}: Make a detailed summary of the image.\\
& \textbf{Model Response}: \\
& This image is an infographic titled "How Brilliant Women Make Business \\
& Bloom." \\
&It highlights the growth and impact of women-owned businesses in the United \\
& States.  Here's a detailed summary of the content:\\
& 1. Title and Introduction:\\
& \qquad $\cdot$ The infographic begins with the title "How Brilliant Women Make Business\\
&  \qquad Bloom." \\
&  \qquad $\cdot$ A brief introduction mentions that women entrepreneurs are making   \\
& \qquad significant strides in the marketplace. It states that women-owned businesses  \\
& \qquad grew by 44\% between  1997 and 2007 ...\\
& 2. Where Girls Rule:\\
& \qquad $\cdot$ Women-Owned Businesses by State:\\
& \qquad \quad $\cdot$ California: 1,000,000 businesses (13.3\% of total) \\
& \qquad \quad ...\\
& \qquad $\cdot$ Women-Owned Businesses by Metro Area:\\
& \qquad \quad $\cdot$ Los Angeles - Long Beach - Santa Ana: 410,000 businesses. \\
& \qquad \quad ...\\
& 3. Industries:\\
& \qquad The infographic lists the top 5 industries with the most women-owned \\
& \qquad businesses: \\
& \qquad 1. Other Services (not public administration): 1,300,000 businesses.\\
& \qquad ... \\
& 4. Confronting the Challenges:\\
& ...\\
& Additional Information:\\
& \qquad $\cdot$ The infographic also includes a note on financial tips and deals, mentioning  \\ 
& \qquad CreditDonkey.com, which helps small business owners with "donkey-proof" \\
& \qquad financial advice.\\
& \qquad ... \\
& Visual Elements:\\
& ...\\
\\
& \textbf{Prompt}: Which state has the second highest number of women-owned businesses? \\
& Answer the question with detailed explanation.\\
& \textbf{Model Response}: \\
& According to the text, Texas has 610,000 women-owned businesses, which is the   \\
& second highest number after California. The text also provides percentages for the    \\
& number of women-owned businesses in different industries and metro areas.      \\
& However, the question specifically asks for the state with the second highest number   \\
& of women-owned businesses, which is Texas.\\
\hline

\end{tabular}
\caption{Demonstration of AndesVL’s ability of accurately extracting and comprehensing texts and visual elements from an image, understanding the user’s query, and retrieving the answer from the document. }
\label{tab:case_ocr_1}
\end{table}

\begin{table}[h]
\centering
\begin{tabular}{l|l}
\hline
\multirow{16}*{{\includegraphics[width=3cm]{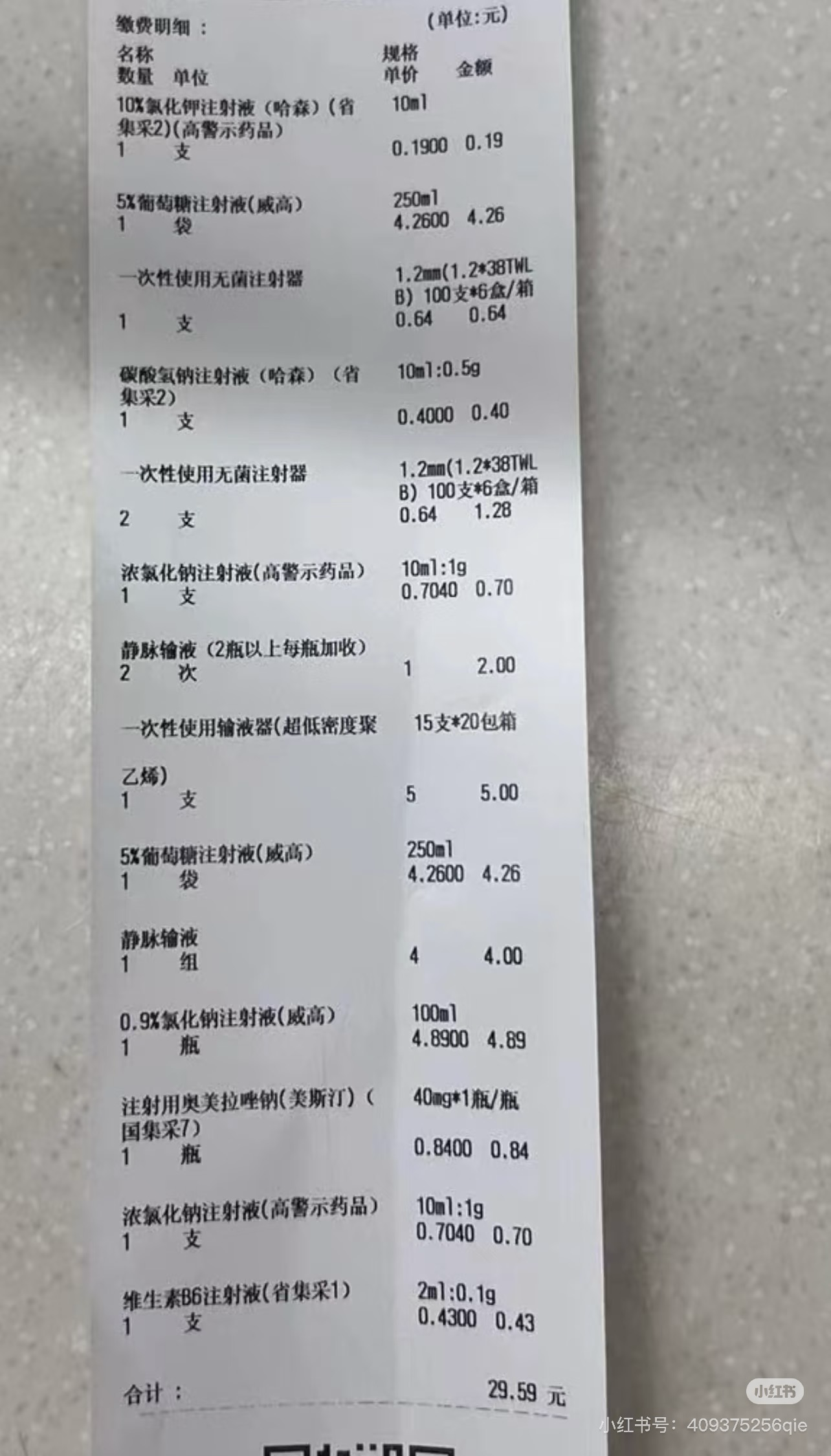}}} \\

& \textbf{Prompt}: Parse the image into a markdown table.\\
& \textbf{Model Response}: \\
& ```markdown \\
& |10\%氯化钾注射液（哈森）（省集采2）(高警示药品)|10ml|1 支|0.1900|0.19| \\
& |5\%葡萄糖注射液(威高)|250ml|1 袋|4.2600|4.26| \\
& |一次性使用无菌注射器|1.2mm(1.2*38TWLB)100支*6盒/箱|1 支|0.64|0.64| \\
& |碳酸氢钠注射液（哈森）（省集采2）|10ml:0.5g|1 支|0.4000|0.40| \\
& |一次性使用无菌注射器|1.2mm(1.2*38TWLB)100支*6盒/箱|2 支|0.64|1.28| \\
& |浓氯化钠注射液(高警示药品)|10ml:1g|1 支|0.7040|0.70| \\
& |静脉输液（2瓶以上每瓶加收）|1|2 次|2.00|2.00| \\
& |一次性使用输液器(超低密度聚乙烯)|15支*20包/箱|1 支|5.00|5.00| \\
& |5\%葡萄糖注射液(威高)|250ml|1 袋|4.2600|4.26| \\
& |静脉输液|4|1 组|4.00|4.00| \\
& |0.9\%氯化钠注射液(威高)|100ml|1 瓶|4.8900|4.89|\\
& |注射用奥美拉唑钠(美斯汀)（国集采7）|40mg*1瓶/瓶|1 瓶|0.8400|0.84|\\
& |浓氯化钠注射液(高警示药品)|10ml:1g|1 支|0.7040|0.70|\\
& |维生素B6注射液(省集采1)|2ml:0.1g|1 支|0.4300|0.43|\\
& |合计：|29.59|\\
& ```\\
\hline
\end{tabular}
\caption{An example of extracting multilingual texts from a long receipt. }
\label{tab:case_ocr_2}
\end{table}

\begin{table}[h]
\centering
\begin{tabular}{c|l}
\hline
\multirow{30}*{{\includegraphics[width=4cm]{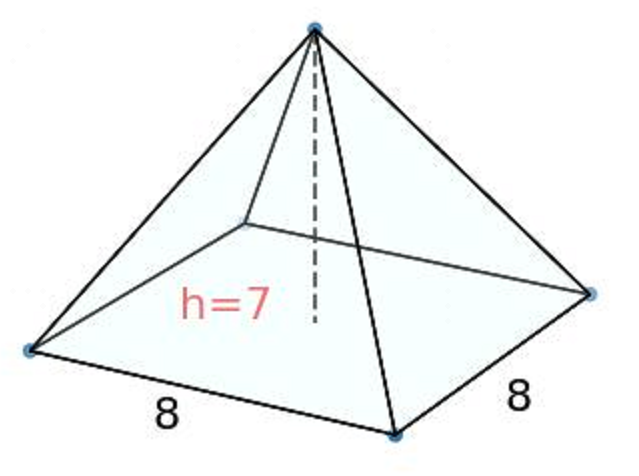}}}
& \textbf{Prompt}: What is the volume of this azure right square pyramid?\\
& \textbf{Model Response}: \\
& <think> \\
& Got it, let's see. The problem is to find the volume of a right square pyramid. \\
& First, I need to recall the formula for the volume of a pyramid. The volume \( V \) \\
& of a pyramid is \( \frac{1}{3} \times \text{base area} \times \text{height} \). \\
& First, identify the base. The base is a square, so the base area is side length \\
& squared.\\
& From the diagram, the side length of the square base is 8. The height \( h \) of the \\
& pyramid is given as 7.\\
& So, base area \( = 8 \times 8 = 64 \). Then, volume \( = \frac{1}{3} \times 64 \times 7 \).\\
& ... \\
& So the answer is 149.33 (or 149.333... but let's present it as a floating number).\\
& </think> \\
& To find the volume of a right square pyramid, we use the formula:\\
& $V = \frac{1}{3} \times \text{base area} \times \text{height}$
\\
&  Step 1: Determine the base area \\
& The base of the pyramid is a square with side length \( 8 \).\\
& The area of a square is \( \text{side}^2 \), so:\\
& $\text{Base area} = 8 \times 8 = 64$\\
&  Step 2: Identify the height of the pyramid\\
& From the diagram, the height \( h \) of the pyramid is \( 7 \).\\
&  Step 3: Calculate the volume\\
& Substitute the base area and height into the volume formula:\\
& $V = \frac{1}{3} \times 64 \times 7$\\
& First, multiply \( 64 \times 7 = 448 \). Then divide by 3:\\
& $V = \frac{448}{3} \approx 149.333\ldots $\\
&  Step 4: Present the result as a floating number\\
& Rounding to two decimal places, the volume is \( 149.33 \).\\
& the answer is 149.33\\
\hline
\end{tabular}
\caption{An example of solving a geometry problem.}
\label{tab:case_math_1}
\end{table}

\begin{table}[h]
\centering
\begin{tabular}{c|l}
\hline
\multirow{40}*{{\includegraphics[width=4cm]{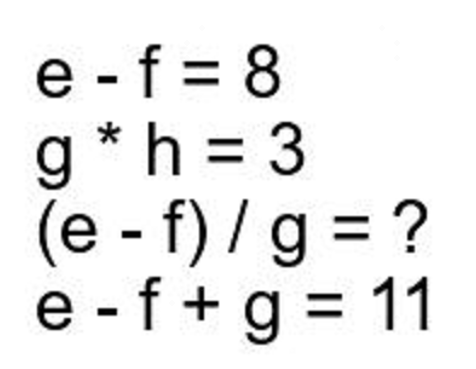}}}
& \textbf{Prompt}: What is the value of (e - f) / g?\\
& \textbf{Model Response}: \\
& <think> \\
& Got it, let's try to solve this problem. First, we need to find the value of (e - f)/g.\\
& Let's list out the given equations: \\
& 1. e - f = 8 \\
& 2. g * h = 3 \\
& 3. e - f + g = 11 \\
& First, let's find the value of g. From equation 3: e - f + g = 11. We know from \\
& equation 1 that e - f = 8. So substitute that into equation 3: 8 + g = 11. \\
& Solving for g, we get g = 11 - 8 = 3. So g is 3.\\
& Now, we need to find (e - f)/g. We know e - f is 8 (from equation 1) and g is 3. \\
& So substitute those values in: 8 / 3. Let's compute that. 8 divided by 3 is \\
& approximately 2.666..., which as a floating number is 2.666666... (repeating). \\
& But we need to present it as a floating number. Let's check if we need to round or \\
& keep the exact value. Since it's a math problem, 8/3 is approximately 2.666...,  \\
& so we can write it as 2.6667 (if we round to four decimal places) or just  \\
& 2.666666... but let's confirm. \\
& Wait, let's make sure we didn't miss anything. Let's check the other equations. \\
& ...\\
& he problem says "convert the final result as a floating number". So 8 divided by \\
&  3 is approximately 2.666666..., so we can write it as 2.666666... or 2.6667.  \\
& Let's go with 2.666666... (six decimal places) or 2.6667 (four decimal places).  \\
& Let's pick 2.666666... as the floating number.\\
& </think>\\
& To solve for the value of (e - f)/g, we analyze the given equations \\
& step by step:\\
& Step 1: Find the value of g \\
& We know: \\
&  e - f = 8 (from the first equation).\\
& e - f + g = 11 (from the fourth equation).\\
& Substitute e - f = 8 into e - f + g = 11:\\
& 8 + g = 11 \\
& Solve for g: \\
& g = 11 - 8 = 3 \\
& Step 2: Calculate (e - f)/g \\
&...\\
& the answer is 2.666666\\
\hline
\end{tabular}
\caption{An example of solving an algebra problem. }
\label{tab:case_math_2}
\end{table}

\begin{table}[h]
\centering
\begin{tabular}{c|l}
\hline
\multirow{40}*{{\includegraphics[width=4cm]{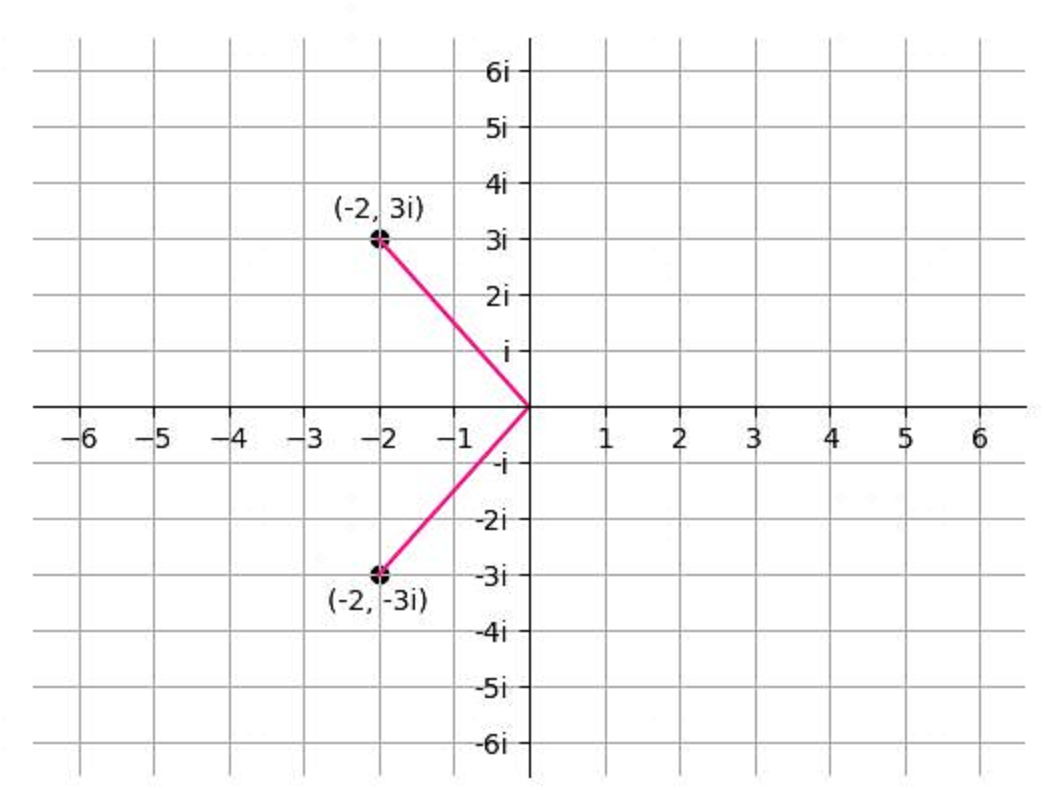}}}
& \textbf{Prompt}: Find the value of $z * (z^*)$.\\
& \textbf{Model Response}: \\
& <think> \\
& Got it, let's try to solve this problem. First, we need to find the value of z \\
 &multiplied by its complex conjugate $z^*$. \\
 &Let's start by identifying the complex number z from the graph.\\
& Looking at the graph, the point (-2, 3i) corresponds to the complex number z = -2 + 3i. \\
& ... \\
& Alternatively, using the formula (a + bi)(a - bi) = a² + b². \\
 &Here, a = -2, b = 3. So a² = (-2)² = 4, b² = 3² = 9.\\
 &Then 4 + 9 = 13.\\
 &So the value of $z *(z^*)$ is 13.\\
& </think>\\
& To solve for $z * (z^*)$, follow these steps:\\
& Step 1: Identify the complex number $z$ from the graph\\
& The graph shows a point at $(-2, 3i)$. \\
 &For a complex number $z = a + bi$, the real part $a$ is the x - coordinate, \\
 &and the imaginary part $b$ is the y - coordinate. Thus, $z = -2 + 3i$.\\
& Step 2: Find the complex conjugate $z^*$\\
& The complex conjugate of $z = a + bi$ is $z^* = a - bi$.\\
 & For $z = -2 + 3i$, the conjugate is $z^* = -2 - 3i$.\\
& Step 3: Calculate $z * z^*$\\
& For any complex number $z = a + bi$, \\
 &the product $z * z^*$ simplifies to $a^2 + b^2$ \\
 &(derived from multiplying $(a + bi)(a - bi)$).\\
& For $z = -2 + 3i$:\\
 &- Real part $a = -2$, so $a^2 = (-2)^2 = 4$.\\
 &- Imaginary part $b = 3$, so $b^2 = 3^2 = 9$.\\
 &Adding these: $a^2 + b^2 = 4 + 9 = 13$.\\
&  The answer is $13$.\\ 
\hline
\end{tabular}
\caption{An example of solving an analytic geometry problem. }
\label{tab:case_math_3}
\end{table}

\begin{table}[h]
\centering
\begin{tabular}{l|l}
\hline
\multirow{16}*{{\includegraphics[width=3cm]{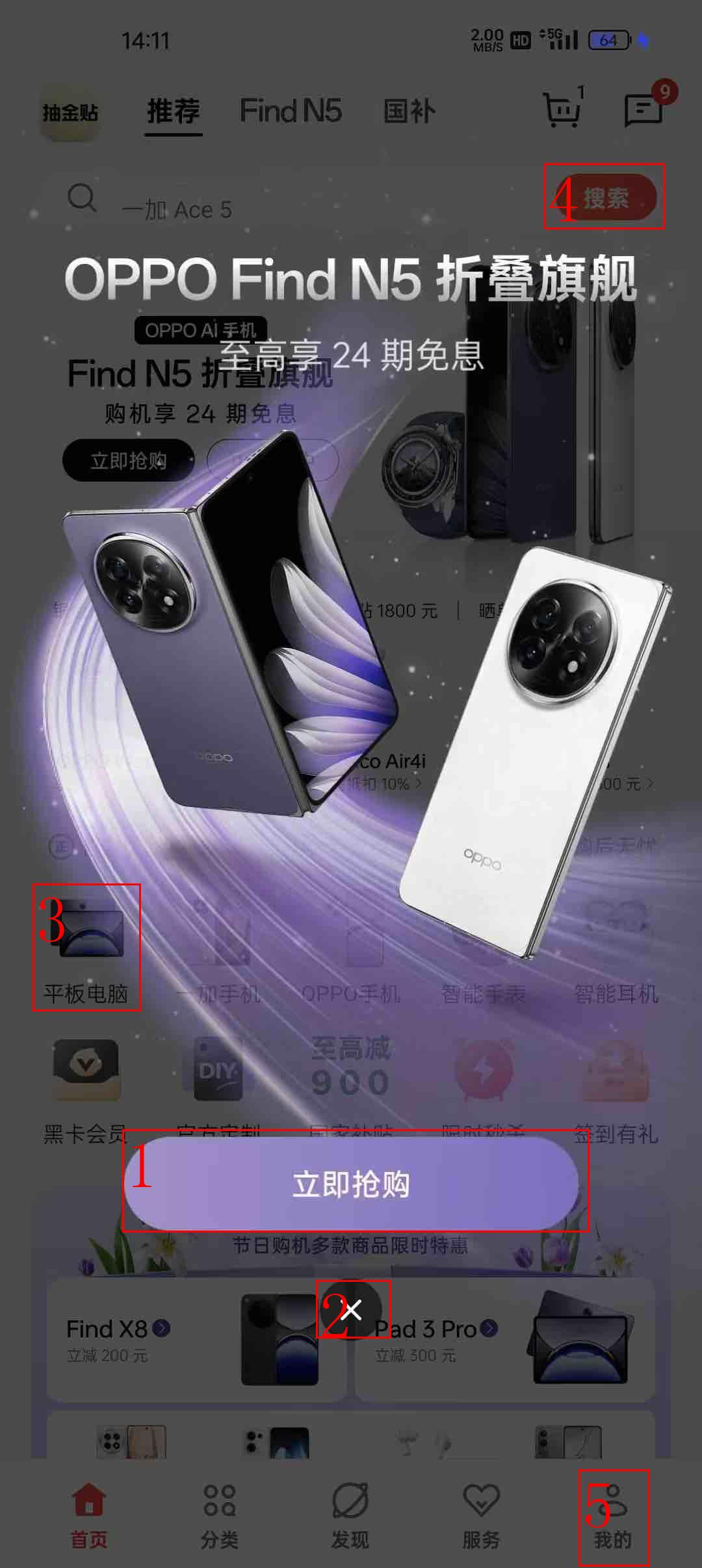}}} \\

& \textbf{Prompt}: 当前页面有弹窗吗？\\
& \textbf{Model Response}: 是的，当前页面有一个弹窗，\\
& 控件(1741,7204),(8398,7858)是一个文本按钮, 描述为“立即抢购”。\\
& \textbf{Prompt}: 如何关闭当前页面的弹窗？\\
& \textbf{Model Response}: 您可以点击控件(4509,8167),(5565,8533)，\\
& 它是一个图标（Icon），描述为'关闭'。\\
& \textbf{Prompt}: 我想找平板电脑，应该怎么办？\\
& \textbf{Model Response}: 您可以点击控件(472,5638),(2000,6450)，\\
& 它是一个文本按钮，描述为“平板电脑”。\\
& \textbf{Prompt}: 我想查看我的订单，应该怎么办？\\
& \textbf{Model Response}: 您可以点击控件(8241,9371),(9250,9988)，\\
& 它是一个多标签（Multi-Tab），描述为“我的”，进入后可以查看您的订单。\\
\\
\\
\\
\\
\hline
\end{tabular}
\caption{An example of Mobile UI understanding and reasoning.}
\label{tab:case_ui_1}
\end{table}

\begin{table}[h]
\centering
\begin{tabular}{l|l}
\hline
\multirow{16}*{{\includegraphics[width=3cm]{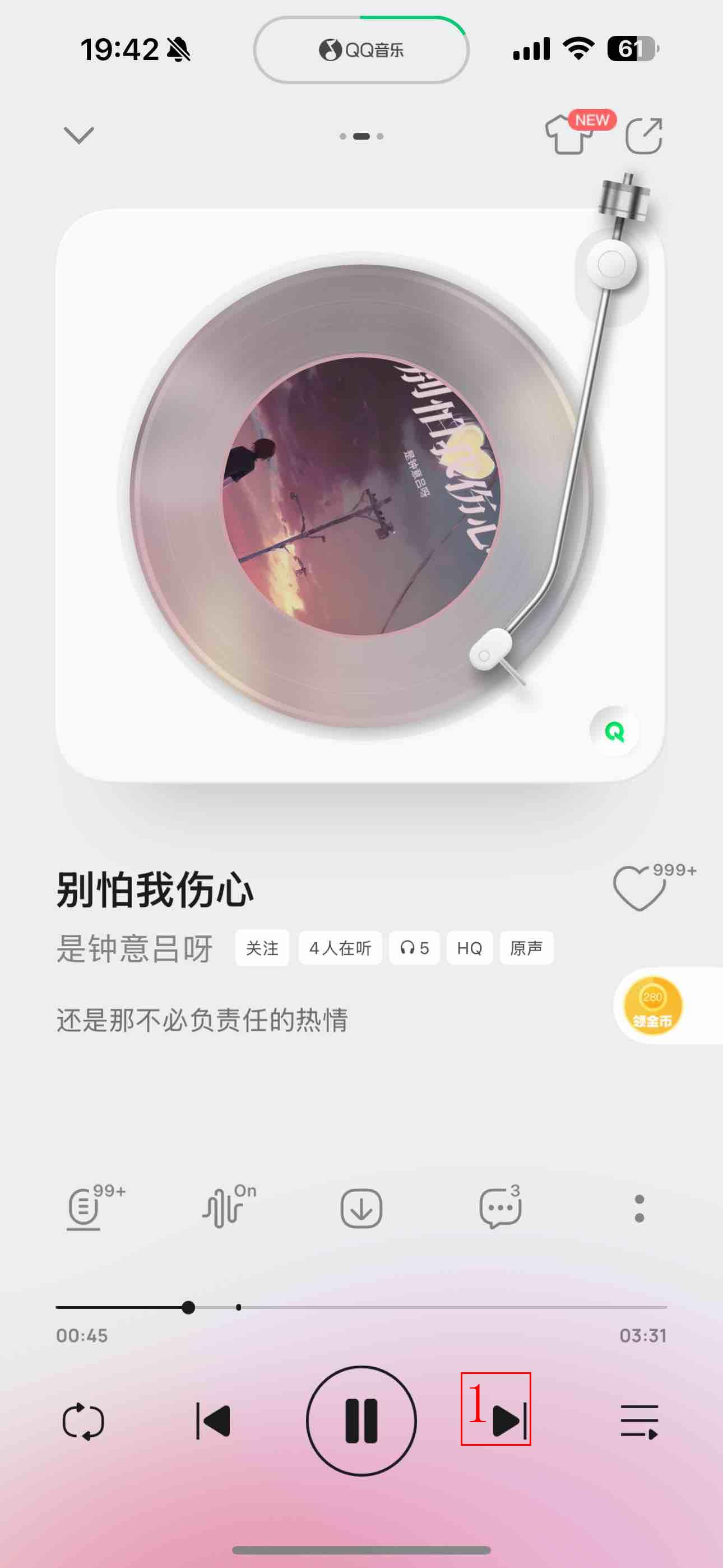}}} \\

& \textbf{Prompt}: 这是什么页面？\\
& \textbf{Model Response}: 这是音乐APP中的播放页面，用于控制和播放音乐。\\
& \textbf{Prompt}: 现在正在播放什么音乐？\\
& \textbf{Model Response}: 正在播放的音乐是'别怕我伤心'。\\
& \textbf{Prompt}: 如何切换下一曲？\\
& \textbf{Model Response}: 点击控件(6376,8752),(7342,9218)。\\
\\
\\
\\
\\
\\
\\
\\
\\
\\
\\
\hline
\end{tabular}
\caption{An example of Mobile UI understanding and reasoning.}
\label{tab:case_ui_2}
\end{table}

\end{CJK}
\end{document}